\documentclass[journal]{IEEEtran}
\usepackage{pgf}
\usepackage{tikz}
\usepackage{cite}
\usepackage{amsmath,amssymb,amsfonts,bm}
\usepackage{color,colortbl}
\usepackage{algorithmic}
\usepackage{array}
\usepackage{url}
\usepackage{algorithm}           
\usepackage{algorithmic}            
\usepackage{float}
\usetikzlibrary{arrows,automata}

\graphicspath{{images/}}

\ifCLASSOPTIONcompsoc
  \usepackage[caption=false,font=normalsize,labelfont=sf,textfont=sf]{subfig}
\else
  \usepackage[caption=false,font=footnotesize]{subfig}
\fi

\begin{document}
\title{Patchwise Joint Sparse Tracking with Occlusion Detection}
\author{Ali~Zarezade,
       Hamid R. Rabiee,~\IEEEmembership{Senior Member,~IEEE}, Ali~Soltani-Farani, Ahmad Khajenezhad
\thanks{A. Zarezade, H. R. Rabiee, A. Soltani-Farani, and A. Khajenezhad are with the Department of Computer Engineering, Sharif University of Technology, Tehran, Iran.}}

\maketitle

\begin{abstract}
This paper presents a robust tracking approach to handle challenges such as occlusion and appearance change. Here, the target is partitioned into a number of patches. Then, the appearance of each patch is modeled using a dictionary composed of corresponding target patches in previous frames. In each frame, the target is found among a set of candidates generated by a particle filter, via a likelihood measure that is shown to be proportional to the sum of patch-reconstruction errors of each candidate. Since the target's appearance often changes slowly in a video sequence, it is assumed that the target in the current frame and the best candidates of a small number of previous frames, belong to a common subspace. This is imposed using joint sparse representation to enforce the target and previous best candidates to have a common sparsity pattern. Moreover, an occlusion detection scheme is proposed that uses patch-reconstruction errors and a prior probability of occlusion, extracted from an adaptive Markov chain, to calculate the probability of occlusion per patch. In each frame, occluded patches are excluded when updating the dictionary. Extensive experimental results on several challenging sequences shows that the proposed method outperforms state-of-the-art trackers.
\end{abstract}


\IEEEpeerreviewmaketitle

\section{Introduction}
\IEEEPARstart{V}{isual tracking} is a classic computer vision problem that remains challenging after more than three decades.
There is a high demand for visual tracking in many computer vision applications, where it is directly or indirectly required to track an object in a video sequence. To name a few, consider traffic control, robotics, sports, gaming, augmented reality, and gesture recognition.

Given a bounding box defining the object of interest (target) in the first frame of a video sequence, the goal of a tracker, is to determine the object's bounding box in subsequent frames.
Primary challenges encountered in visual tracking are target appearance change and occlusion, while other challenges arise from variation in illumination, scale, and camera motion (see Table \ref{tbl:ds-attr}).
A general tracker is composed of three main components: 1) A \textit{Motion model}, which generates a number of potential bounding boxes as target candidates in each frame. 2) An \textit{Appearance model}, which models the appearance of the target  throughout the video sequence. 3) An \textit{Observation Model}, which defines a measure of similarity (likelihood) to determine which candidate is most likely to be generated by the appearance model. This candidate is then selected as the target in the current frame. The focus of this paper is on the two latter components.

Recently, sparse representation has shown appealing results in computer vision applications, such as face recognition \cite{Wright2009}, image super-resolution \cite{Yang2010}, background subtraction \cite{Cevher2008}, denoising \cite{Aharon2006}, and classification \cite{Mairal12}.
Motivated by the work of Wright et. al \cite{Wright2009} in face recognition, \cite{Mei2011} used sparse representation for object tracking, which paved the way for other sparse trackers \cite{Mei2011a,Bao2012,Jia,Liu2011a,Zhong2012}.
The main idea of \cite{Mei2011} is to model the object with a linear combination of few target images in previous frames (target templates) plus a linear combination of canonical basis vectors (trivial templates). The true candidate is distinguished from other candidates, since it is expected to be represented mostly by the target templates. Accordingly, the likelihood of a candidate is selected to be inversely proportional to its reconstruction error with the target templates.
The main drawback of this method is that, 
 in cases of sever occlusion the nonzero coefficients rarely correspond to the target template \cite{Wright2009} and therefore the likelihood measure will no longer indicate the "true" candidate. Many other trackers \cite{Mei2011,Mei2011a,Bao2012,Zhang,Zhang2012,Zhang2012a,Zhang2013a,Wang2013,Wang2012a} have also taken a holistic view,  in which the target is treated as a single entity. In contrast, other algorithms view the object image as a collection of small patches in order to better handle partial occlusion \cite{Adam2006,Jia,Liu2011a,Zhong2012}. In these methods, occlusion only affects a small subset of the patches where it occurs, and can therefore be treated more efficiently \cite{Wu2013}. 

In this paper, we propose a robust tracking algorithm by taking advantage of the similarity between target objects in consecutive frames (temporal similarity assumption). Inspired by the success of patchwise methods and sparse representation for visual tracking, we consider the target object as a collection of non-overlapping patches and assume that a true candidate patch accompanied by the target patches found in previous frames, have sparse representations with a common sparsity pattern (i.e. belong to a common subspace). This is accomplished by using a regularization term that induces joint-sparsity. The appearance of each patch is modeled by a collection of corresponding target patches in previous frames, called a patch template. To increase the discriminative power of the algorithm, we model the object appearance using a dictionary composed of all the patch templates. This enables us to measure the overall likelihood of a candidate as a sum of patchwise reconstruction errors. The advantage of this likelihood measure is that in cases of  partial occlusion, as long as there are patches which are not totally occluded, it will indicate the best candidate. Another advantage of building the appearance model using patch templates, is that we are able to take advantage of the patchwise reconstruction errors to determine occluded patches, and exclude them from the dictionary update procedure. To summarize, the contributions of this work are as follows:
\begin{itemize}
\item Introduction of a patchwise likelihood measure computed as the sum of patch-reconstruction errors, hence increasing tracker robustness to misalignment and partial occlusion.
\item Taking into account the temporal similarity assumption through joint sparse representation, and therefore increasing tracker robustness to appearance change.
\item 
Removing occluded patches from the dictionary update procedure, using patchwise occlusion detection based on an adaptive Markov model.
\end{itemize}

The remainder of this paper is organized as follows. Section \ref{sec:prior-work} discusses prior work by categorizing tracking algorithms into generative and discriminative classes, and provides a more detailed description of  methods that exploit sparse representation. In Section \ref{sec:pf}, the motion model used in our algorithm, which is based on particle filter is introduced. The proposed tracker is introduced in Section \ref{sec:proposed-method} . Experimental results are presented in Section \ref{sec:experiment}, and the concluding remarks are provided in Section \ref{sec:conclusion}.

\section{Prior Works and Motivation} \label{sec:prior-work}
The literature on visual object tracking is extensive \cite{Yilmaz2006}, and two major categories of visual tracking are discriminative and generative \cite{Bao2012,Mei2011a,Liu2011a}. In this section, we briefly review the main works of each category. Afterwards, a more detailed review of significant tracking methods based on sparse representation is given.
For a recent survey of sparse coding based trackers, the interested reader may refer to \cite{Zhang2013}.
 
\subsection{Discriminative Trackers}
In discriminative methods, 
a classifier is trained to discriminate the target object from the background. Often, a likelihood measure is considered as the classification score. Regarding the training scheme of the classifier, the appearance model may be static \cite{Avidan2004} or adaptive \cite{Avidan2007,Grabner2006,Babenko2011,Kalal2012}.
In static models, the classifier is learned off-line using a training set consisting of object and background images. For example, Avidan proposed an off-line trained SVM for car tracking \cite{Avidan2004}. Although, static discriminative trackers are robust to the extent of the available training data, they can't handle previously unseen appearance variations.
In adaptive approaches, the classifier is trained online during tracking and bootstraps itself with positive and negative training data sampled from the target and background. 
In \cite{Avidan2007,Comaniciu2002,Grabner2006a,Grabner2006,Babenko2011}, an ensemble of weak classifiers is learned online and combined to create a stronger classifier. 
An important issue that causes drift in discriminative trackers is uncertainty in labeling training data for updating the classifier. To overcome this problem, Grabner et al. proposed an online semi-boosting algorithm \cite{Grabner2008}, in which labeled samples are collected only from the first frame. Babenko et al. \cite{Babenko2011}, utilized the idea of multiple instance learning (MIL), in which labels are assigned to a bag of instances instead of each individual instance, to transfer the ambiguity in labeling to the learning algorithm. In \cite{Zeisl2010} both ideas of MIL and semi-supervised learning are combined to take advantage of both approaches. 

\subsection{Generative Trackers}
Generative trackers attempt to model the object appearance and measure the likelihood that a candidate was generated by the model. Generative models can also be static or adaptive. The simplest  appearance model is a single image of the target, called a \emph{template}. Early trackers modeled the object appearance using a fixed template and searched for the candidate with minimum sum of squared distance (SSD) to the template \cite{Hager1998}. The mean-shift tracker \cite{Comaniciu2000,Comaniciu2003}
uses an adaptive appearance model, which is the best candidate of the previous frame. The likelihood measure is defined to measure the similarity between intensity histograms of the appearance model and a candidate. 
Fragment-based tracker \cite{Adam2006} is one of the successful extensions of mean-shift tracker. To handle occlusion it divides the candidate image into a collection of overlapping patches and finds the likelihood by aggregating all patches.

The main limitation of template-based appearance models is that illumination variation can't be handled efficiently.
The work of Turk and Pentland \cite{Turk} called “Eigenfaces”, addressed this limitation. Their technique employed principal component analysis (PCA) to model the target object (face). To overcome the problems caused by illumination variation, it is assumed that the target object belongs to the subspace spanned by a few of the principle components. It has been theoretically proven that an object image under Lambertian reflections belongs to a low dimensional subspace \cite{Basri2003}.
Also it has been empirically observed that even under mild pose change an object belongs to a low-dimensional subspace \cite{Wright2009,Wagner2012,Mei2011}. Based on these properties, Black and Jepson utilized Eigenimage and optical flow to introduce the Eigentracker \cite{Black1998}.
Since the object appearance may change during the video sequence, the appearance model needs to be adaptive and should be updated during tracking. Ross et al. exploit eigenbasis update algorithms and proposed an incremental learning method \cite{Ross2008} which updates eigenbases of the object subspace using incremental SVD with a forgetting factor.

\subsection{Sparse Trackers}
After the aforementioned developments in generative models, sparse representation as a variant of subspace models, was used for visual tracking leading to state-of-the-art results \cite{Mei2011,Mei2011a,Liu2011a,Bao2012,Jia,Zhong2012,Zhang}. 
Generally, a candidate is represented using a linear combination of a few elements (atoms) from a dictionary composed of a number of previously found target images. The coefficients of this representation are used to find the best candidate.
Apart from the ability to handle illumination and mild pose changes, these trackers attempt to tackle occlusion.
Based on how the sparse representation is used, sparse trackers can also be categorized into generative and discriminative approaches. 

Generative sparse approaches model appearance using a dictionary composed of two parts, one of which models the target object while the other is used to model occlusion. 
The likelihood measure is usually defined as a function of the reconstruction error of a candidate using coefficients corresponding to target atoms in the dictionary. As was previously mentioned, the dictionary used by the $\ell_1$ tracker \cite{Mei2011} is composed of target (object) and trivial (occlusion) templates.
Due to the large size of the trivial templates, the computational complexity of this tracker is high, and since it lacks an occlusion detection method, the dictionary may be contaminated by occlusion which causes the tracker to drift and eventually lose the target.
Mei et. al addressed these two drawbacks \cite{Mei2011a} by solving the $\ell_1$ minimization for a reduced set of candidates that are likely to possess a lower reconstruction error and then detecting occlusion by deriving an occlusion map from the sparse coefficients. The dictionary was updated for candidates with occlusion less than some threshold.
To increase tracker accuracy, a new occlusion-aware minimization is proposed in \cite{Bao2012}, which is solved using a customized version of APG \cite{Tseng2008}.
In \cite{Zhang2012a}, the problem of finding the sparse representation of the candidates is cast into low-rank sparse matrix learning. The sparse representation of all candidates is found simultaneously by minimizing their reconstruction error accompanied with an $\ell_1$  and low-rank regularization term.
To take advantage of the relationship between particles and increase tracker speed, Zhang et. al proposed to group all the candidates in a frame and find their joint-sparse representation, simultaneously \cite{Zhang,Zhang2013a}. Since the number of candidates that don't belong to the target object's subspace is usually large, the joint-sparse minimization is likely to compensate for these candidates rather than accurately representing the fewer candidates that belong to the target object's subspace. Therefore, the assumption made by Zhang et. al that all candidates are related through a common low-dimensional subspace seems unrealistic.

Discriminative sparse trackers, employ sparse representation to discriminate the target from its background. In \cite{Wang2012a}, the dictionary is learned from SIFT descriptors extracted from a general dataset of object images, and a linear classifier is learned online with positive and negative samples.
To better discriminate the target from background, in \cite{Liu2011a} a dictionary is learned online from target patches. The sparse representation of candidate patches are obtained via Local Linear Coding (LLC) \cite{Wang2010}. Finally, the best candidate is found via regularized Mean-Shift. In \cite{Zhang2012}, a sparse measurement matrix is adopted to extract low dimensional discriminative features and the tracking problem is formulated as an online naive Bayes classifier. In \cite{Jia}, a dictionary composed of overlapping patches from the target is used to find the sparse representation of candidate patches. A method called alignment-pooling is proposed where the likelihoods are found by summation over a feature vector composed of the similarity of each candidate patch to related patches in the target dictionary.



\section{Particle Filter}\label{sec:pf}


Visual tracking problem may be cast as Bayesian filtering. The goal of Bayesian filtering is to find the posterior pdf of a system state $\mathbf{x}_t$, given all observations $\mathbf{z}_{1:t}$. Bayesian filtering is composed of two steps: prediction and update. In the prediction step, with Markov assumption, the prior pdf of the state is obtained using the Chapman-Kolmogorov equation as:
\begin{equation}\label{equ:pf1}
p(\mathbf{x}_t|\mathbf{z}_{1:t-1}) = \int p(\mathbf{x}_t|\mathbf{x}_{t-1}) p(\mathbf{x}_{t-1}|\mathbf{z}_{1:t-1}) d\mathbf{x}_{t-1}
\end{equation}
When a new observation $\mathbf{z}_t$ becomes available, in update step, the posterior is found via Bayes' rule as:
\begin{equation}\label{equ:pf2}
p(\mathbf{x}_t|\mathbf{z}_{1:t}) = \frac{p(\mathbf{z}_t|\mathbf{x}_t) p(\mathbf{x}_t|\mathbf{z}_{1:t-1})}{p(\mathbf{z}_t|\mathbf{z}_{1:t-1})}
\end{equation}
Solving (\ref{equ:pf1}) and (\ref{equ:pf2}) gives the optimal solution of the filtering problem. Particle filter represents an analytical method for solving these equations. The key idea is to represent the posterior by samples (particles) and their associated probabilities (weights). Therefore, updating the posterior is equivalent to updating the particles and their weights. So far, many variants of particle filters have been developed \cite{Chen2003}. We use Sequential Importance Resampling (SIR) filter \cite{Arulampalam2002} in which particles are resampled in each frame to reduce the degeneracy problem. The update equations of SIR filter are:
\begin{equation} \label{equ:SIR}
\mathbf{x}_t^i \sim p(\mathbf{x}_t|\mathbf{x}_{t-1}^i), \: \mathbf{w}_t^i  \sim p(\mathbf{z}_t|\mathbf{x}_t^i) 
\end{equation}
where $\mathbf{x}_t^i$ is the $i$th particle in frame $t$. In visual tracking, state $\mathbf{x}_t$ denotes the affine motion parameters which define the object region in frame $t$, and observations $\mathbf{z}_{1:t}$ are a collection of frames observed so  far. We notice from SIR equations (\ref{equ:SIR}) that visual tracking requires a likelihood evaluation measure $p(\mathbf{z}_t|\mathbf{x}_t^i)$ and the state transition probability $p(\mathbf{x}_t|\mathbf{x}_{t-1}^i)$. Without any prior knowledge, $p(\mathbf{x}_t|\mathbf{x}_{t-1}^i)$ is considered to be a zero mean Gaussian with predefined variance $\Sigma$.
To evaluate $p(\mathbf{z}_t|\mathbf{x}_t^i)$, first the candidate region associated with particle $\mathbf{x}_t^i$ is cropped from frame $t$, then the probability that it is generated by the appearance model is considered as its likelihood.

\section{Proposed Method}\label{sec:proposed-method}
Given the initial state of the target, the goal is to find the target object among the set of candidates generated by the particle filter, in subsequent frames. In order to accomplish this task, we need to model the target's appearance and find the candidate that is most likely to be generated by this model. We  use a dictionary composed of image patches selected from targets found in previous frames to model the target's appearance. The best candidate is assumed to have the least reconstruction error sum, over its patches. In order to support changes in the appearance of the target object, once the best candidate is found, the model needs to be updated accordingly. We attempt to exclude occluded image patches when updating the dictionary with the best candidate. In this section, we describe these steps in detail and intuitively justify the proposed approach.

\subsection{Notation}
In the sequel, bold lower case and bold upper case letters are used for vectors ($\mathbf{d}$) and matrices ($\bm{D}$), respectively. 
For any variable, the superscript $(i)$ is used to indicate that the variable is related to the $i$th patch. 
Suppose that $\Lambda\subset\{1,2,\ldots,n\}$ is a subset of indices, the complement of which is denoted by $\Lambda^c=\{1,2,\ldots,n\}\backslash \Lambda$. By $\mathbf{d}_\Lambda$ we mean the vector obtained from $\mathbf{d}$ by eliminating indices inside $\Lambda^c$. Similarly, by $\bm{D}_{\Lambda}$ we mean the matrix obtained by removing from $\bm{D}$ the columns indexed by $\Lambda^c$.

\begin{figure}[!t]
\centering
\includegraphics[width=0.35\textwidth]{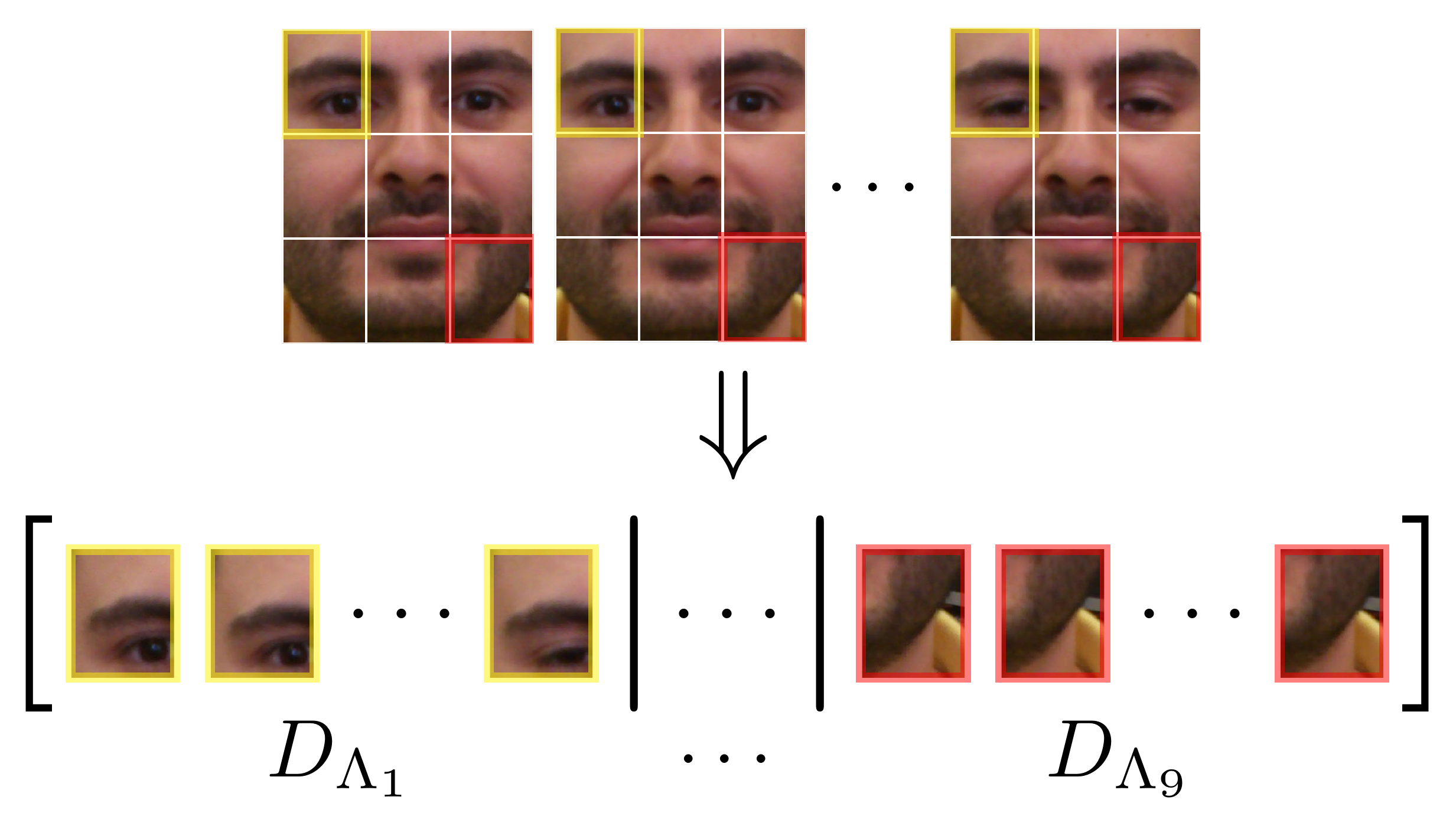}
\caption{\textbf{Appearance Dictionary}. Target appearance is modeled by a dictionary composed of patch templates. Each patch template $\bm{D}_{\Lambda_i}$ is used to model the corresponding target patch, and is composed of a number of target patches found in previous frames.}
\label{fig:dic}
\end{figure}
\subsection{Appearance Model}
Given a dictionary composed of a number of the best candidates found in previous frames, the target's appearance is modeled with a linear combination of a few elements (atoms) from this dictionary. In other words, it is assumed that the target belongs to a low-dimensional linear subspace spanned by a few of the dictionary atoms. Linear subspaces are rich appearance models that can handle illumination and pose change. It has been theoretically proven that an object under different illumination lies on a low-dimensional subspace \cite{Basri2003}. Also, \cite{Wagner2012} empirically showed that a dictionary of pose varying face images is able to model pose change. We take a patchwise view and model the target's appearance with a dictionary $\bm{D}$, composed of patches selected from $n$ previous best candidates (see Fig. \ref{fig:dic}):
\begin{equation}\label{equ:pbg_dict}
\bm{D} = \Bigg[\overbrace{\mathbf{d}^{(1)}_1 \cdots \mathbf{d}^{(1)}_n}^{\bm{D}_{\Lambda_1}} \Big| \cdots \Big| \overbrace{\mathbf{d}^{(m)}_1 \cdots \mathbf{d}^{(m)}_n}^{\bm{D}_{\Lambda_m}}\Bigg]  
\end{equation}
where $\mathbf{d}^{(j)}_{i} \in \mathbb{R}^M$ is a vectorized grayscale image of the $j$th patch of the $i$th target placed in the dictionary, $\Lambda_j$ denotes the set of indices of the atoms that belong to the $j$th patch in all targets, $\bm{D}_{\Lambda_j}$ is the $j$th patch template, and $\bm{D} \in R^{M \times N}$. The number of dictionary atoms is $N=nm$, where $n$ and $m$ are the number of targets and number of patches in each target image, respectively.

Before further discussing the model, we look into two issues regarding the proposed appearance model, namely feature extraction and patchwise treatment of the target. Among features used by sparse trackers are grayscale \cite{Mei2011,Liu2011a,Jia}, SIFT \cite{Wang2012a}, block-division based covariance \cite{Zhang2012c}, and covariance matrix \cite{Wu2011}. Most feature extraction methods involve only linear (or approximately linear) operations, and it has been shown that linear transform of grayscale features under mild conditions has no considerable effect on face recognition using sparse representation \cite{Wright2009}. On the other hand, feature extraction methods such as SIFT or Haar \cite{Babenko2011,Grabner2006a,Grabner2008} are usually accompanied by information loss, which reduce the discriminative ability of the tracker. Considering these observations and to reduce computational complexity, we use grayscale features. The other major issue is the target representation scheme. The common approach is to view the target as a single entity, in which the dictionary is a collection of vectorized  images of the target in previous frames, in contrast to the dictionary in (\ref{equ:pbg_dict}) which is composed of vectorized patches. Wright et. al \cite{Wright2009} showed that severe occlusions cause a considerable decrease in face recognition performance. Also, the results of a recent benchmark \cite{Wu2013} shows that trackers with a patchwise view  \cite{Adam2006,Jia,Liu2011a} are more robust to occlusion than methods with a holistic view \cite{Mei2011,Bao2012,Zhang}. The advantages of the proposed target appearance model are twofold: improved performance in handling occlusion and considerable speed up due to reduced dictionary size compared to \cite{Mei2011}.

\begin{figure*}[t]
\centering
\includegraphics[width=0.85\textwidth]{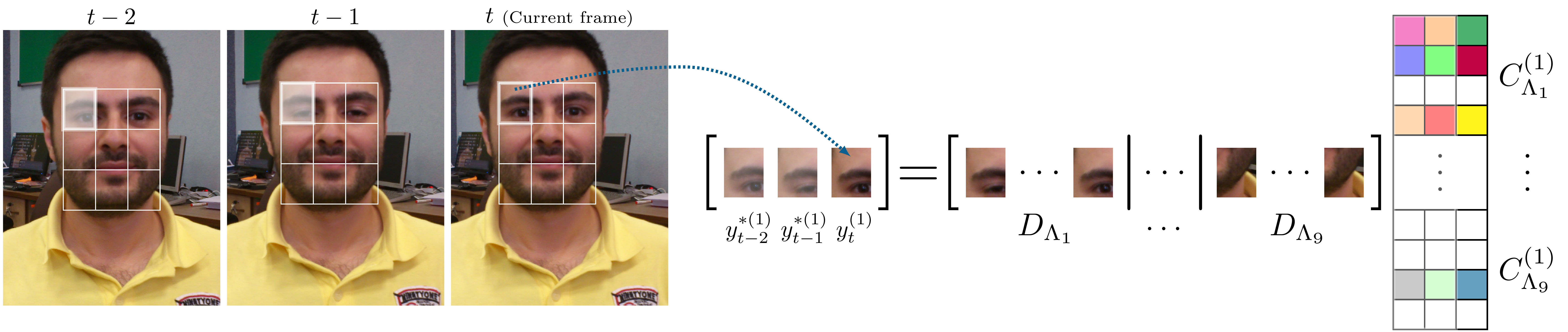}
\caption{\textbf{Joint sparse representation of a patch.} To find the sparse representation of a candidate patch, it is grouped with corresponding patches from the two previous best candidates. The joint-sparse minimizations (\ref{equ:l2l0}) or (\ref{equ:l2l1}), force the sparse coding of the signals in a group to have similar support, as depicted in the row-sparse matrix $C^{(1)}$.
}
\label{fig:grouping}
\end{figure*}

In the current frame, the $i$th patch of the target $\mathbf{y}$, denoted by $\mathbf{y}^{(i)}$, is assumed to belong to a low-dimensional subspace spanned by the columns of the $i$th patch template $\bm{D}_{\Lambda_i}$. Considering the appearance variations of the target and presence of noise, each patch template is expected to be a full-rank matrix. Since different patch templates $\bm{D}_{\Lambda_j}$ correspond to different non-overlapping patches of the target, and each patch template is likely to be full-rank, it is reasonable to assume that the collection of patch templates $\bm{D}_{\Lambda_j},j\ne i$, denoted by $\bm{D}_{\Lambda_i^c}$ is likely to be overcomplete. Hence, any occlusion or noise in $\mathbf{y}^{(i)}$ may be modeled as $e^{(i)}=\bm{D}_{\Lambda_i^c}c^{(i)}_{\Lambda_i^c}$ leading to our representation of $\mathbf{y}^{(i)}$ as:
\begin{equation} \label{equ:patch-repres-noise}
\mathbf{y}^{(i)} = \bm{D}_{\Lambda_i} \mathbf{c}^{(i)}_{\Lambda_i} + e^{(i)}= \bm{D}_{\Lambda_i} \mathbf{c}^{(i)}_{\Lambda_i} + \bm{D}_{\Lambda_i^c} \mathbf{c}^{(i)}_{\Lambda_i^c} = \bm{D} \mathbf{c}^{(i)}
\end{equation}
where $c^{(i)}$ is the vector of representation coefficients. As mentioned before, the appearance model dictionary $\bm{D}$ is expected to be overcomplete. Therefore, the system of linear equations (\ref{equ:patch-repres-noise}) is underdetermined, and has no unique solution for $\mathbf{c}^{(i)}$. When occlusion and noise is negligible, we can expect $\mathbf{c}^{(i)}$ to be sparse. This sparsity may be imposed by solving:
\begin{equation}\label{equ:l0-minimization}
\underset{\mathbf{c}^{(i)}}{\text{argmin}} \; \|\mathbf{c}^{(i)}\|_0\quad\text{s.t.} \quad \mathbf{y}^{(i)}=\bm{D}\mathbf{c}^{(i)} 
\end{equation}
Taking into account any shortcomings due to the linear nature of the model, a relaxed form \cite{Bruckstein2009} is usually solved by using at most $L$ atoms:
\begin{equation}\label{equ:l0-minimization2}
\underset{\mathbf{c}^{(i)}}{\text{argmin}} \;  \frac{1}{2}\|\mathbf{y}^{(i)}-\bm{D}\mathbf{c}^{(i)}\|_2^2 \quad \text{s.t.} \quad \|\mathbf{c}^{(i)}\|_0 \le L
\end{equation}

Considering that the target's appearance changes smoothly, it is plausible to assume that the target objects, over a few consecutive frames, belong to the same subspace. We apply this assumption by enforcing the sparse representation of the current target patch $\mathbf{y}^{(i)}$ to be jointly sparse (i.e. have the same sparsity pattern)  with the same patch in the previous $k$ best candidates, i.e. $\mathbf{y}_{t-k}^{*(i)},\ldots,\mathbf{y}_{t-1}^{*(i)}$. To this end, we solve the following optimization problem:
\begin{equation}\label{equ:l2l0}
\underset{\bm{C}^{(i)}}{\text{argmin}} \; \frac{1}{2}\|\bm{Y}^{(i)}-\bm{DC}^{(i)}\|_F^2 \quad \text{s.t.} \quad \|\bm{C}^{(i)}\|_{2,0} \le L
\end{equation}
where $\bm{Y}^{(i)}=[\mathbf{y}^{*(i)}_{t-k},\cdots,\mathbf{y}^{*(i)}_{t-1},\mathbf{y}^{(i)}]$, the columns of $\bm{C}^{(i)}$ are the corresponding sparse representations, and $\|\bm{C}^{(i)}\|_{2,0}$ is the $\ell_0$ pseudo-norm computed over the $\ell_2$ norms of rows of $\bm{C}^{(i)}$, i.e. counts the number of nonzero rows of $\bm{C}^{(i)}$ (see Fig. \ref{fig:grouping}).

This optimization problem belongs to the class of NP-Hard problems \cite{Zelnik2012}, and is therefore usually solved approximately, by using a greedy algorithm or convex relaxation. Different convex relaxations have been proposed for this problem which often take the form of:
\begin{equation}\label{equ:l2l1}
\arg\min_{\bm{C}^{(i)}} \frac{1}{2}\|\bm{Y}^{(i)}-\bm{DC}^{(i)}\|_F^2 + \lambda \|\bm{C}^{(i)}\|_{p,q}
\end{equation}
where $\|\bm{C}^{(i)}\|_{p,q}$ is the $\ell_q$ norm computed over the $\ell_p$ norms of rows of $\bm{C}^{(i)}$, and $\lambda$ is a regularization parameter, which balances reconstruction error with model complexity. For (\ref{equ:l2l1}) to become a convex problem, we require $p,q\ge1$. This is the well known convex formulation of the joint sparse recovery \cite{Baraniuk10} or simultaneous sparse approximation \cite{Tropp06} problem, also known as the Multiple Measurement Vector (MMV) \cite{Cotter05} problem in the compressed sensing community. Other convex formulations for this problem follow the general $\ell_q$/$\ell_p$ form with $q\ge1$ and $p=1$, among which $\ell_2$/$\ell_1$ and $\ell_\infty$/$\ell_1$ formulations are more widely used. We adhere to the $\ell_2$/$\ell_1$ form because the objective function of (\ref{equ:l2l1}) is strongly convex for $k>0$, and therefore has a unique solution. Several algorithms exist that efficiently solve this problem \cite{Cotter05,Malioutov05,Lu11,Rakotomamonjy11}. 

A common greedy strategy used to solve (\ref{equ:l2l0}) is the Simultaneous Orthogonal Matching Pursuit (SOMP) algorithm \cite{Tropp06_2} which is an extension of the well known Orthogonal Matching Pursuit (OMP) algorithm. In SOMP, it is initially assumed that $\bm{C}^{(i)}=0$, i.e. no atoms are used in the representation. Then in each iteration, the remainder $\bm{R}=\bm{Y}^{(i)}-\bm{DC}^{(i)}$ is computed, and the dictionary atom $\mathbf{d}_j$ that has the largest correlation with $\bm{R}$, i.e. $\arg\max_j \|\bm{R}^T\mathbf{d}_j\|_2$,  is added to the set of active atoms in the representation. The sparse representation, $\bm{C}^{(i)}$ is then updated to use the set of active atoms.

Among the algorithms proposed to solve (\ref{equ:l2l1}), Malioutov, et al.\cite{Malioutov05} show that the optimization may be posed as Second Order Cone Programming (SOCP) for which off-the-shelf optimizers are available. The inner loops of SOCP are computationally expensive and the algorithm is only suitable for small-size problems \cite{Malioutov05,Lu11}. The work of Lu, et al. \cite{Lu11} extends the Alternating Directions Method (ADM) of \cite{Yang11} to solve the MMV recovery. Although the proposed algorithm is quite fast within an acceptable solution accuracy, there is no guarantee that each iteration of the algorithm will reduce the objective function. We employ the regularized M-FOCUSS algorithm of Cotter et al. \cite{Cotter05} which is simple, efficient, and also guaranteed to reduce the objective function in each iteration. The algorithm works by estimating the $\ell_2$ norm of each row of $\bm{C}^{(i)}$, and then updating $\bm{C}^{(i)}$ based on that estimate. 
The update rule may be written as:
\begin{equation}\label{MFOCUSS_update}
\bm{C}^{(i)} = \Lambda \bm{D}^T (\bm{D}\Lambda \bm{D}^T + \lambda \mathbf{I})^{-1}\bm{Y}^{(i)}
\end{equation}
where $\Lambda = \text{diag}(\|\bm{C}^{(i)}_j\|_2)$ is computed using the previous estimate of $\bm{C}^{(i)}$. The algorithm may be initialized from any random point for which all rows of $\bm{C}^{(i)}$ have a nonzero norm,
and is terminated when the difference between consecutive estimates of $\bm{C}^{(i)}$ is smaller than some threshold. 

\subsection{Observation Model}
The model defined in (\ref{equ:patch-repres-noise}) represents both the target object and any occlusion or noise, through the sparse coefficients. For a given sparse representation $\mathbf{c}^{(i)}$ of patch $i$, $\mathbf{c}^{(i)}_{\Lambda_i}$ are the coefficients corresponding to the patch template $i$, and represent the target, while $\mathbf{c}^{(i)}_{\Lambda_i^c}$ represents any occlusion or noise. In order to measure the likelihood that a given candidate is generated by the target model, we only pay attention to the portion of sparse coefficients that represent the target, and disregard the representation of the occlusion which we later use for occlusion detection in Section \ref{subsec:dict-update}.

Given the current frame image $\mathbf{z}$, and a region defined by the state $\mathbf{x}$ which is provided by the motion model as a candidate for the target, we define the likelihood of the candidate image $\mathbf{y}$ as the probability that it is generated by the appearance model dictionary $\bm{D}$ through the portion of sparse coefficients that represent the target found by minimization (\ref{equ:l2l0}) or (\ref{equ:l2l1}). 
Hence we define:
\begin{equation}\label{equ:image_likelihood}
p(\mathbf{z}|\mathbf{x})  \triangleq  p(\mathbf{y}|\bm{D},\mathbf{c}^{(1)}_{\Lambda_1},\ldots,\mathbf{c}^{(m)}_{\Lambda_m})
\end{equation}
Assuming independence between patches of a candidate, we can evaluate a candidate's likelihood as:
\begin{align}\label{equ:likelihood_decomp}
p(\mathbf{y}|\bm{D},\mathbf{c}^{(1)}_{\Lambda_1},\ldots,\mathbf{c}^{(m)}_{\Lambda_m}) =  \prod_i{p(\mathbf{y}^{(i)}|\bm{D}_{\Lambda_i},\mathbf{c}^{(i)}_{\Lambda_i})}
\end{align}
Considering a zero mean Gaussian error $\mathbf{e}^{(i)} \sim N(0,\sigma^{(i)} \mathbf{I})$ in equation (\ref{equ:patch-repres-noise}), 
the likelihood of each patch is obtained from:
\begin{equation}\label{equ:patch-likelihood}
p(\mathbf{y}^{(i)}|\bm{D}_{\Lambda_i},\mathbf{c}^{(i)}_{\Lambda_i}) = \exp(-\frac{\|\mathbf{y}^{(i)}-\bm{D}_{\Lambda_i}\mathbf{c}^{(i)}_{\Lambda_i}\|_2^2}{\sigma^{(i)^2}})
\end{equation}
Taking the log-likelihood of (\ref{equ:image_likelihood}), using (\ref{equ:likelihood_decomp}) and (\ref{equ:patch-likelihood}), and assuming equal variance for all patches, we have the final measure for selecting the best candidate:
\begin{equation}
\log p(\mathbf{z}|\mathbf{x})\propto-\sum_i{\|\mathbf{y}^{(i)}-\bm{D}_{\Lambda_i}\mathbf{c}^{(i)}_{\Lambda_i}\|_2^2}
\end{equation}
Intuitively, this log-likelihood measure will favor the candidate with the lowest sum of reconstruction errors calculated for each patch separately, via its representation in the corresponding patch template. The main benefit of this new likelihood measure is due to patchwise treatment of the candidates. When a partial or even a severe occlusion occurs, as long as there are some patches which are not totally occluded, this measure will indicate the true best candidate. A pseudocode for the proposed tracker is presented in the Algorithm \ref{alg:tracker}.

We expect to get better results by using some heuristic to select appropriate values for $\sigma^{(i)}$. As the target  can not usually be inscribed inside a rectangle, marginal patches often contain the background. Moreover, occlusion usually occurs at the boundaries of the target, and therefore marginal patches are more affected. To alleviate this undesirable effect we can set the values of $\sigma^{(i)}$s proportional to the inverse distance of the patch center from the center of the rectangle enclosing the target. In the experiments of Section \ref{sec:experiment}, we have used the simple case with equal variance for all patches. 

\begin{figure}
\centering
\includegraphics[width=6.5cm]{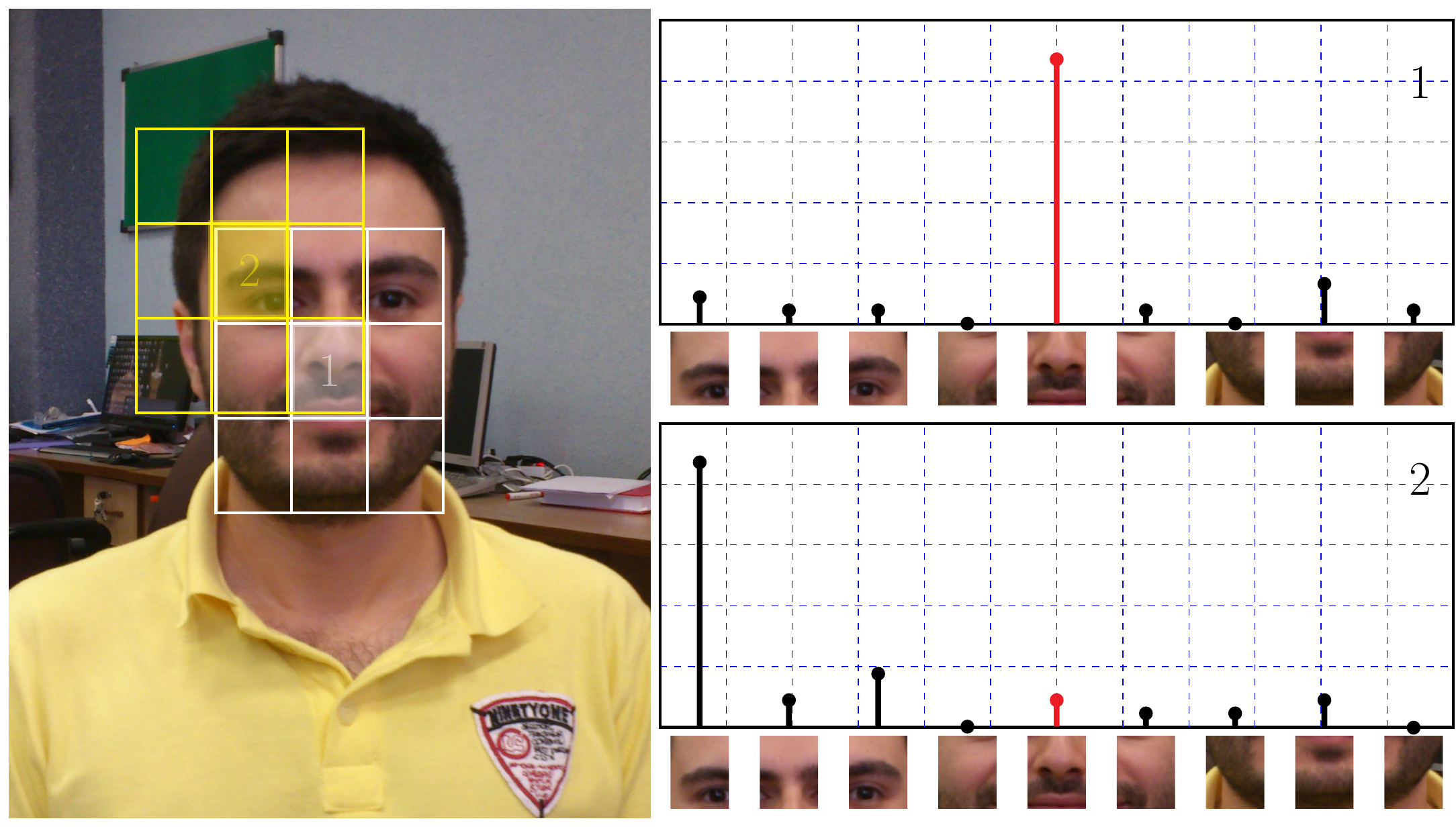}
\caption{\textbf{Misalignment effect.} Illustrative sparse representation of patch 1 (highlighted in white) and patch 2 (highlighted in yellow).
The center patch of the white candidate is aligned with the target and has low reconstruction error in its corresponding patch template. The center patch in the yellow candidate is slightly shifted away from the target and is mainly reconstructed by a non-corresponding patch template, hence increasing its reconstruction error in the corresponding patch template.}
\label{fig:patch-effect}
\end{figure}

Calculating the negative log-likelihood of a candidate as the sum of its patch-reconstruction errors, increases the robustness of the tracker to misalignments. When a candidate is slightly shifted away from the target, the reconstruction error of its patches will substantially increase. As depicted in Fig. \ref{fig:patch-effect}, this is because the shifted candidate is no longer aligned with the target, hence the sparse representation of any of its patches is less likely to use atoms from its corresponding patch template. Since the reconstruction error is computed using the coefficients of the corresponding patch template, the log-likelihood of a misaligned candidate will be quite small.

Since the target's appearance often changes slowly, obtaining the sparse representation of a candidate patch, jointly with its corresponding patches from the best candidates of previous frames, increases the robustness of the tracker to ``false'' candidates. Grouping a ``true'' candidate patch with corresponding patches from the best previous candidates is expected to increase its chance of being represented by atoms from the correct patch template even in case of a moderate pose change, hence increasing its likelihood. Our intuition is that the joint sparse nature of the solution of minimization (\ref{equ:l2l1}) is likely to tolerate an increase in the reconstruction error of the ``true'' patch, in order to best represent the whole group of patches.

\begin{figure*}
\centering
\includegraphics[width=0.78\textwidth]{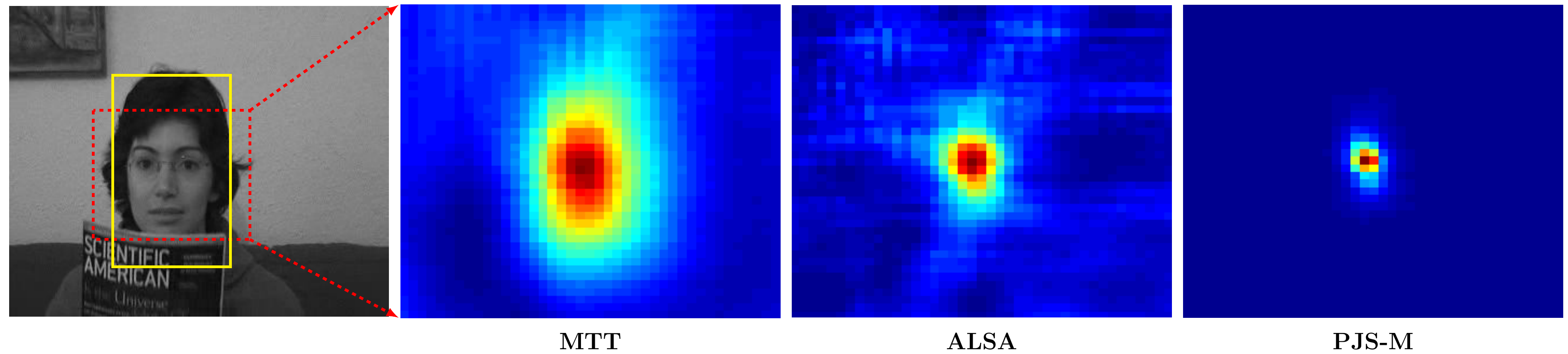}
\caption{\textbf{Likelihood measure.} Comparison of likelihood maps obtained by MTT \cite{Zhang2013a}, ASLA \cite{Jia}, and proposed tracker. These maps show the likelihood of candidates generated by sweeping the ground truth bounding box (yellow rectangle) inside the red rectangular region.}
\label{fig:likelihood-var}
\end{figure*}

To illustrate the above discussions, we have depicted in Fig. \ref{fig:likelihood-var}, the likelihood measure for a set of regularly sampled candidates centered inside a rectangular region around the target, for various methods. As expected, the likelihood peaks at the center of the region, but more steeply for the proposed method. In other words, our likelihood measure has less variance compared to the other methods.

\begin{algorithm}[h!]
\caption{Proposed tracker} 
\label{alg:tracker}
{\small 
\begin{algorithmic}[1]
\REQUIRE 
Previous particles $\mathcal{X}_{t-1} =\{\mathbf{x}_{t-1}^i\}_{i=1}^{N}$, frame $\mathbf{z}_t$, dictionary $\bm{D}$, previous $k$ targets $\{\mathbf{y}^*_{t-1},\ldots,\mathbf{y}^*_{t-k}\}$.

\ENSURE
Target $\mathbf{y}$, current particles $\mathcal{X}_{t} =\{\mathbf{x}_{t}^i\}_{i=1}^{N}$, dictionary $\bm{D}$.

\STATE{
Draw particles $\mathcal{X}_{t}$ from $\mathcal{X}_{t-1}$ using $\mathbf{x}_{t}^i \sim \mathcal{N}(\mathbf{x}_{t-1}^i,\Sigma)$.}

\FOR{$i=1$ to $N$}

\STATE{
Generate candidate $\mathbf{y}$ corresponding to particle $\mathbf{x}_t^i$ and resize it to a predefined size.
}

\STATE{
Partition candidate $\mathbf{y}$ in to equal-sized non-overlapping patches $\{\mathbf{y}^{(1)},\mathbf{y}^{(2)},\ldots,\mathbf{y}^{(m)}\}$.}


\FOR{$j=1$ to $m$}

\STATE{$\bm{Y}^{(j)} \leftarrow [\mathbf{y}^{*(j)}_{t-k},\cdots,\mathbf{y}^{*(j)}_{t-1},\mathbf{y}^{(j)}]$}


\STATE {$\bm{C}^{(j)} \leftarrow$ solution of minimization (\ref{equ:l2l0}) or (\ref{equ:l2l1})}

\STATE{$\mathbf{c}^{(j)} \leftarrow \bm{C}^{(j)}_{k+1}$}

\STATE{$p(\mathbf{y}^{(j)}|\bm{D}_{\Lambda_i},\mathbf{c}^{(i)}_{\Lambda_i}) \leftarrow \exp(-\|\mathbf{y}^{(j)}-\bm{D}_{\Lambda_j}\mathbf{c}^{(j)}_{\Lambda_j}\|_2^2)$}

\ENDFOR
\STATE{$\log p(\mathbf{z}_t|\mathbf{x}_t^i)   \leftarrow  \sum_{j} \log p(\mathbf{y}^{(j)}|\bm{D}_{\Lambda_j},\mathbf{c}^{(j)}_{\Lambda_j})$}


\ENDFOR

\STATE{
Select the candidate with the largest $\log p(\mathbf{z}_t|\mathbf{x}_t^i)$ as the target $\mathbf{y}$ in current frame.}

\STATE{
Update dictionary $\bm{D}$ with $\mathbf{y}$ using Algorithm \ref{alg:dic-update}.}
\end{algorithmic}
}
\end{algorithm}

\subsection{Dictionary Update} \label{subsec:dict-update}
The appearance model needs to be updated over time in order to handle variations in the target's pose and illumination. Commonly, the best candidate found in the current frame is used to update the appearance model. Many schemes have been presented to update generative and discriminative trackers. In most generative sparse trackers, the target is modeled with a dictionary and is updated by replacing its atoms. Two issues are important when updating the dictionary. The first issue is the policy used to select the dictionary atom(s) that are going to be replaced, while the second issue is related to detecting and perhaps removing any occlusion present in the new atom(s). If severely occluded atoms are allowed to enter, and hence corrupt the dictionary, the tracker is prone to produce erroneous results leading to further corruption of the dictionary and gradual failure of the tracker. This is known as the problem of drift. Therefore, an accurate occlusion detection approach is essential to alleviate this problem.

To select an old atom to be replaced with a newly found target, the policy used by \cite{Mei2011,Bao2012,Mei2011a,Zhang,Zhang2013a} is based on assigning weights to the dictionary atoms. The weight assigned to each atom is proportional to its corresponding coefficient in the sparse representation of the best candidate. The weights are updated in each frame, after the best candidate is found. The atom with the lowest weight is selected to be replaced by the target. To alleviate the drift problem, similar to \cite{Jia}, we randomly select dictionary atoms according to a predefined distribution. This distribution is defined so that recent atoms are more likely to be removed.

Based on our patchwise approach, we propose a new occlusion detection scheme. We investigate whether each patch is occluded or not by defining an occlusion probability. A patch is assumed to be occluded if its occlusion probability is larger than $\frac{1}{2}$. Then the occluded patches are excluded from the best candidates, when updating the dictionary. The occlusion probability of a patch $\mathbf{y}^{(i)}$, is defined as:
\begin{align}\label{equ:occ_prob}
& p({o}^{(i)}|\bm{D},\mathcal{O}^{(i)}, \mathbf{y}^{(i)},\mathbf{c}^{(i)}) \propto \\ & \quad p(\mathbf{y}^{(i)}|\bm{D},\mathcal{O}^{(i)},{o}^{(i)},\mathbf{c}^{(i)}) p({o}^{(i)}|\bm{D},\mathcal{O}^{(i)},\mathbf{c}^{(i)}) \nonumber
\end{align}
where $o^{(i)}=1$, and $o^{(i)}=0$ indicate the occluded and  non-occluded states, respectively, and $\mathcal{O}^{(i)}=\{ {o}^{(i)}_1,\ldots,{o}^{(i)}_{t-1} \}$ is the history of occlusion states for patch $i$. Assuming that given the current occlusion state of $\mathbf{y}^{(i)}$, this patch is independent of previous occlusion occurrences, and considering a first order Markov model for occlusion state of patches, we have:
\begin{align}
p(\mathbf{y}^{(i)}|\bm{D},\mathcal{O}^{(i)},{o}^{(i)},\mathbf{c}^{(i)}) &= p(\mathbf{y}^{(i)}|\bm{D},{o}^{(i)},\mathbf{c}^{(i)})  \label{equ:occ-part1}\\
p({o}^{(i)}|\bm{D},\mathcal{O}^{(i)},\mathbf{c}^{(i)}) &=p({o}^{(i)}|{o}^{(i)}_{t-1}) \label{equ:occ-part2}
\end{align}
To determine the probability of occlusion in (\ref{equ:occ_prob}), it suffices to define the above two distributions. Considering the model in (\ref{equ:patch-repres-noise}), the target patch is represented by the sparse coefficients $\mathbf{c}^{(i)}_{\Lambda_i}$, while severely occluded   patches are expected to be represented mainly by $\mathbf{c}^{(i)}_{\Lambda_i^c}$. Since the best candidate is already found, we no longer need to impose the temporal similarity assumption, and $\mathbf{c}_{\Lambda_i}$ is obtained from (\ref{equ:l2l1}) with $k=0$. 
Hence, to discriminate an occluded patch form a non-occluded patch we define the likelihood of patch $i$ in (\ref{equ:occ-part1})  for the case of  $o^{(i)} = 0$ and $o^{(i)} = 1$, respectively:
\begin{align}
p(\mathbf{y}^{(i)}\big|\bm{D},\mathbf{c}^{(i)},{o}^{(i)}=0) &= \exp(-\|\mathbf{y}^{(i)}-\bm{D}_{\Lambda_i} \mathbf{c}^{(i)}_{\Lambda_i}\|_2^2) 
\label{equ:occ_likelihood_1} \\
p(\mathbf{y}^{(i)}|\bm{D},\mathbf{c}^{(i)},{o}^{(i)}=1) &= \exp(-\|\mathbf{y}^{(i)}-\bm{D}_{\Lambda_i^c} \mathbf{c}_{\Lambda_i^c}^{(i)}\|_2^2)
\label{equ:occ_likelihood_2}
\end{align}
\begin{figure}
\centering
\includegraphics[width=4.5cm]{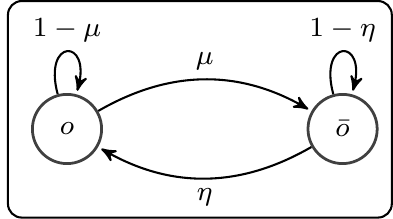}
\caption{\textbf{Markov chain of occlusion state.} Two-state Markov chain with adaptive transition probabilities used to define occlusion prior of each patch. States $o$ and $\bar{o}$ represent occluded and non-occluded states, respectively.}
\label{fig:markov}
\end{figure}
To define a prior for the occlusion state of patch $i$ in (\ref{equ:occ-part2}), we consider a simple two state Markov chain for each patch, as shown in Fig \ref{fig:markov}. The transition probabilities of this model are updated online during tracking, by using Maximum A Posteriori (MAP) estimation. Removing the patch superscript $(i)$ from occlusion states $o^{(i)}$ and  restoring the frame subscript for notational convenience, we have:
\begin{align}\label{equ:occ-prior}
p(o_t|o_{t-1}) = & \, \mu^{o_{t-1}(1-o_t)} (1-\mu)^{o_{t-1} o_t} \\ 
& \, \eta^{(1-o_{t-1})o_t} (1-\eta)^{(1-o_{t-1})(1-o_{t})} \nonumber
\end{align}
Since the transition probabilities are between 0 and 1, an appropriate prior for these parameters is the beta distribution, namely $\mu \sim \beta(a,b)$, and $\eta \sim \beta(c,d)$. By changing the parameters of these distributions we can fully reflect any prior knowledge about the transition probabilities that determine patch occlusion. With knowledge of occlusion states in previous frames and prior distribution of the transition probabilities, $\mu,\eta$ can be found using MAP estimation as follows:
\begin{align}
& \arg\max_{\mu,\eta} \, p(\mu,\eta | o_1,o_2,\cdots,o_t) =   \\
& \qquad  \arg\max_{\mu,\eta} \, \beta(\mu|a,b) \beta(\eta|c,d) \prod_{i=1}^{t} p(o_i|o_{i-1},\mu,\eta)  \nonumber 
\end{align}
Substituting (\ref{equ:occ-prior}) and solving this minimization, the MAP estimation of $\mu$ and $\eta$ can be found as:
\begin{align}
\hat{\mu}_{_\text{MAP}}  &= \frac{a-1+n_{o\bar{o}}}{a-1+n_{o\bar{o}}+b-1+n_{oo}}  \label{equ:mu}\\
\hat{\eta}_{_\text{MAP}} &= \frac{c-1+n_{\bar{o}o}}{c-1+n_{\bar{o}o}+d-1+n_{\bar{o}\bar{o}}} 
\label{equ:eta}
\end{align}
where for example $n_{o\bar{o}}$ counts the number of transitions between occlusion state $o_{i-1}=1$, and non-occluded state $o_{i}=0$ in the occlusion history $\mathcal{O}$. A pseudo-code of the dictionary update procedure is represented in Algorithm \ref{alg:dic-update}.

\begin{algorithm}[t!]
\caption{Dictionary update} 
\label{alg:dic-update}
{\small 
\begin{algorithmic}[1]
\REQUIRE 
Dictionary $\bm{D}$, newly found target image $\mathbf{y}$, 
state transition counters, occlusion history of patches $\{\mathcal{O}^{(i)}\}_{i=1}^m$.

\ENSURE
Updated dictionary $\bm{D}$,
updated state transition counters, updated occlusion history of patches $\{\mathcal{O}^{(i)}\}_{i=1}^m$.

\STATE{Randomly select an old dictionary target image $\mathbf{y}_{\text{old}}$, with higher probability in selection of recent targets.}
\FOR{$i=1$ to $m$}

\STATE{Find sparse representation of the target patch: \\
$\mathbf{c}^{(i)} \leftarrow $
$\arg\min_{\mathbf{c}} \frac{1}{2}\|\mathbf{y}^{(i)}-\bm{D}\mathbf{c}\|_2^2 + \lambda \|\mathbf{c}\|_1$
}
\STATE{Find occlusion likelihood of the target patch according to (\ref{equ:occ_likelihood_1}) and (\ref{equ:occ_likelihood_2}).
}

\STATE{Update the transition probabilities of Markov chain according to (\ref{equ:mu}) and (\ref{equ:eta}).
}

\STATE{Evaluate the prior probability of occlusion according to (\ref{equ:occ-prior}).
}

\STATE{Calculate occlusion probability according to (\ref{equ:occ_prob}).
}

\STATE{Upade dictrionary with non-occluded patches of the target:}
\IF{$p({o}^{(i)}=0|\bm{D},\mathcal{O}^{(i)}, \mathbf{y}^{(i)},\mathbf{c}^{(i)}) \geq 0.5$}
\STATE{$\mathbf{y}_{\text{old}}^{(i)} \leftarrow \mathbf{y}^{(i)}, \: o^{(i)} \leftarrow 0$}
\ELSE
 \STATE{$o^{(i)} \leftarrow 1$}
\ENDIF
\STATE{Update counters using occlusion states $o^{(i)}$ and $o_{t-1}^{(i)}$.}
\STATE{
$\mathcal{O}^{(i)} \leftarrow \mathcal{O}^{(i)} \cup \{o^{(i)}\}$
}
\ENDFOR

\end{algorithmic}
}
\end{algorithm}

\section{Experiments}\label{sec:experiment}
In this section, we present qualitative and quantitative experimental results to evaluate the performance of the proposed trackers.
Qualitative comparisons involve demonstrating each tracker's performance on a number of sample frames.
Since this measure is limited to observing performance for a few hand picked frames, it is essential to evaluate different tracking algorithms with quantitative measures such as Center Location Error (CLE), PASCAL overlap score (VOC), and success rate. 
The experimental results are organized as follows. In Section \ref{subs:base-tr&ds} an overview of the different datasets used for experiments, and the trackers used for comparison, is provided. Some important implementation details are discussed in Section \ref{subs:implem-detail}. The comparison of different trackers according to qualitative and quantitative measures are presented in Sections \ref{subsec:qualitative} and \ref{subsec:quantitative}, respectively. 
 
\subsection{Base Trackers and Dataset}\label{subs:base-tr&ds}
\begin{figure}[!t]
\centering
\includegraphics[width=0.46\textwidth]{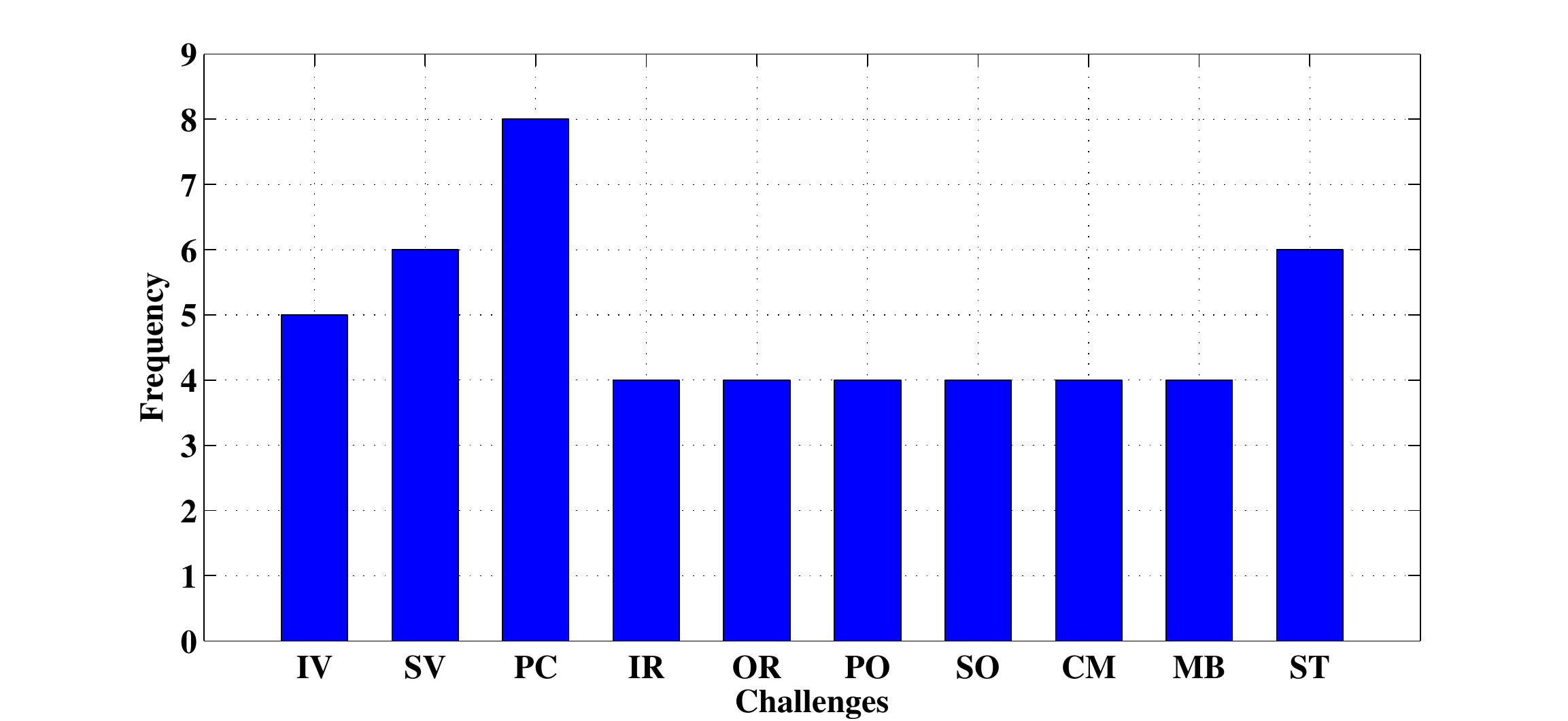}
\caption{\textbf{Challenge histogram.} Frequency of challenges listed in Table \ref{tbl:ds-attr}.}
\label{fig:attr-dist}
\end{figure}

For evaluation we have selected a set of 10 publicly available datasets that cover the common challenges prevalent in visual tracking. The datasets are commonly known as \textit{board}, \textit{crossing}, \textit{david}, \textit{dollar}, \textit{faceocc2}, \textit{skating1}, \textit{stone}, \textit{sylv}, \textit{trellis}, and \textit{walking}, that are available along with their ground truth at \cite{Wu2013}.
A list of the challenges posed by these datasets is provided in Table \ref{tbl:ds-attr}. We have annotated the datasets according to these challenges as listed in Table \ref{tbl:ds-desc},
followed by their different attributes, namely, target size, resolution, and number of frames.  A histogram that shows the number of datasets that pose each challenge is shown in Fig. \ref{fig:attr-dist}. This histogram approximately has an uniform distribution, indicating the diversity of the selected datasets.


We compare our results with four well known trackers, namely OAB \cite{Grabner2006}, MIL \cite{Babenko2011}, Frag \cite{Adam2006}, and IVT \cite{Ross2008}. We also report results of two recent sparse trackers, denoted as APG \cite{Bao2012}, and MTT \cite{Zhang}. For these trackers, we have used the source codes either available from the corresponding author's website or provided by the tracking benchmark \cite{Wu2013}. The proposed trackers are denoted by PJS-M\footnote{Patchwise Joint Sparse-M-FOCCUS} and PJS-S\footnote{Patchwise Joint Sparse-SOMP}. The two trackers differ in that PJS-M solves the convex joint sparse minimization of (\ref{equ:l2l1}), while PJS-S solves the non-convex formulation of (\ref{equ:l2l0}). We use the implementation of SOMP available by the SPAMS \cite{Mairal2009,Mairal2010} package to solve (\ref{equ:l2l0}), and our implementation of regularized M-FOCCUS for (\ref{equ:l2l1}). All source codes for our experiments are available at \url{http://ssp.dml.ir/research/pjs}.

\begin{table}[t]
\centering
\caption{List of challenges used as attributes for dataset annotation in Table \ref{tbl:ds-desc}.}
\includegraphics[width=0.4\textwidth]{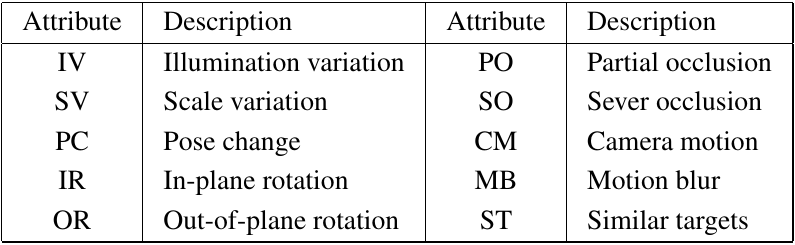}
\label{tbl:ds-attr}
\end{table}
\begin{table}[t]
\centering
\caption{Dataset characteristics: resolution, length, target size, and challenges listed in Table \ref{tbl:ds-attr}}
\includegraphics[width=0.48\textwidth]{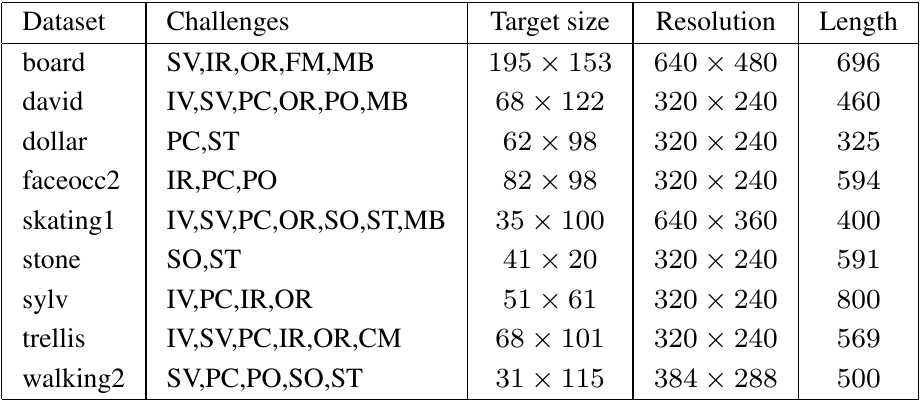}
\label{tbl:ds-desc}
\end{table}


\subsection{Implementation Details}\label{subs:implem-detail}
For methods that use pixel intensities as features (IVT, APG, MTT, PJS-M, and PJS-S) the target object was resized to $32\times 32$ pixels. For OAB, MIL, and Frag that extract features from raw images, the parameters were set as provided by the authors. The patch size was set to $8\times 8$ pixels, resulting in 16 non-overlapping patches for our method. Hence, the dictionary is composed of 16 patch templates each of which contains 10 vectorized grayscale patches. The dictionary is initialized by patches from the given target object and nine slightly (1 to 2 pixels) shifted versions. The nine shifted versions are replaced by the found target in the next nine frames, after which the dictionary is updated according to the scheme described in Section \ref{subsec:dict-update}. The dictionary update parameters of (\ref{equ:mu}) and (\ref{equ:eta}) were set equal to $a=d=4$, and $b=c=8$. For the joint sparse recovery minimizations of (\ref{equ:l2l0}) and  (\ref{equ:l2l1}) we use $L=4$, $\gamma=0.001$, and $k=4$. For methods that require a particle filter motion model, the number of particles was set to 600 and their variance was set to $\sigma = diag([6,6,0.02,0.002,0.002,0])$. The first two diagonal elements of  $\sigma$ correspond to spatial variance in horizontal and vertical directions, and the others correspond to rotation, scale, and skew. These parameters were fixed across all datasets and trackers.

Motion models, especially those which are based on a particle filter, usually involve randomness. To fairly evaluate the trackers and reduce the effects of randomness, we run each experiment 10 times and report the average results. In order to provide reproducible results, the random seed for the 10 runs was set between $0$ to $9$ in MATLAB.
\begin{figure*}[t]
\centering
\includegraphics[width=0.49\textwidth]{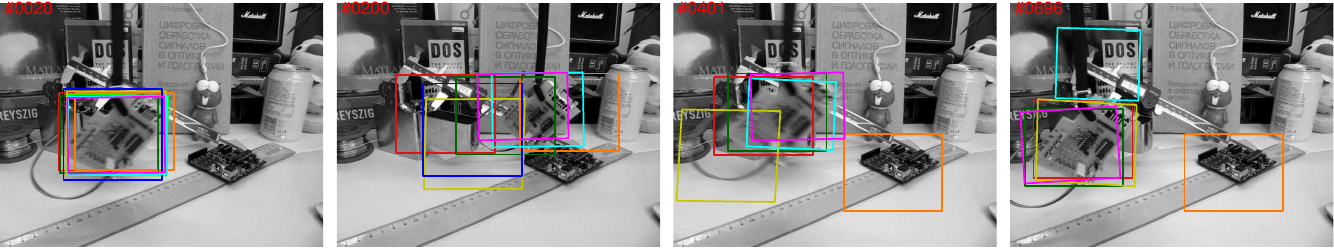}\hspace{2mm}
\includegraphics[width=0.49\textwidth]{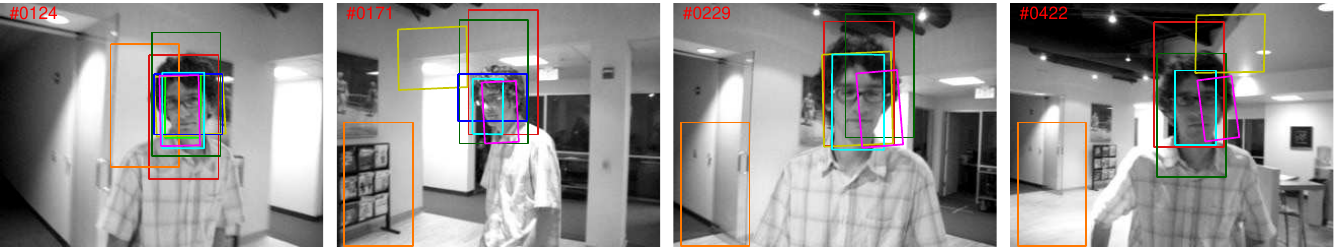}
\includegraphics[width=0.49\textwidth]{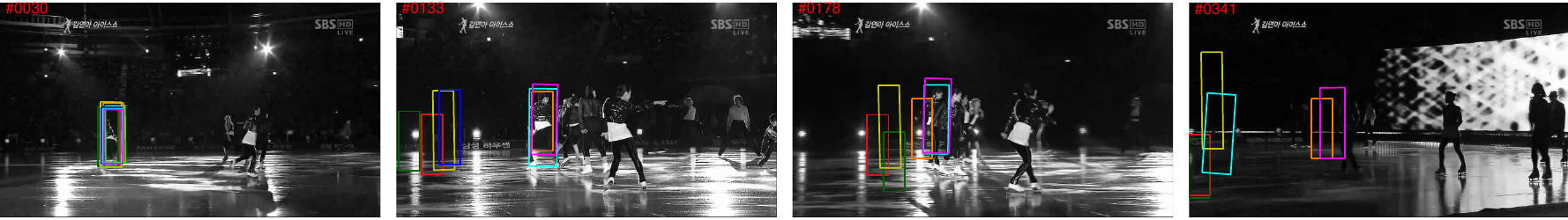}\hspace{2mm}
\includegraphics[width=0.49\textwidth]{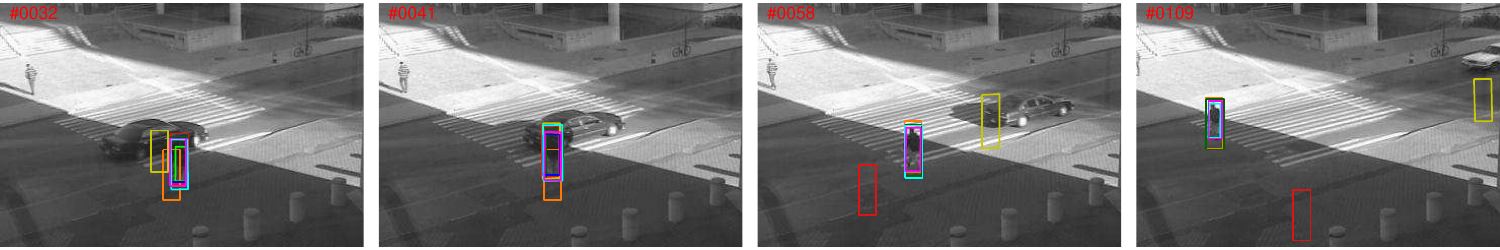}
\includegraphics[width=0.49\textwidth]{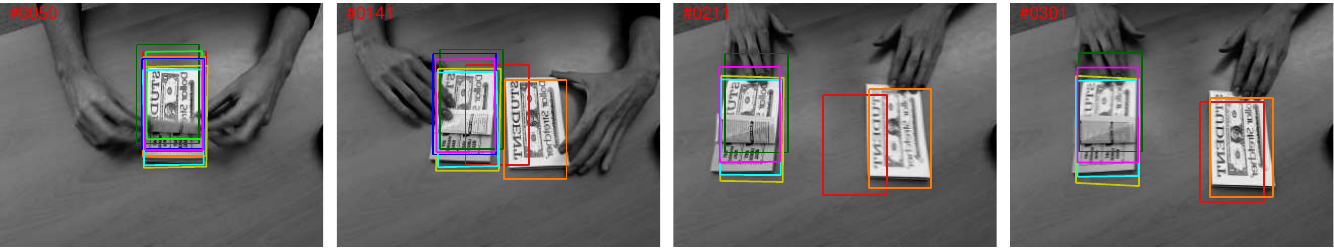}\hspace{2mm}
\includegraphics[width=0.49\textwidth]{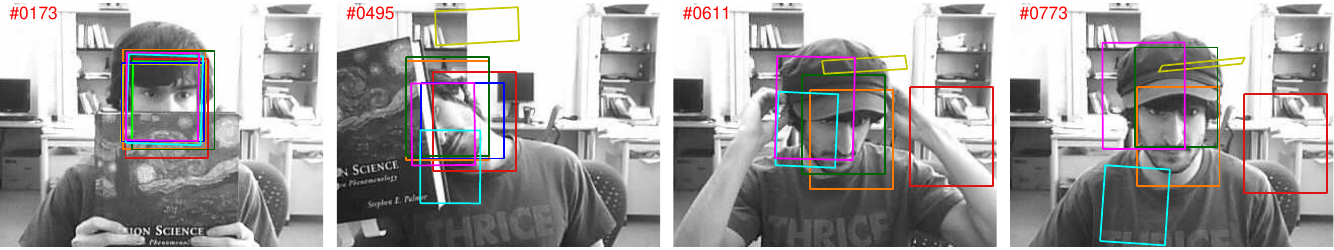}
\includegraphics[width=0.49\textwidth]{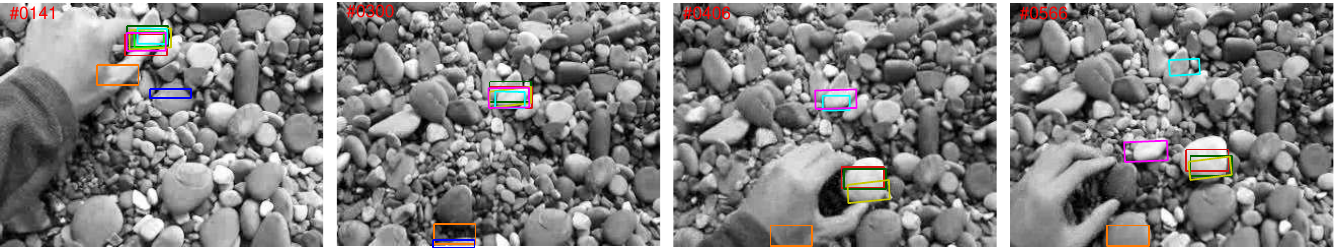}\hspace{2mm}
\includegraphics[width=0.49\textwidth]{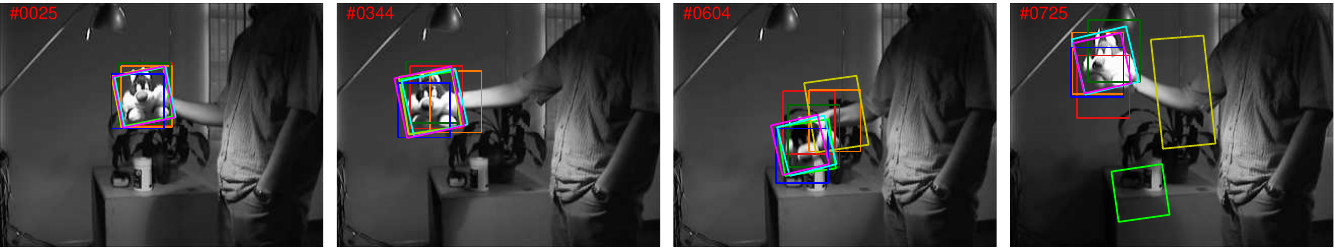}
\includegraphics[width=0.49\textwidth]{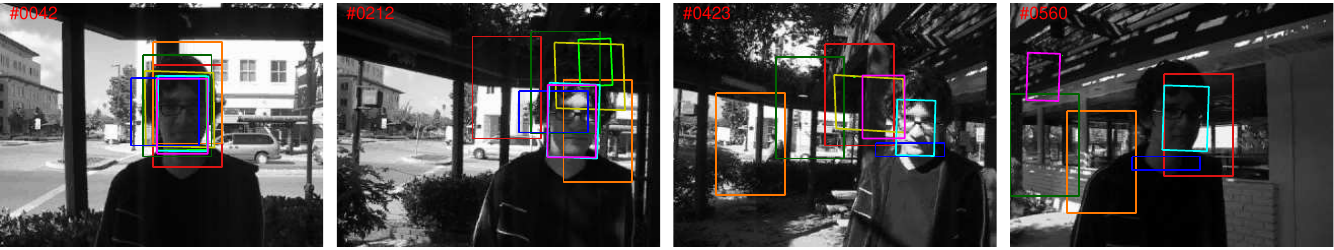}\hspace{2mm}
\includegraphics[width=0.49\textwidth]{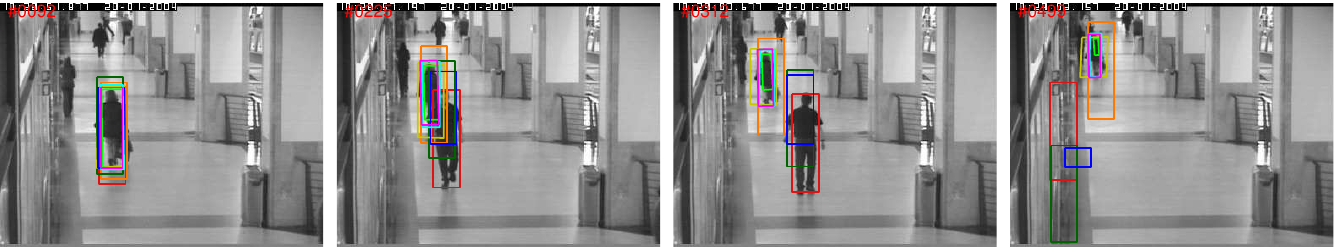}
\vspace{-1mm}
\includegraphics[width=0.6\textwidth]{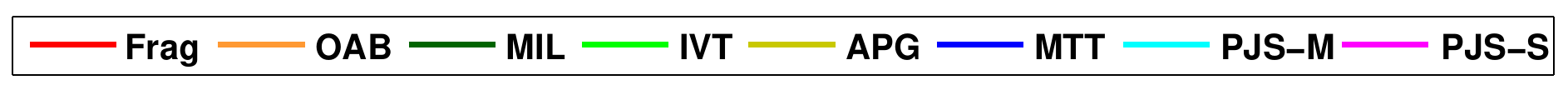}
\vspace{-3mm}
\caption{\textbf{Sample frame illustration}. Tracking results of different trackers on 10 datasets: board, david, skating1, crossing, dollar, faceocc2, stone, sylv, trellis, and walking2, from left to right and top to bottom respectively.}
\label{fig:sampleFrame2}
\end{figure*}
\subsection{Qualitative Comparison}\label{subsec:qualitative}
Figure \ref{fig:sampleFrame2} shows sample frames and the bounding boxes found by each tracker in each dataset. The illustrated results  show that the proposed trackers have been successful in challenging scenarios where previous trackers have often failed. To better reflect on the robustness of the trackers, the next subsection is dedicated to quantitative comparisons. In this section we qualitatively describe how the trackers have dealt with some of the challenges posed by the selected datasets.

\subsubsection{\textbf{board}} In this video, an electronic board undergoes different visual changes such as scale variation, motion blur, and in-plane and out-of-plane rotation. This is quite a challenging dataset, and all trackers except PJS-S and PJS-M have lost track of the target prior to frame $100$. In frame $400$, the target passes its previous location, at which point Frag and MIL were able to track the target again, but lost it again after $60$ or so frames. Only PJS-S was able to track the target throughout the sequence.

\subsubsection{\textbf{david}} This dataset shows a young man's face as he and the camera move, resulting in challenges such as pose change, illumination variation, partial occlusion, and out-of-plane rotation. Prior to frame 124, which has illumination variation and minor pose change, only OAB drifted from the target. In subsequent frames, before the target rotates about $90^{\circ}$, APG, Frag, and MIL also lost the target. After the target returned to its initial pose, all trackers except PJS-S drifted.

\subsubsection{\textbf{skating1}} In this video sequence, the target object and camera move, and the target undergoes sever illumination change, pose variation, and occlusion by a similar object. Around frame 73, the target goes through rapid pose change, as a result only OAB, PJS-M, and PJS-S continued to track the target. Finally, after a sever illumination change, only PJS-S tracked the target until the sequence ended.

\subsubsection{\textbf{crossing}} In this sequence a man (target) crosses a street and a car passes by, momentarily changing the target's background. Almost all trackers were able to follow the target up to frame $41$, when the car appears. After this, due to low resolution and decreased discrimination between the car and the target, Frag, IVT, APG, and MTT lost the target. A close look at the frames shows that PJS-M and PJS-S produce better fitting bounding boxes and are more robust against scale variation and low resolution.

\subsubsection{\textbf{dollar}} The main challenge in this sequence is the existence of an object similar to the target, which is a ticket pass. The ticket is folded in frame 50, and then a similar unfolded ticket is introduced. All trackers except Frag and OAB were able to follow the ticket till the end of sequence.


\subsubsection{\textbf{faceocc2}} In this video,  the target undergoes pose change, occlusion, and rotation. Before frame $336$, the target is not severely occluded and all trackers except the two proposed trackers exhibited a minor drift. In subsequent frames, this drift increases for all the competing methods, except for the two proposed trackers. For IVT, although the center location error is low, the overlap ratio decreased drastically. At the end, APG, IVT, PJS-S, and PJS-M had less drift compared to the other trackers.


\subsubsection{\textbf{stone}} The main challenge here is the existence of many objects that are similar to the target (which may also occlude the target). In frame 385, a similar object occludes the target, at which point Farg, MIL, and APG lost the target. In frame 510, the target passes near another similar object, where only PJS-S was able to overcome this challenge.

\subsubsection{\textbf{sylv}} In this sequence, the camera is fixed and a stuffed cat (target) moves in front of a light source, resulting in illumination change, and in and out-of-plane rotation. During the final frames, and after large out-of-plane rotations, only PJS-S and PJS-M were able to track the target with the largest overlap.

\subsubsection{\textbf{trellis}} This dataset is similar to \textit{david} but exhibits larger pose change and sever illumination variations. After frame 212, a large change in illumination caused all trackers except PJS-S and PJS-M to drift or lose the target.

\subsubsection{\textbf{walking2}} The main challenge of this sequence is sever occlusion by a similar object. In frame 200, after occlusion by the similar object, Frag, MIL, and MTT followed the occlusion, and APG and OAB drifted. This drift continued for the remainder of the sequence, and only PJS-S and PJS-M succeed to track the target with accurate alignment. This sequence also shows the ability of our methods to handle large scale changes.
\begin{figure*}[!t]
\centering
\includegraphics[width=0.194\textwidth]{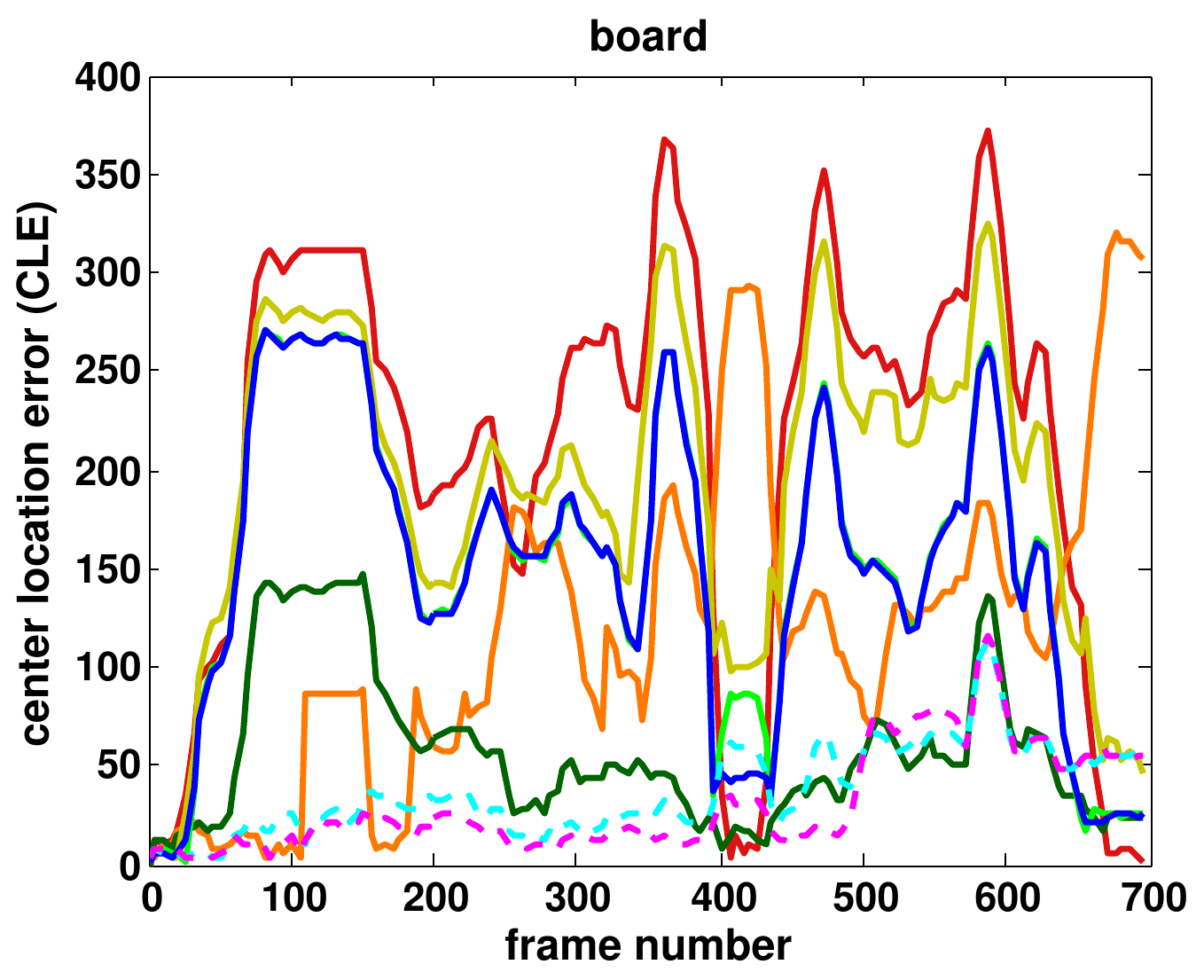}
\includegraphics[width=0.194\textwidth]{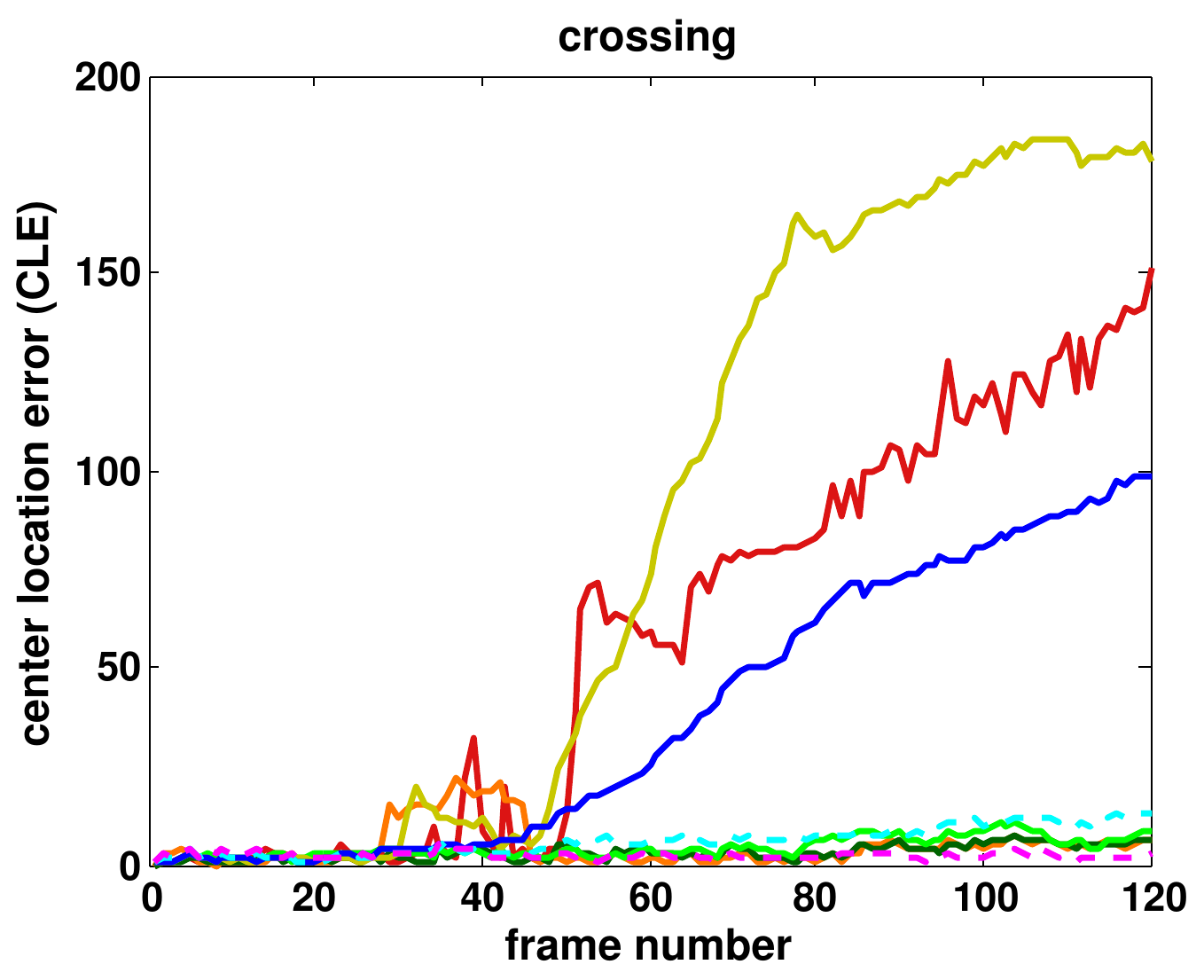}
\includegraphics[width=0.194\textwidth]{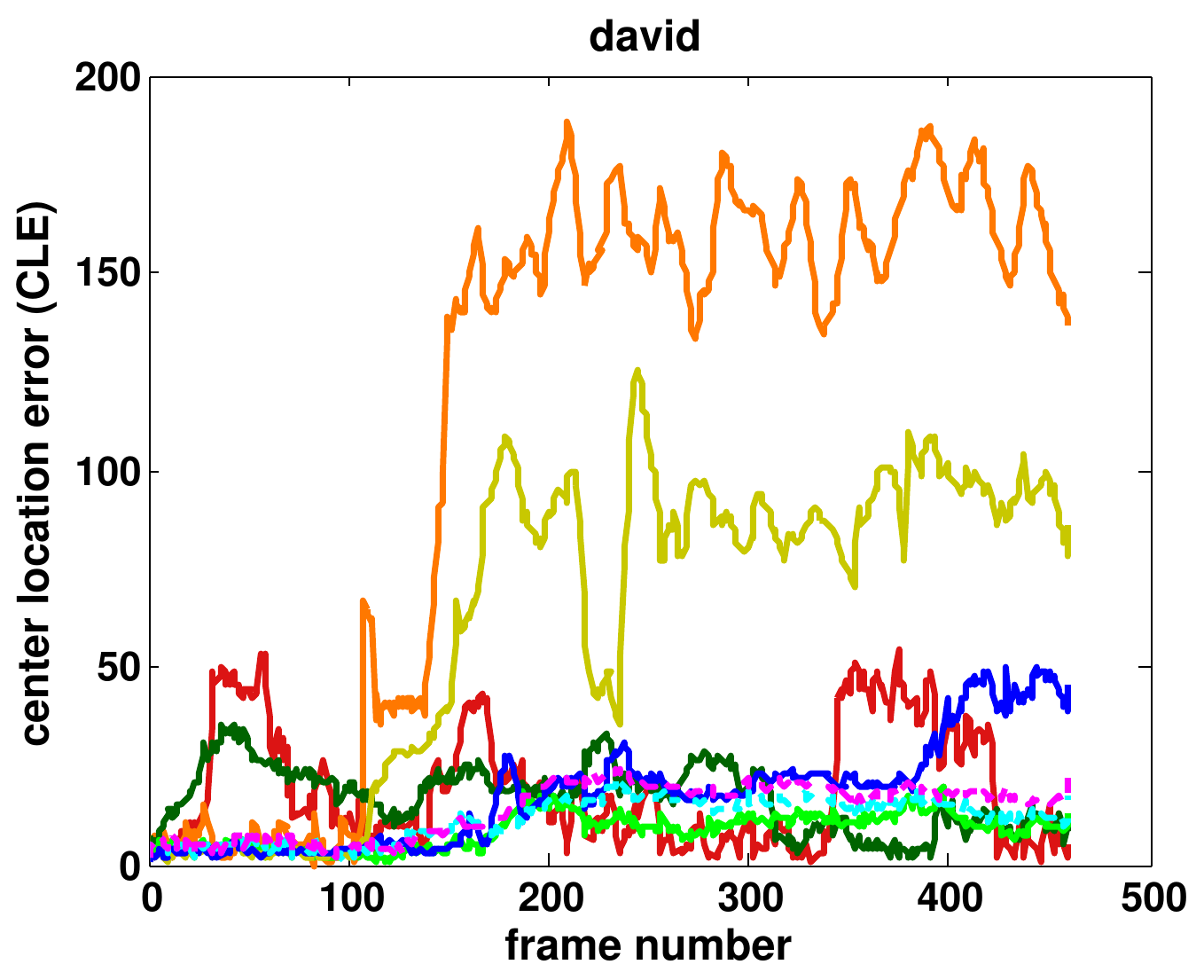}
\includegraphics[width=0.194\textwidth]{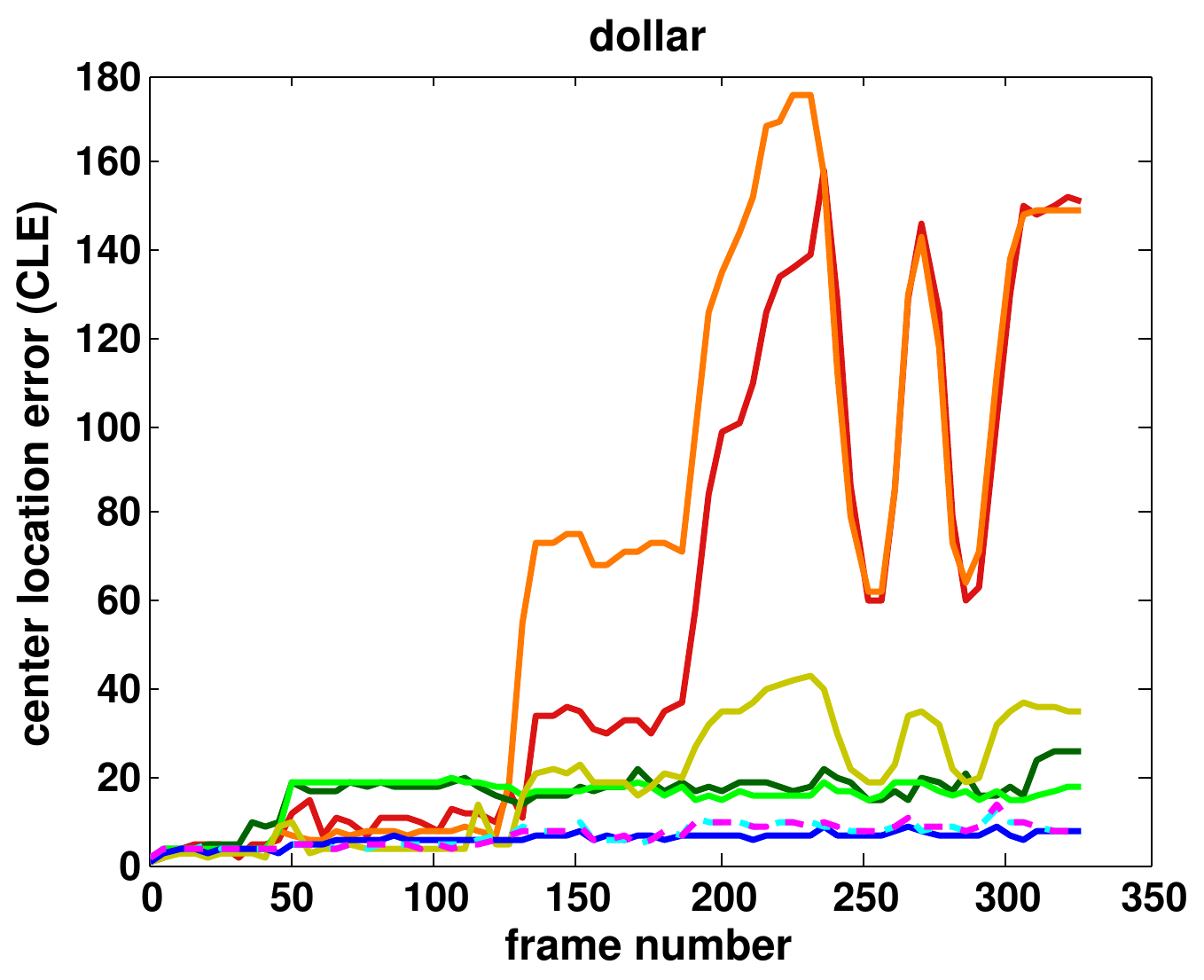}
\includegraphics[width=0.194\textwidth]{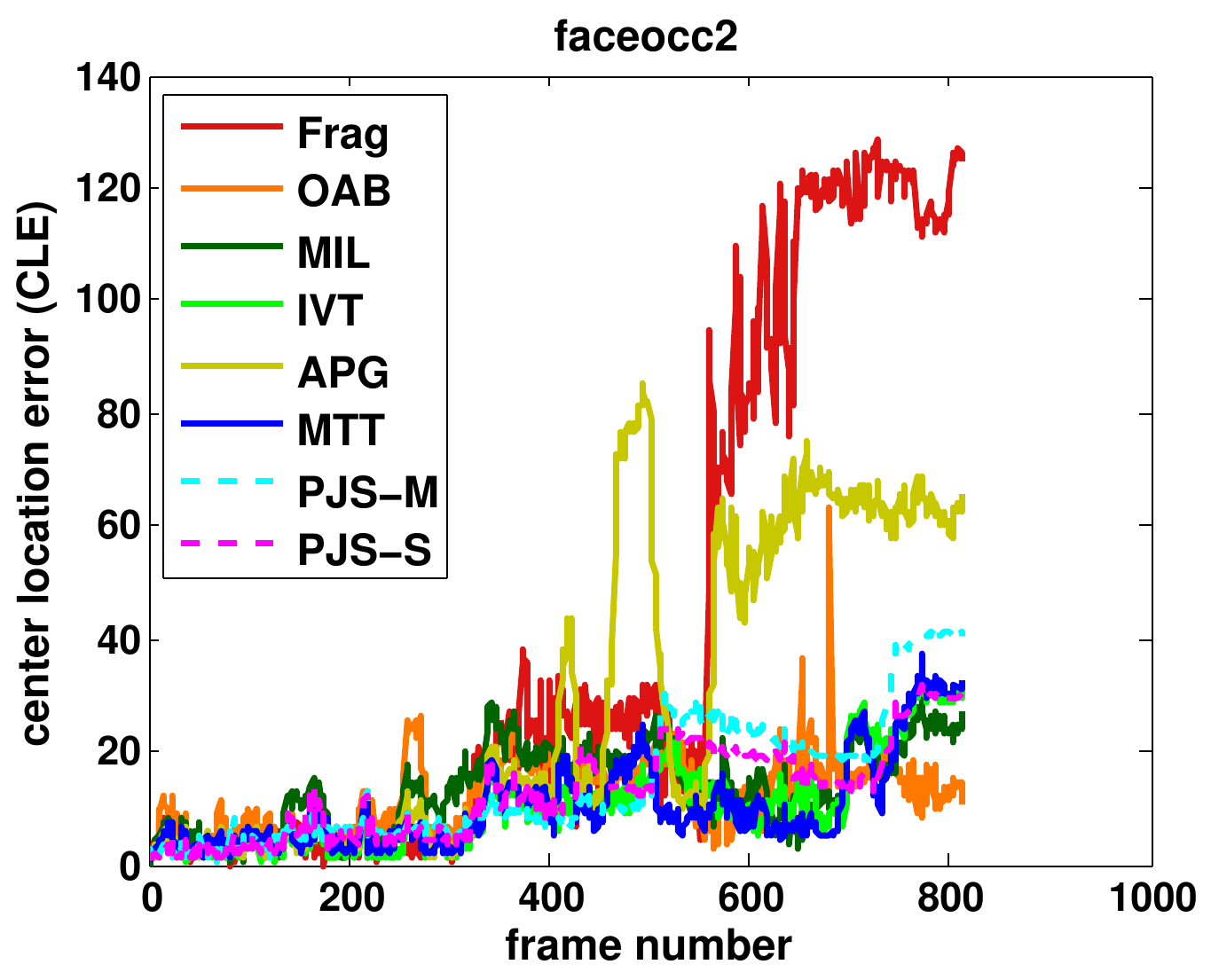}
\includegraphics[width=0.194\textwidth]{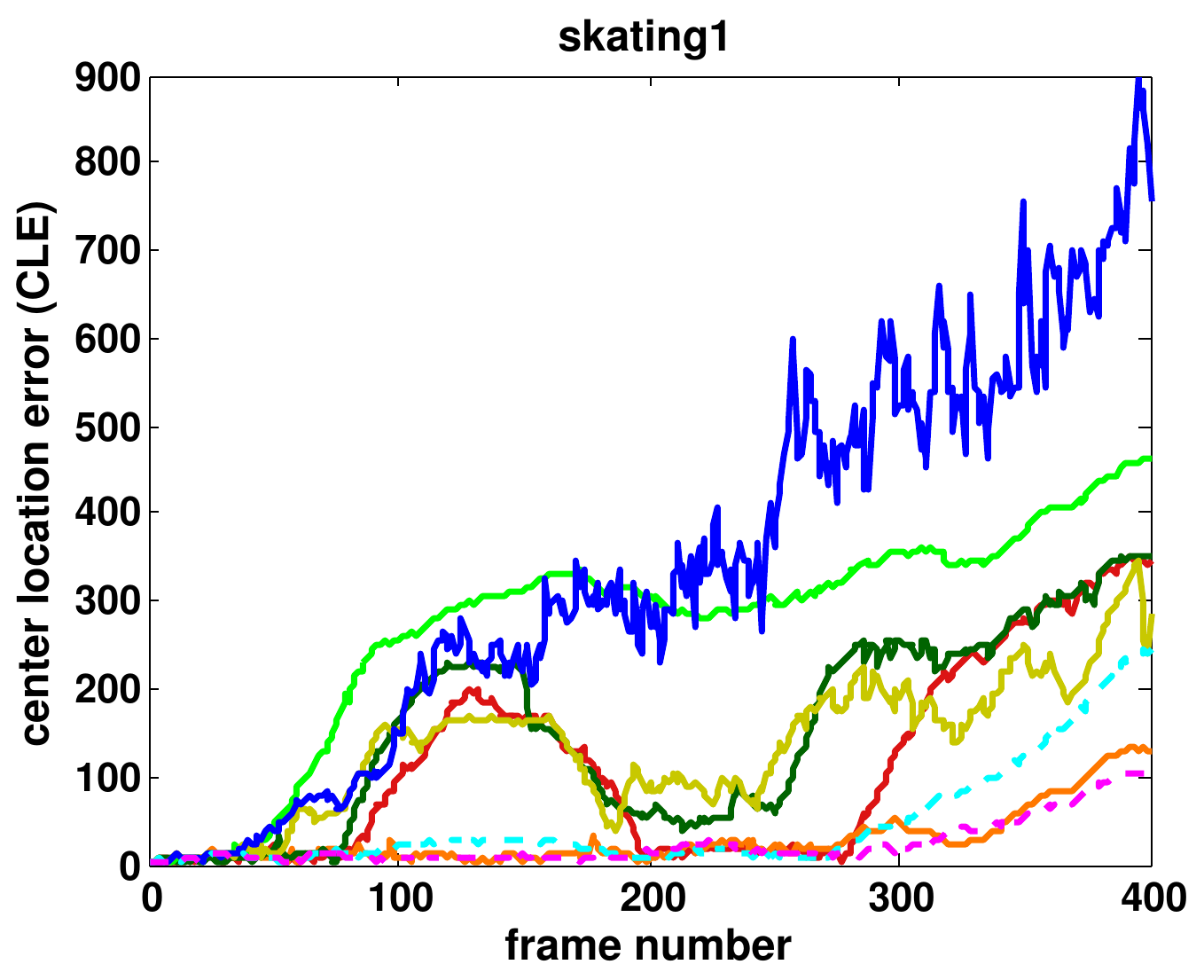}
\includegraphics[width=0.194\textwidth]{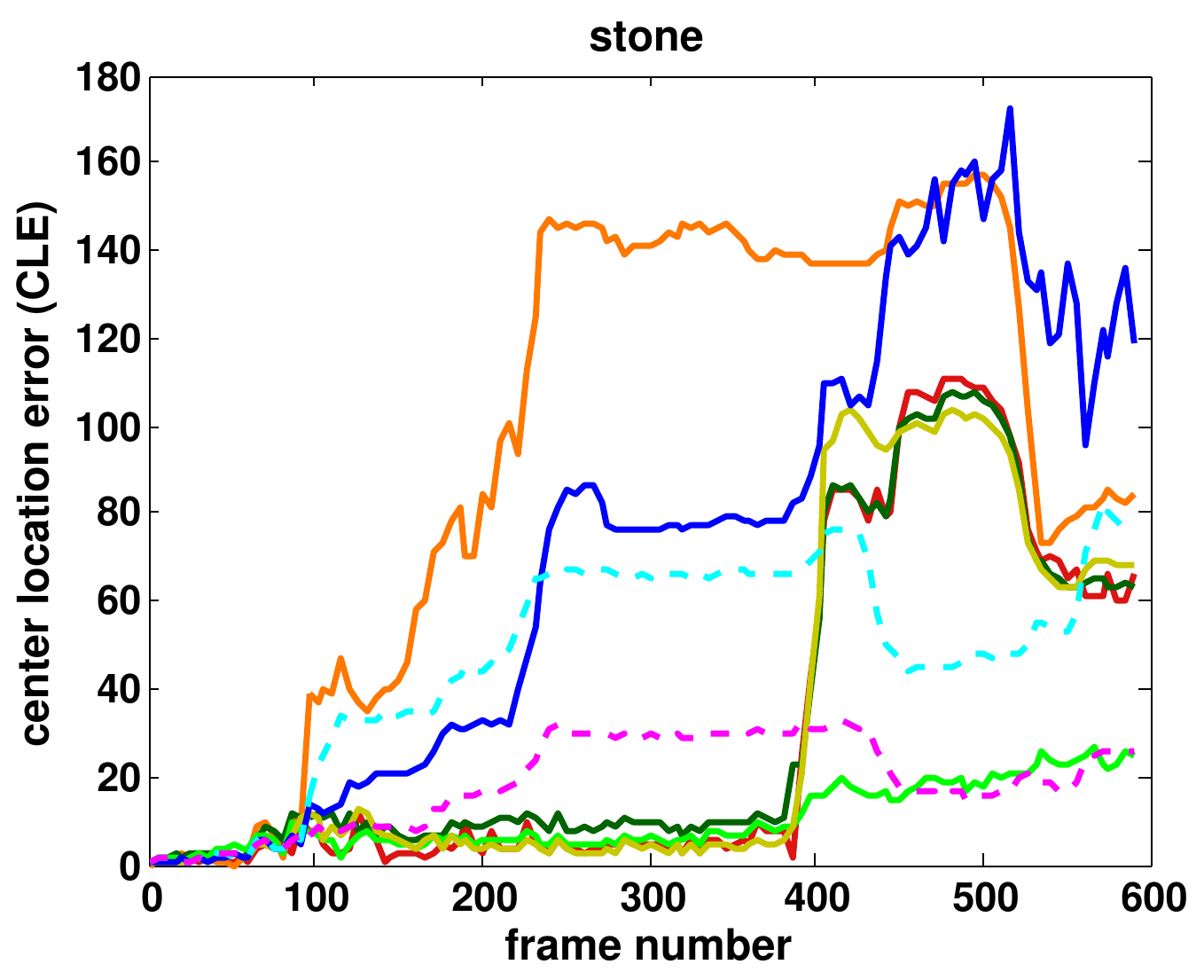}
\includegraphics[width=0.194\textwidth]{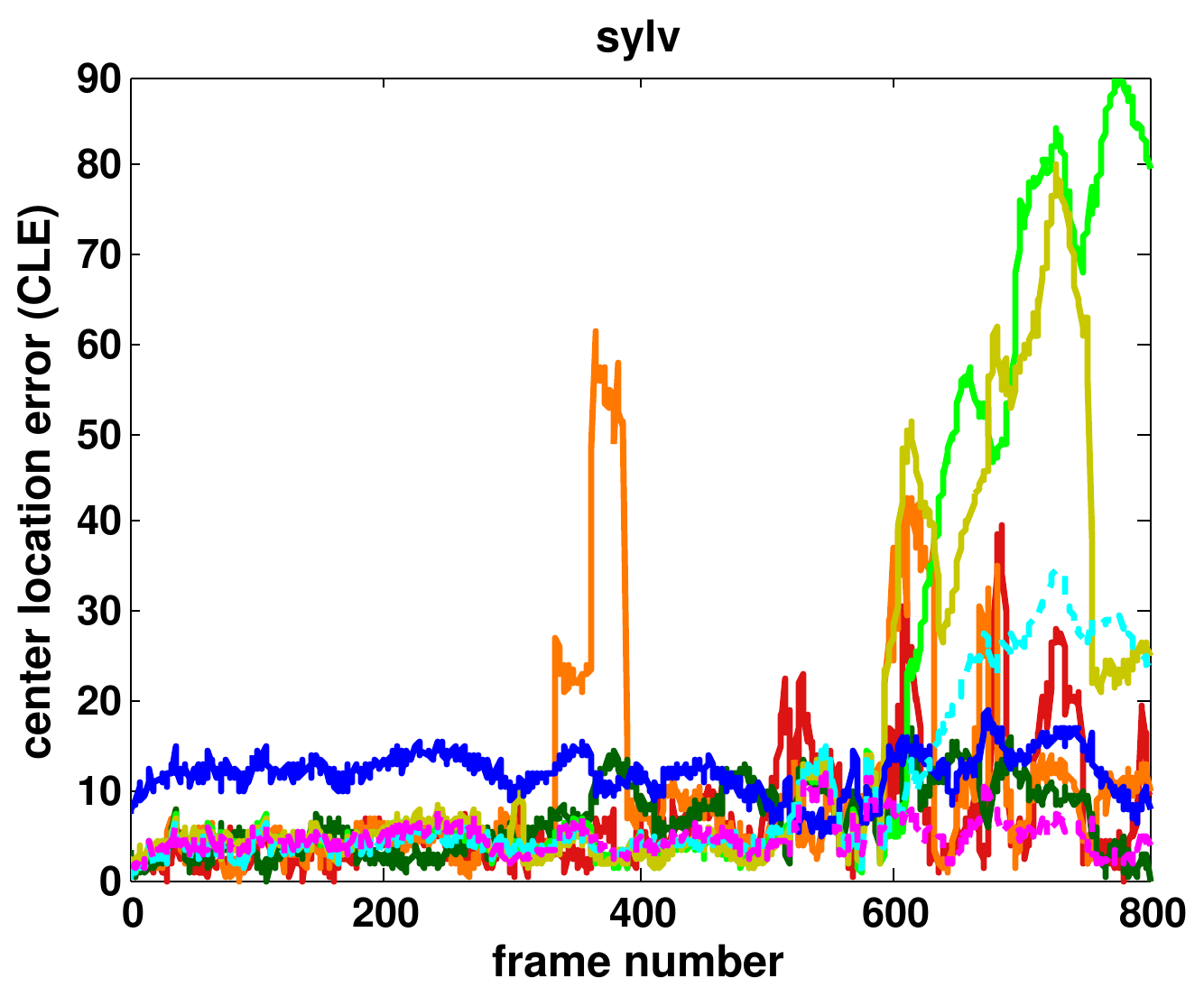}
\includegraphics[width=0.194\textwidth]{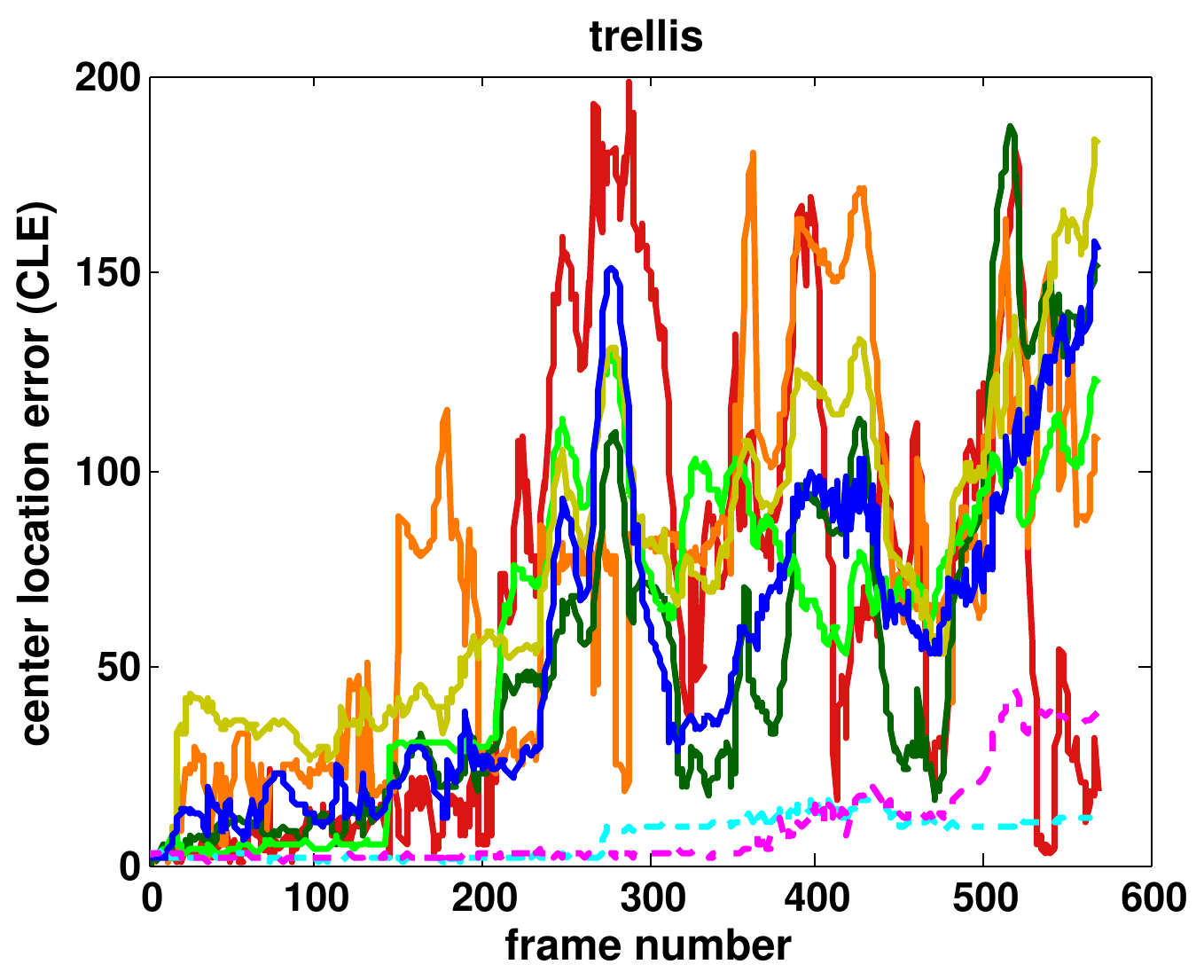}
\includegraphics[width=0.194\textwidth]{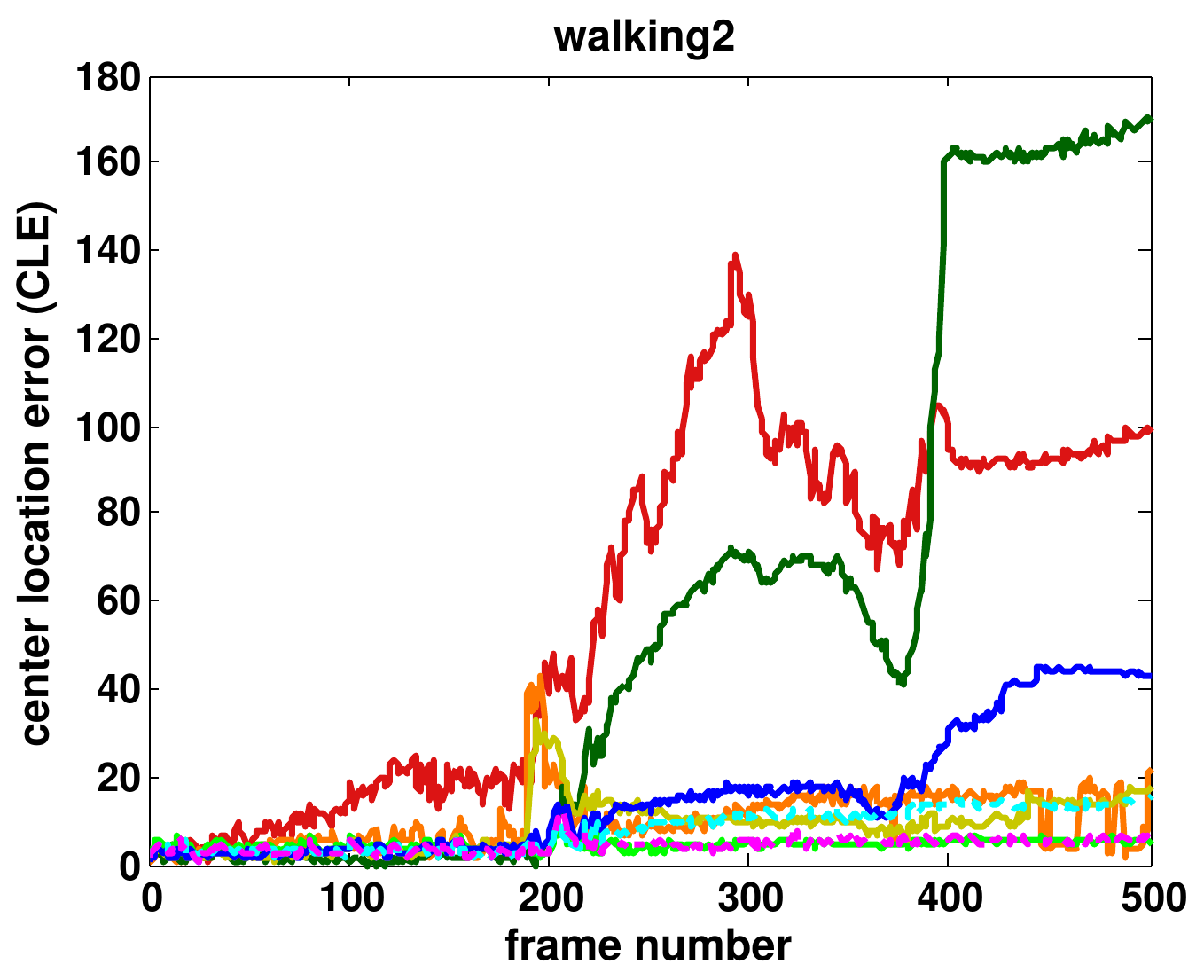}
\caption{\textbf{Center location error plots}. Distance between tracker and ground truth bounding boxes vs frame number. Results were averaged over 10 runs.}
\label{fig:avgCLE}
\end{figure*}

\begin{figure*}[!t]
\centering
\includegraphics[width=0.194\textwidth]{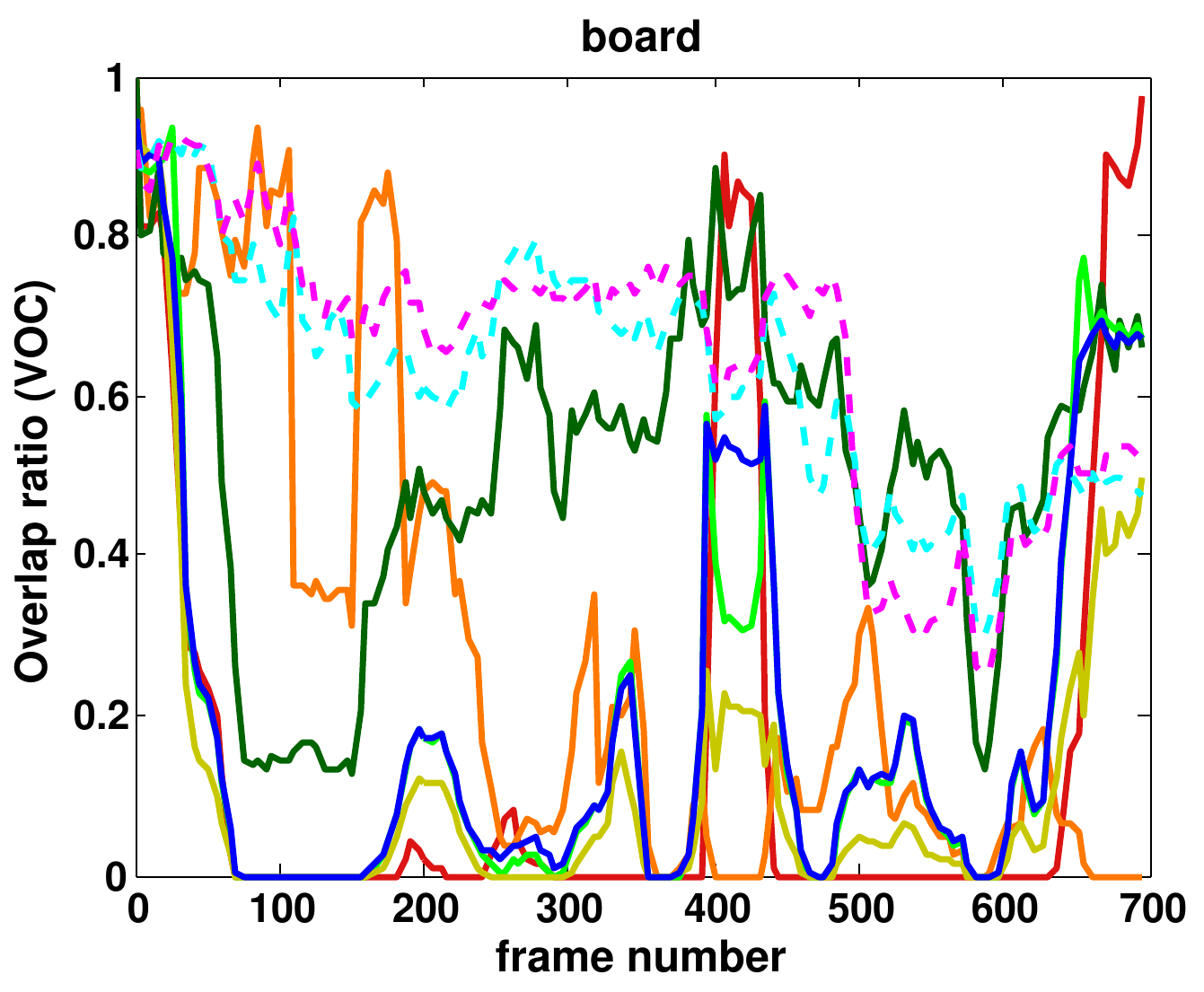}
\includegraphics[width=0.194\textwidth]{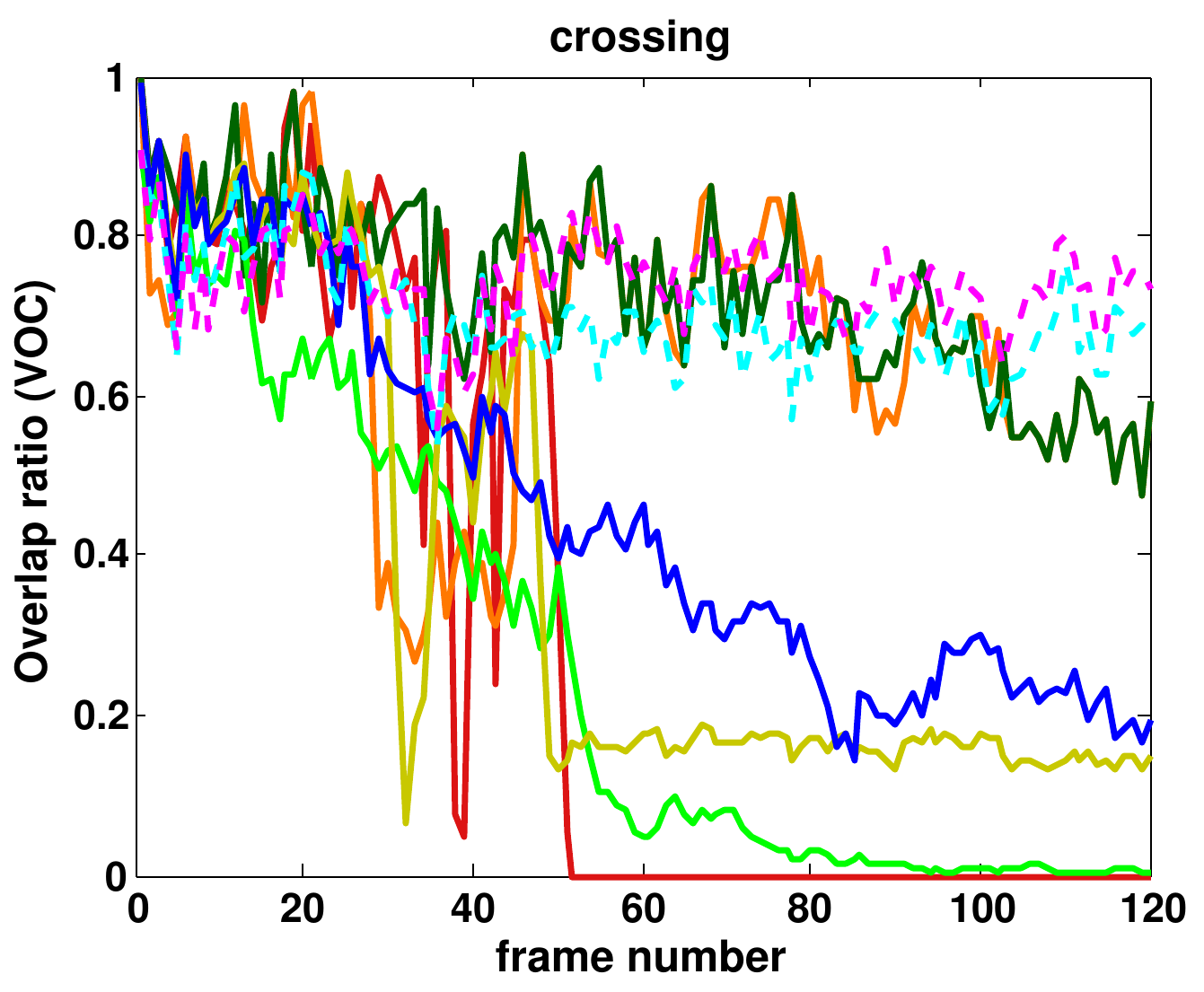}
\includegraphics[width=0.194\textwidth]{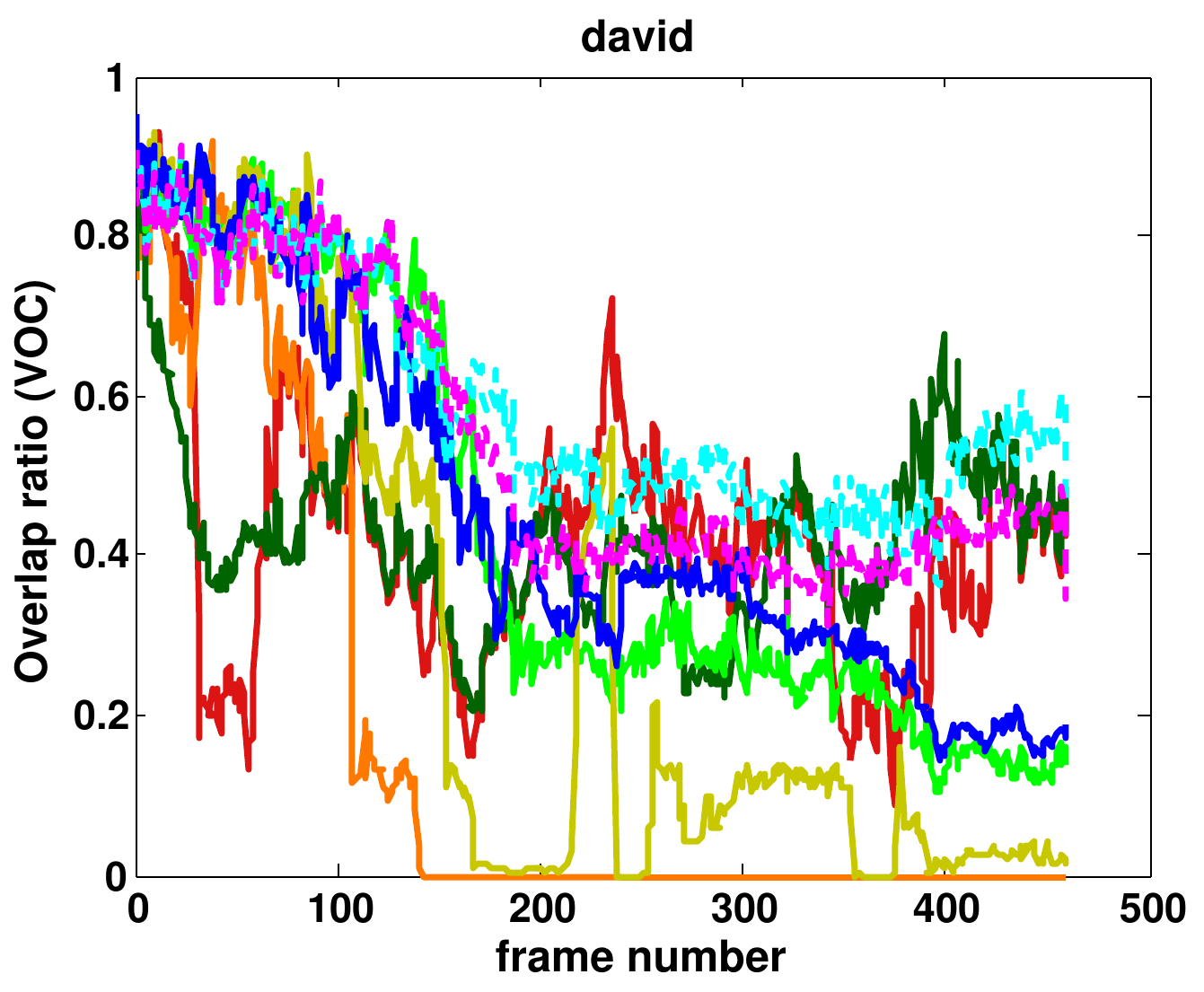}
\includegraphics[width=0.194\textwidth]{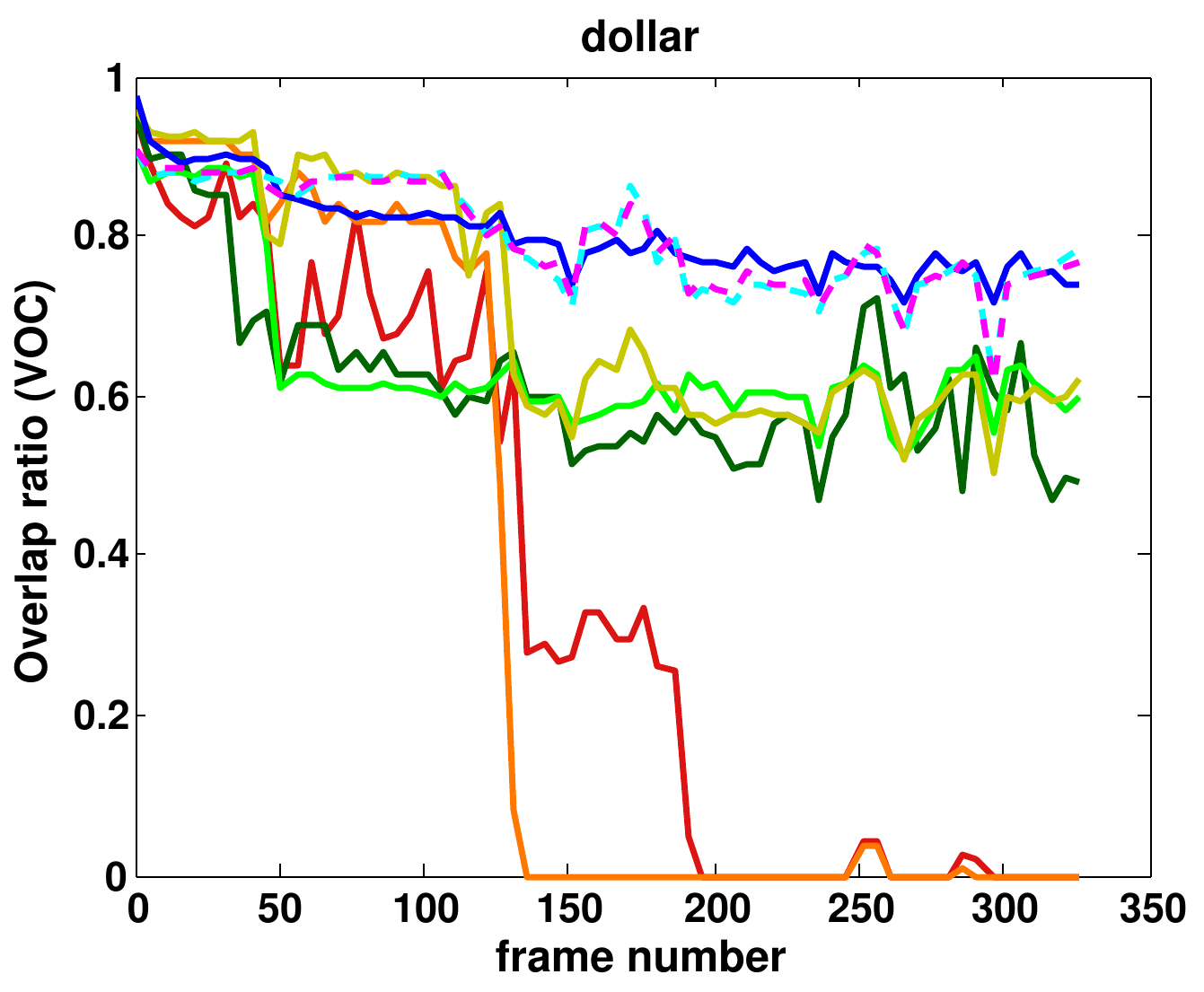}
\includegraphics[width=0.194\textwidth]{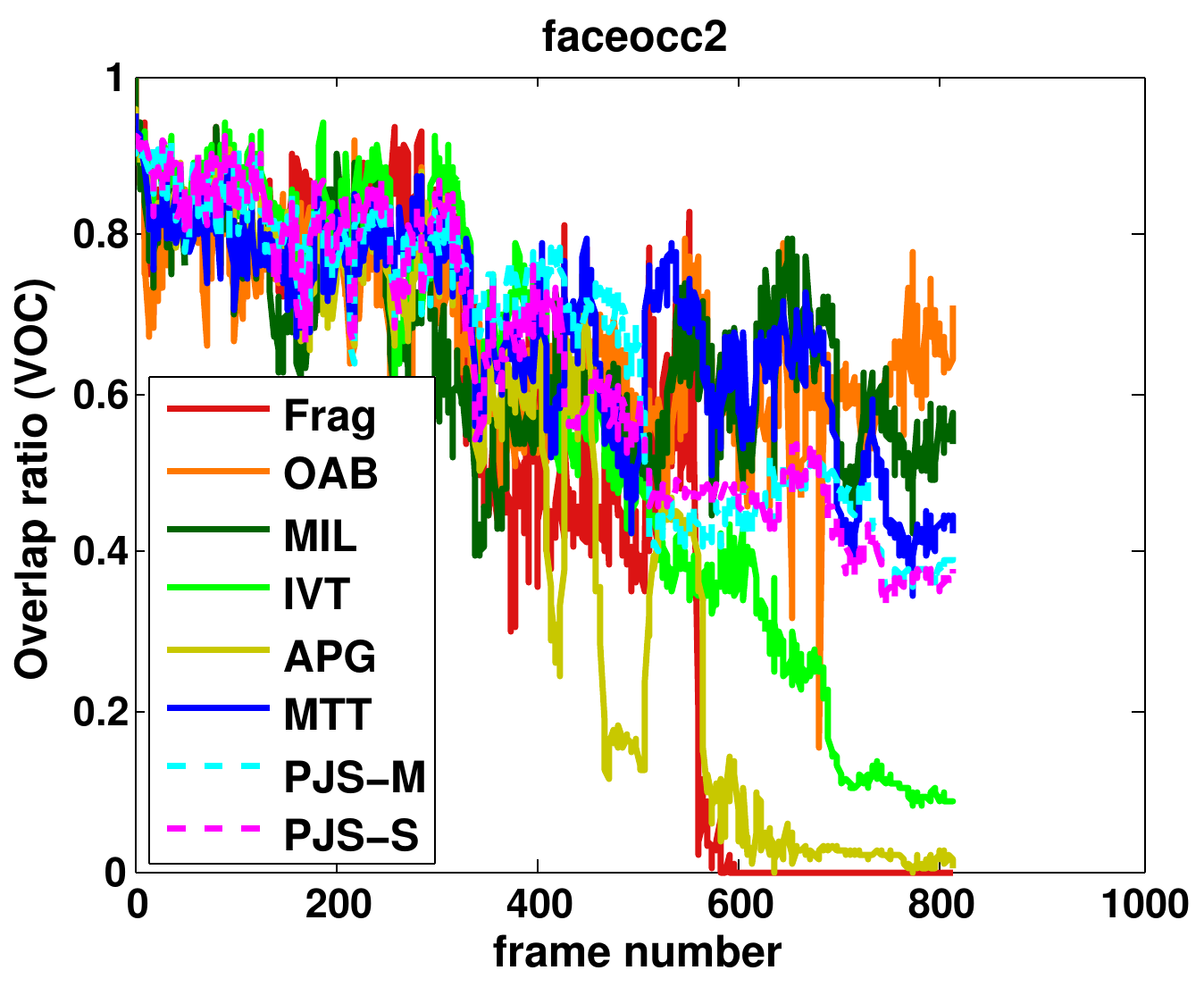}
\includegraphics[width=0.194\textwidth]{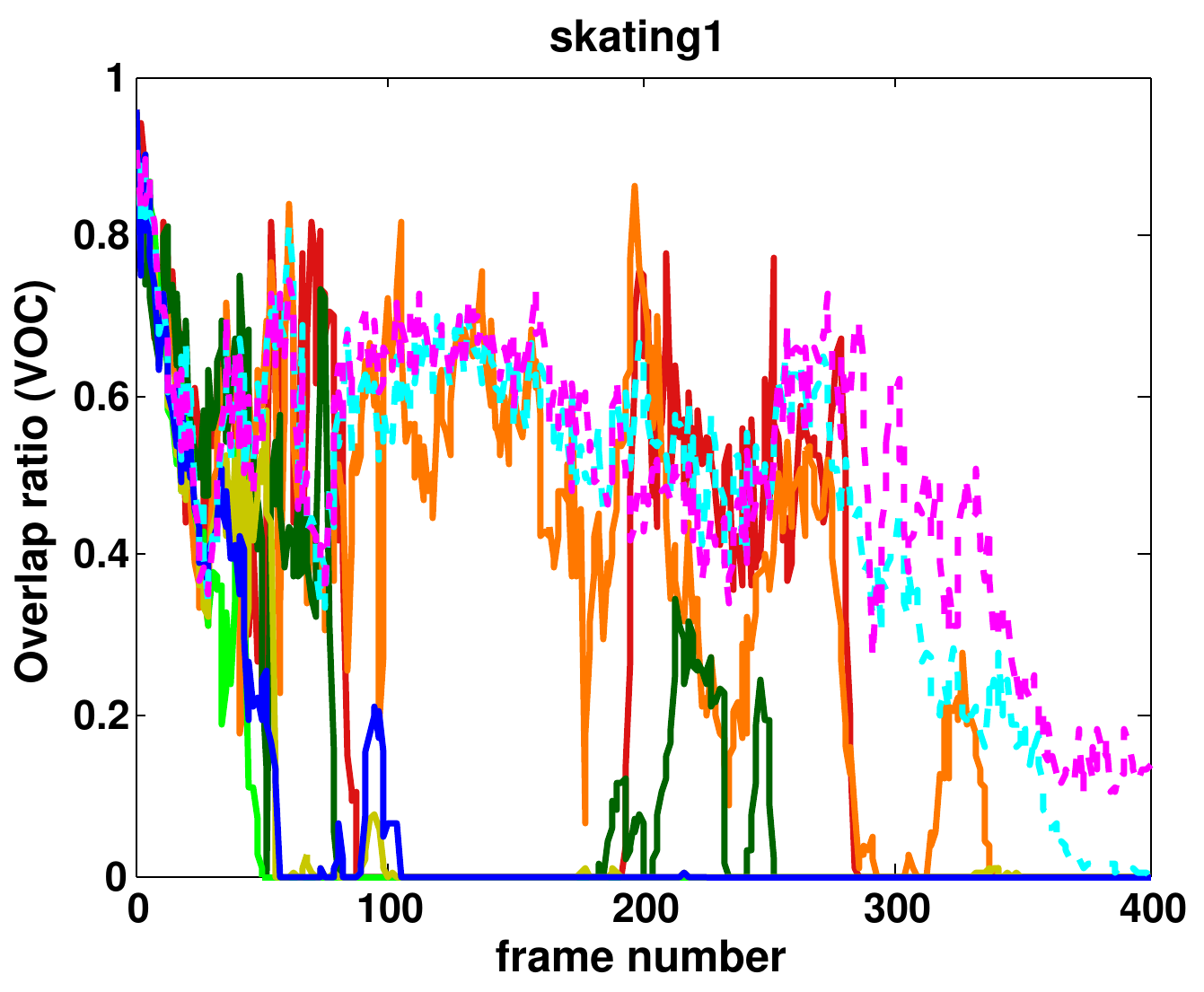}
\includegraphics[width=0.194\textwidth]{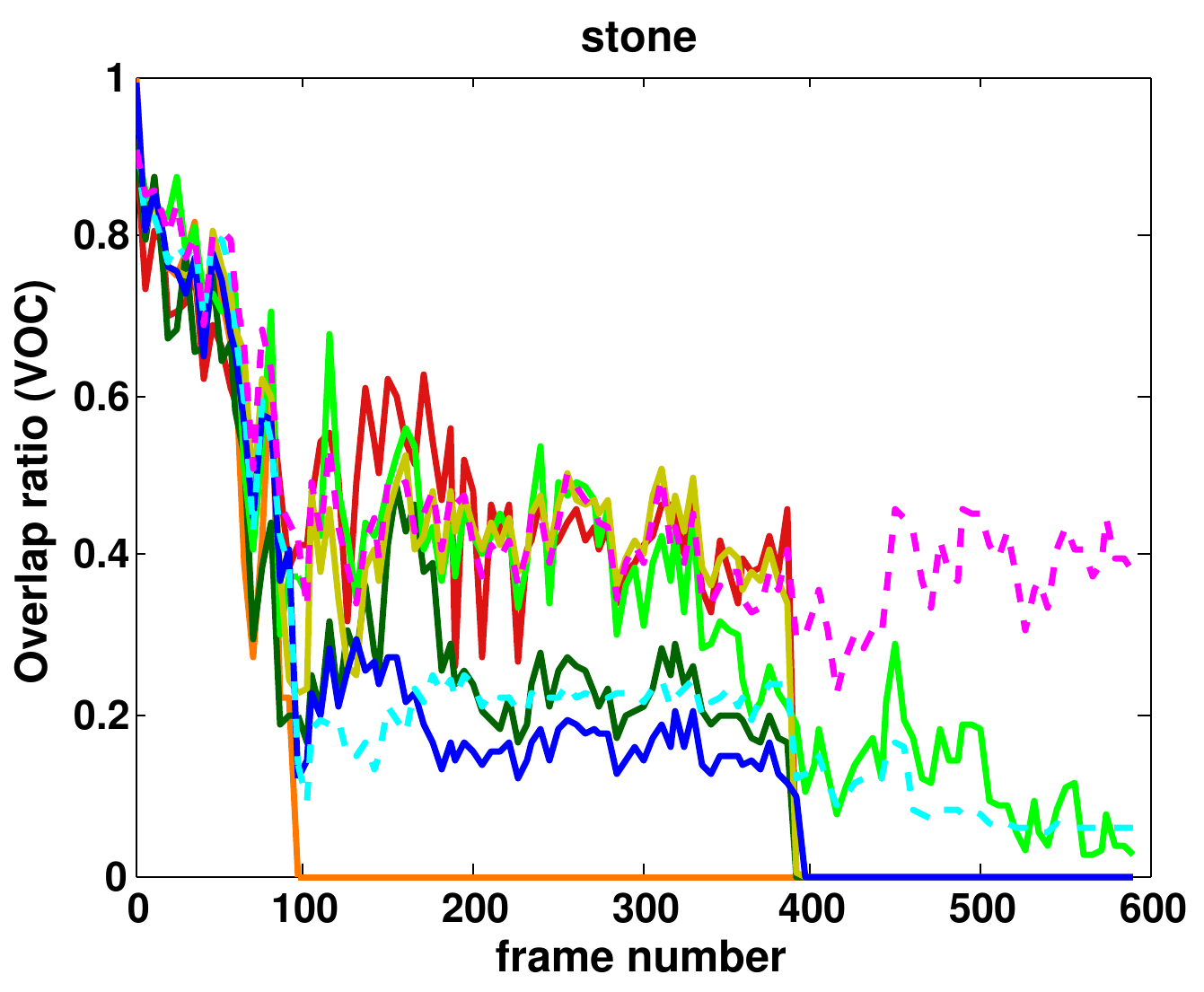}
\includegraphics[width=0.194\textwidth]{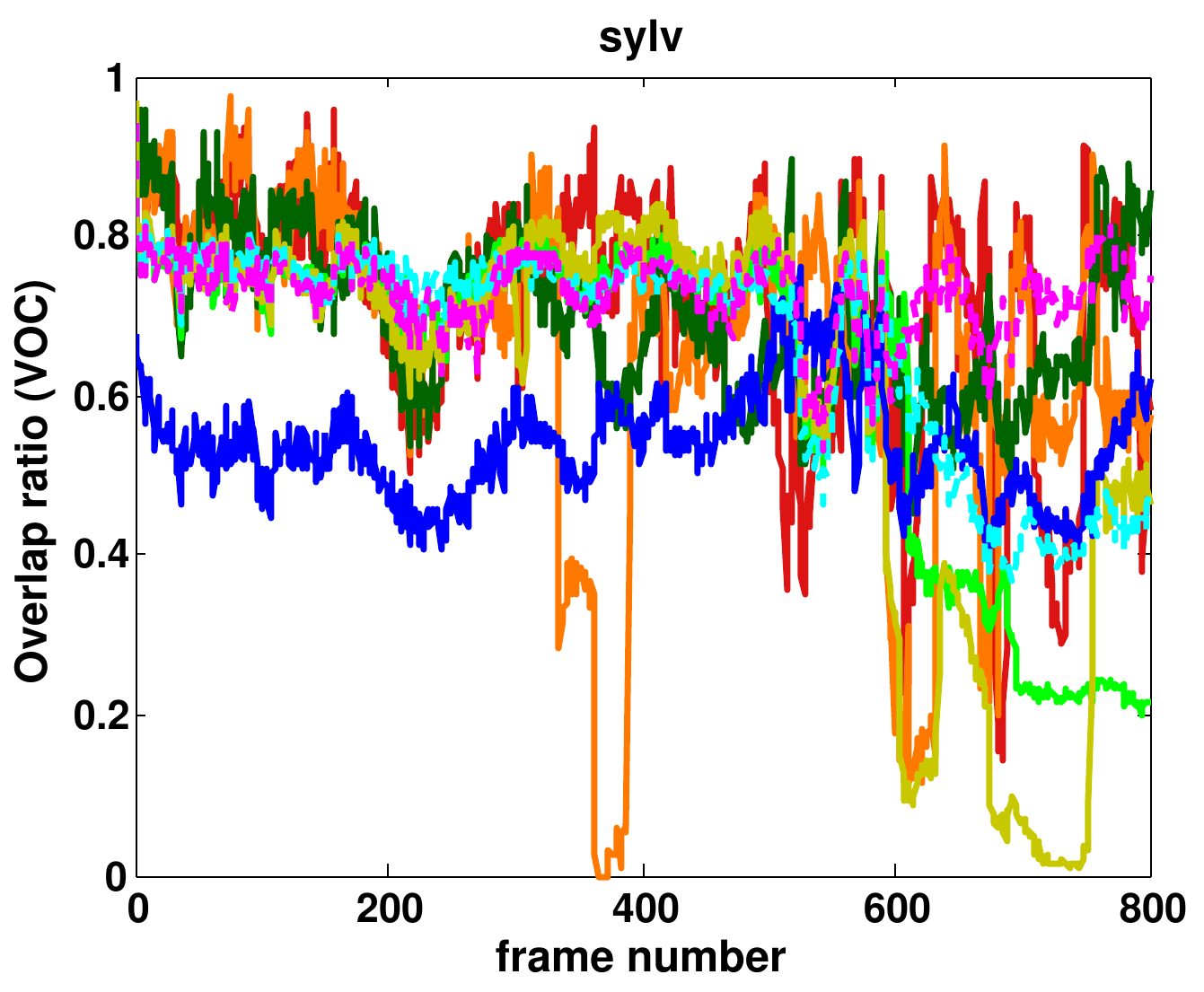}
\includegraphics[width=0.194\textwidth]{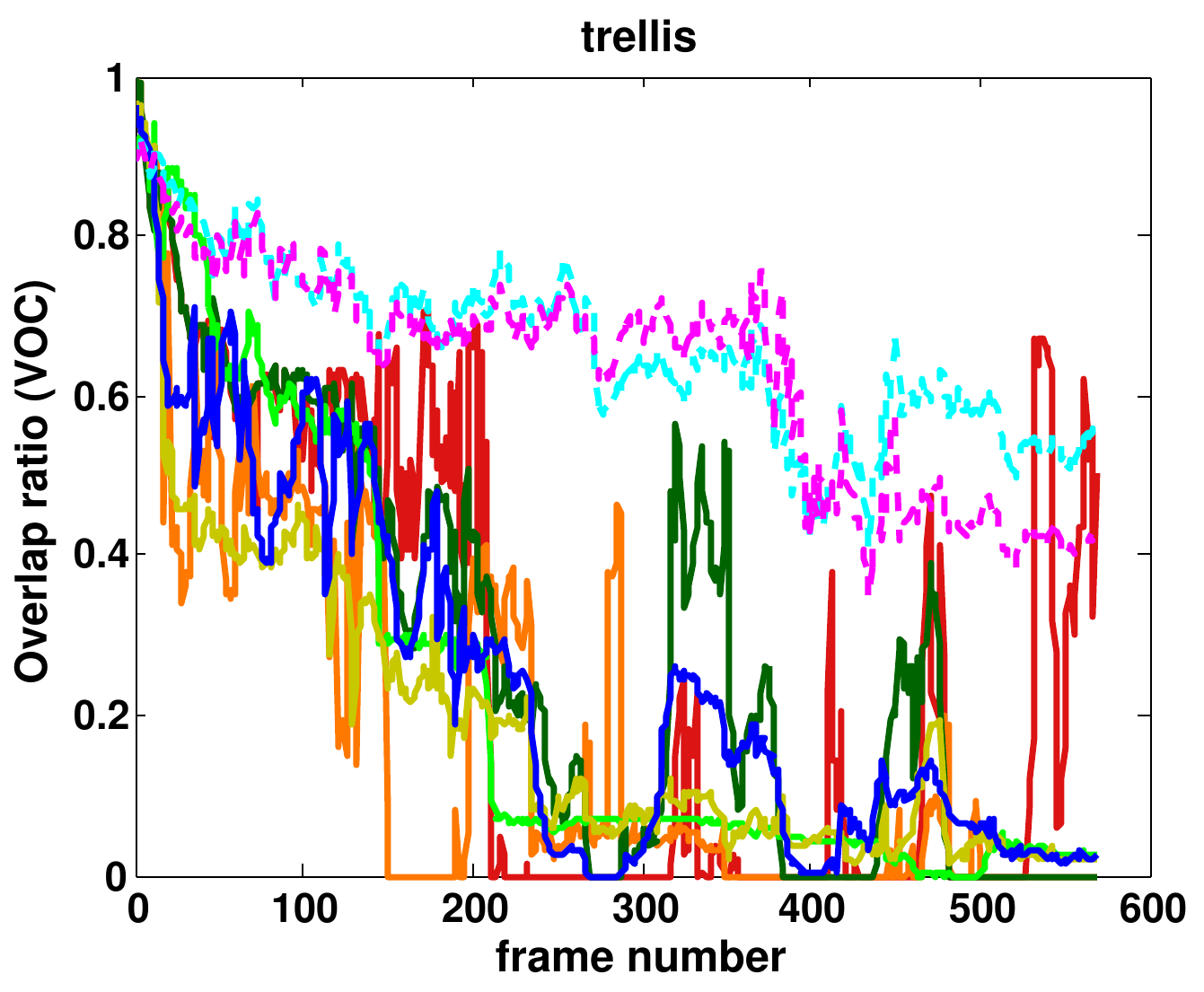}
\includegraphics[width=0.194\textwidth]{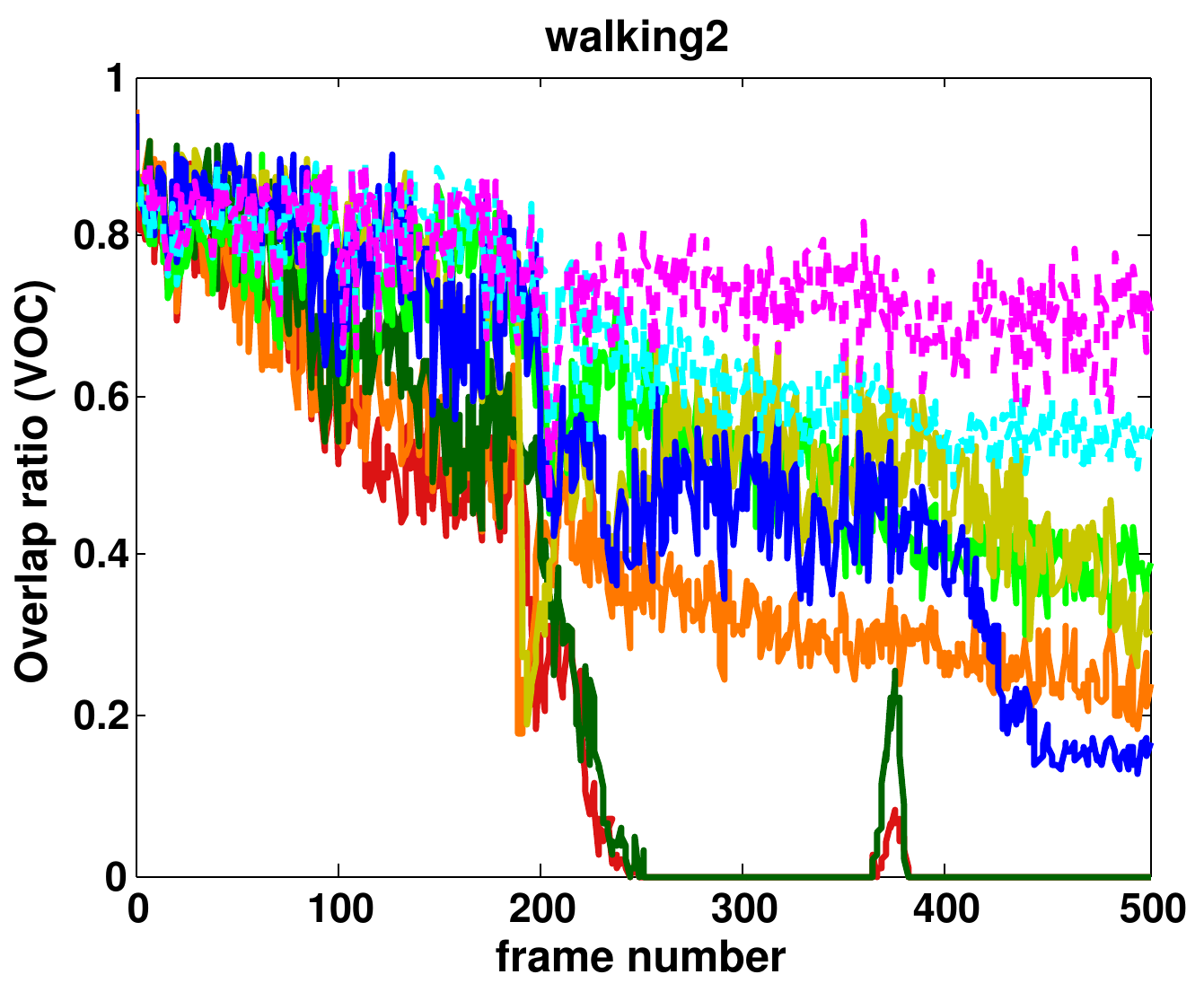}
\caption{\textbf{Overlap ratio plots}. Overlap ratio of tracker and ground truth bounding boxes vs frame number. Results were averaged over 10 runs.}
\label{fig:avgVOC}
\end{figure*}
\begin{figure*}[!t]
\centering
\includegraphics[width=0.194\textwidth]{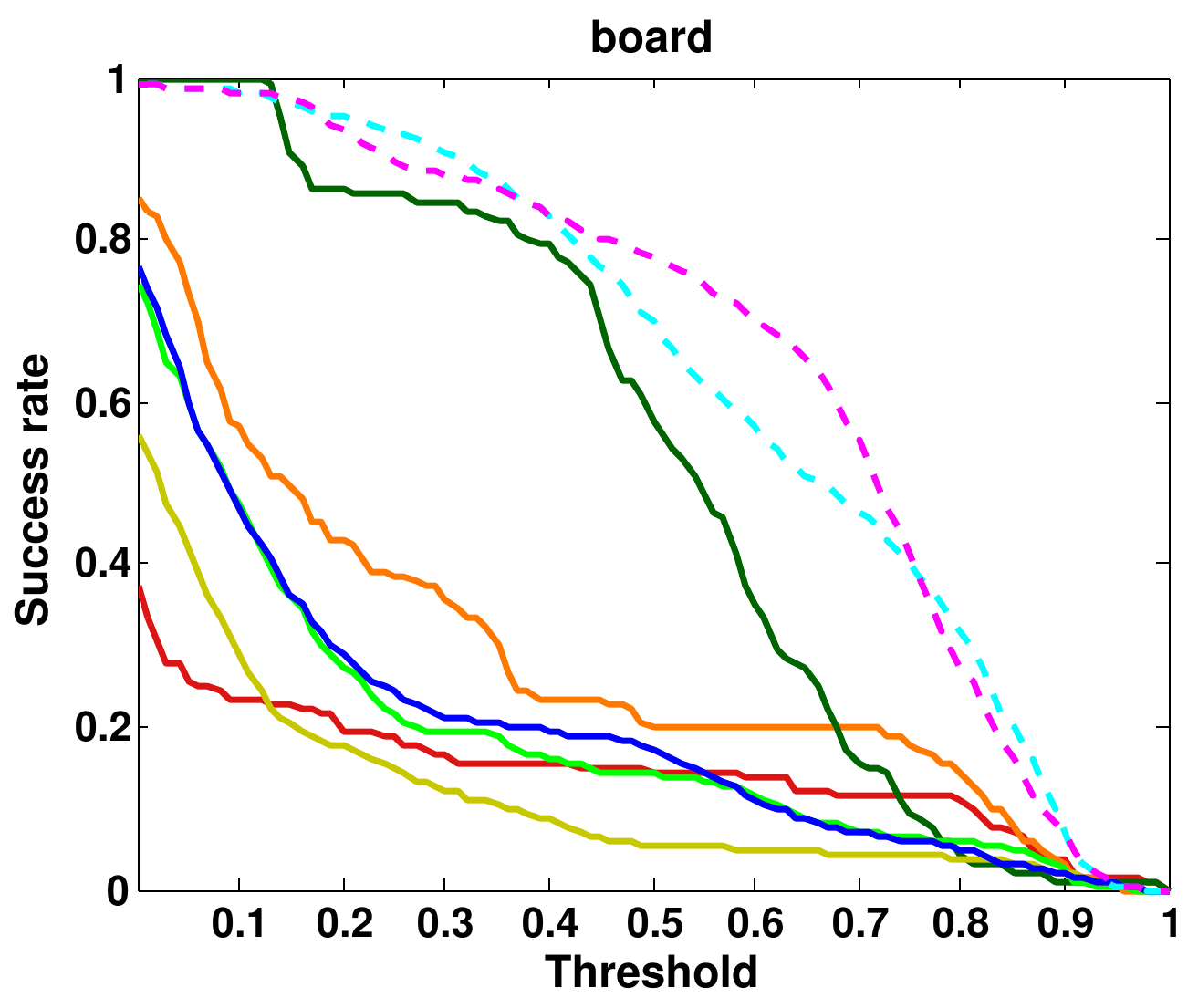}
\includegraphics[width=0.194\textwidth]{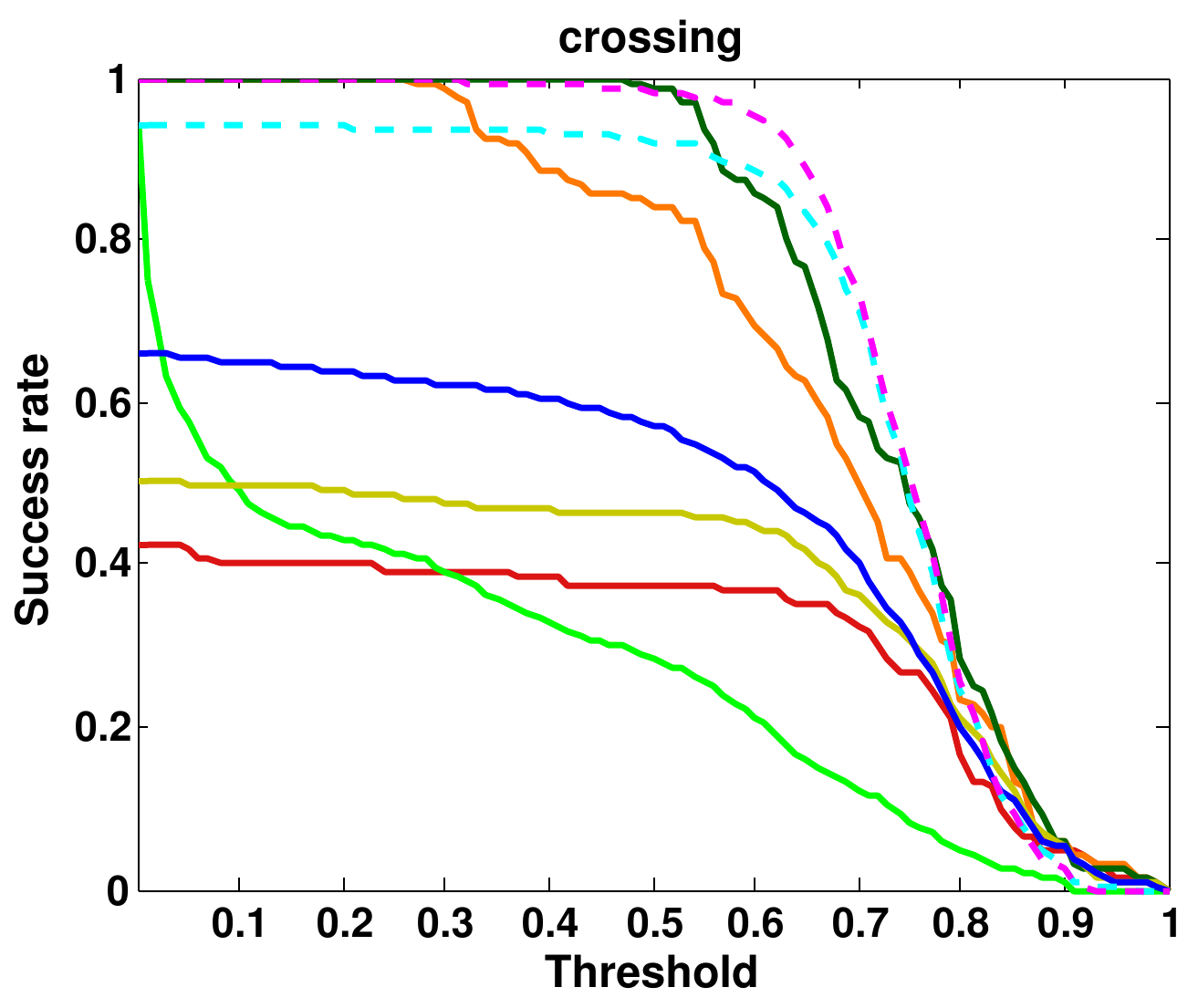}
\includegraphics[width=0.194\textwidth]{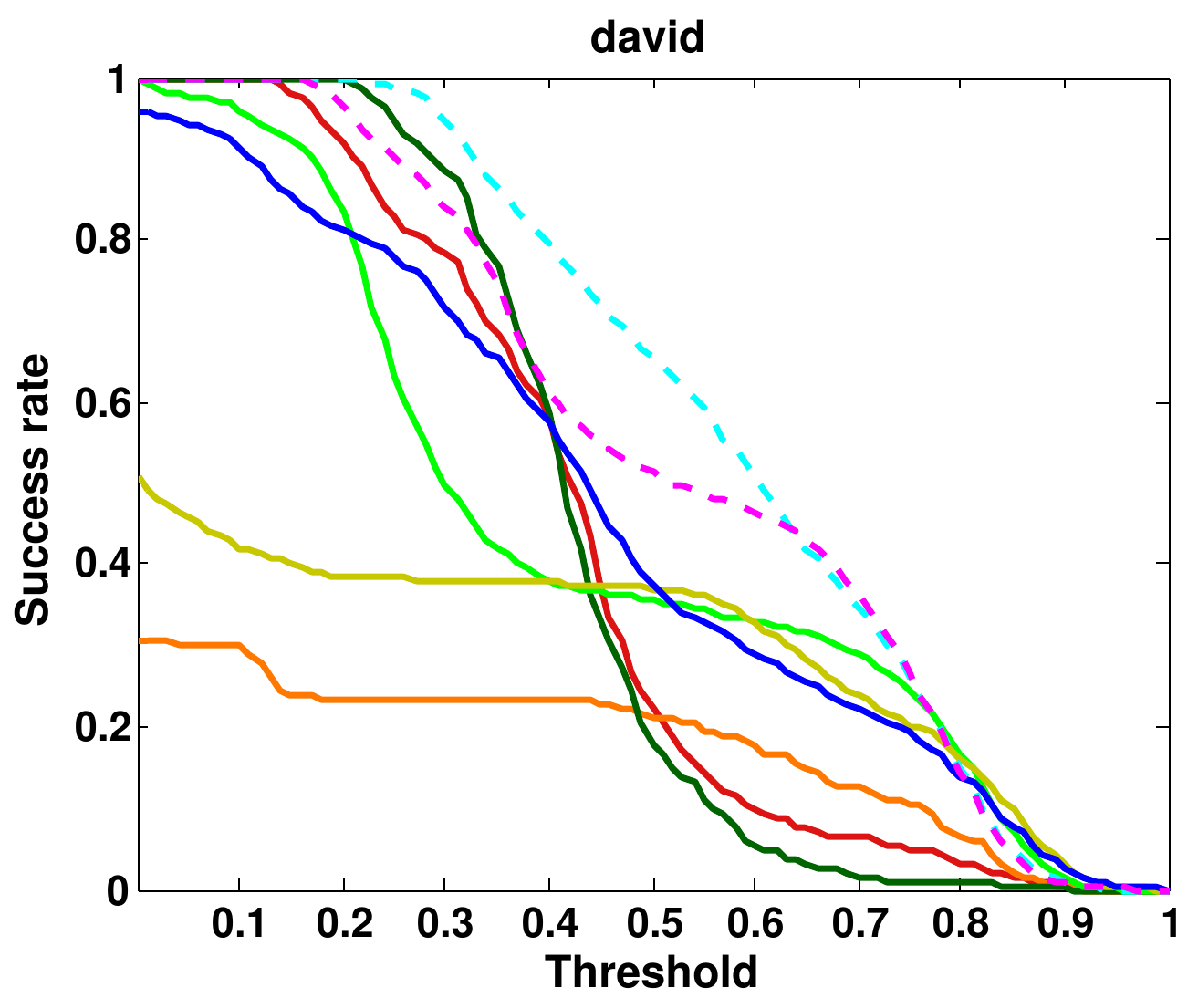}
\includegraphics[width=0.194\textwidth]{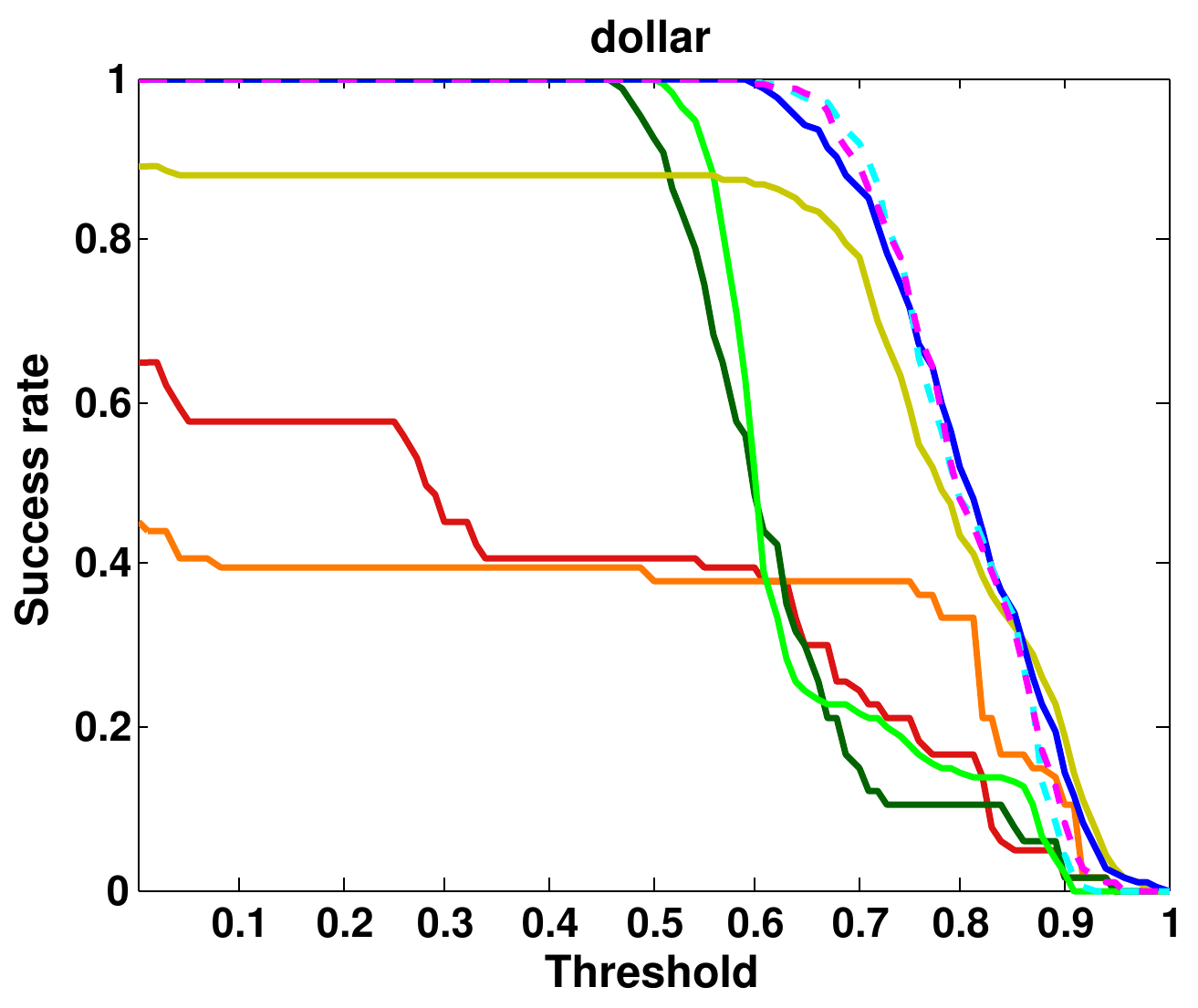}
\includegraphics[width=0.194\textwidth]{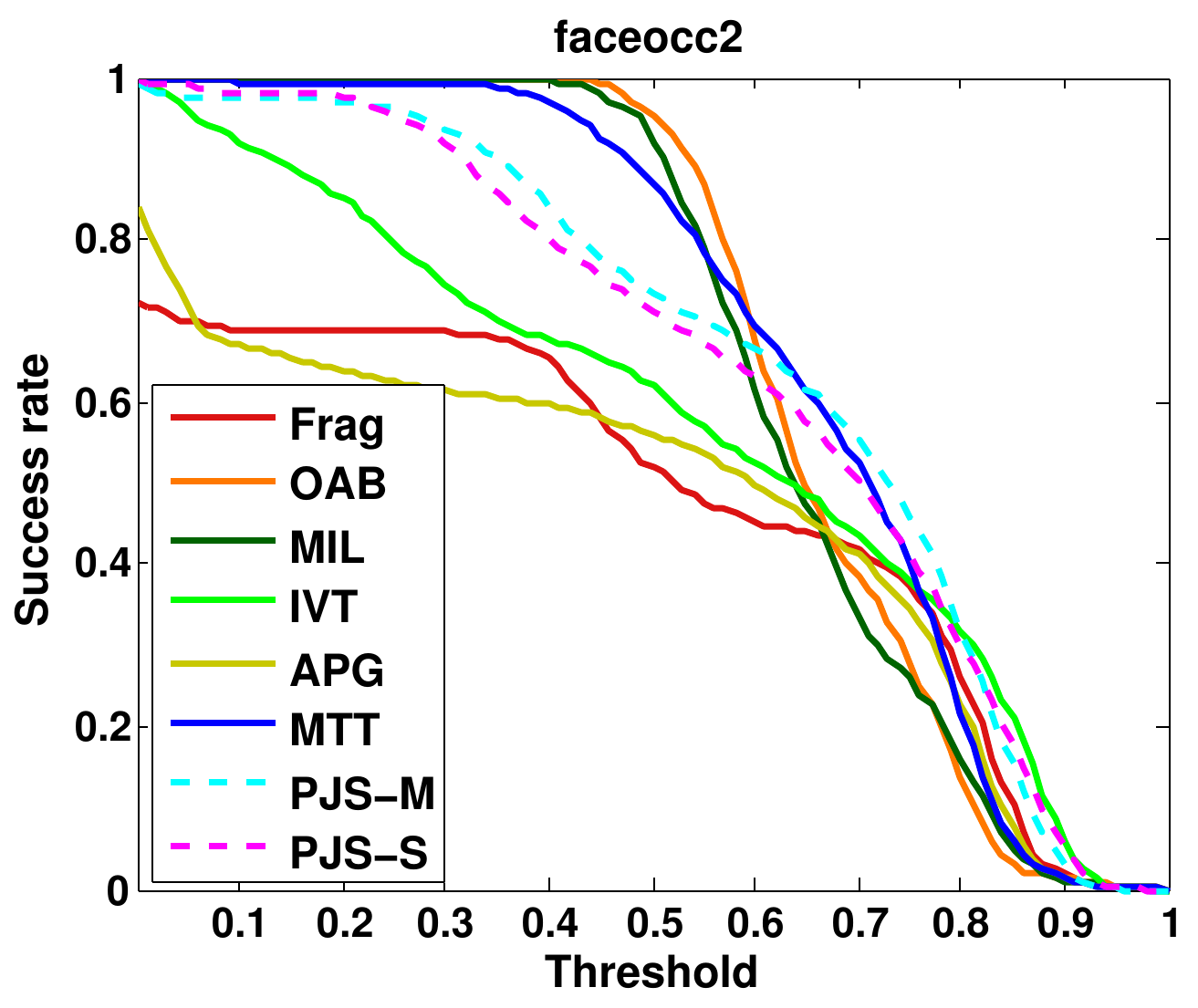}
\includegraphics[width=0.194\textwidth]{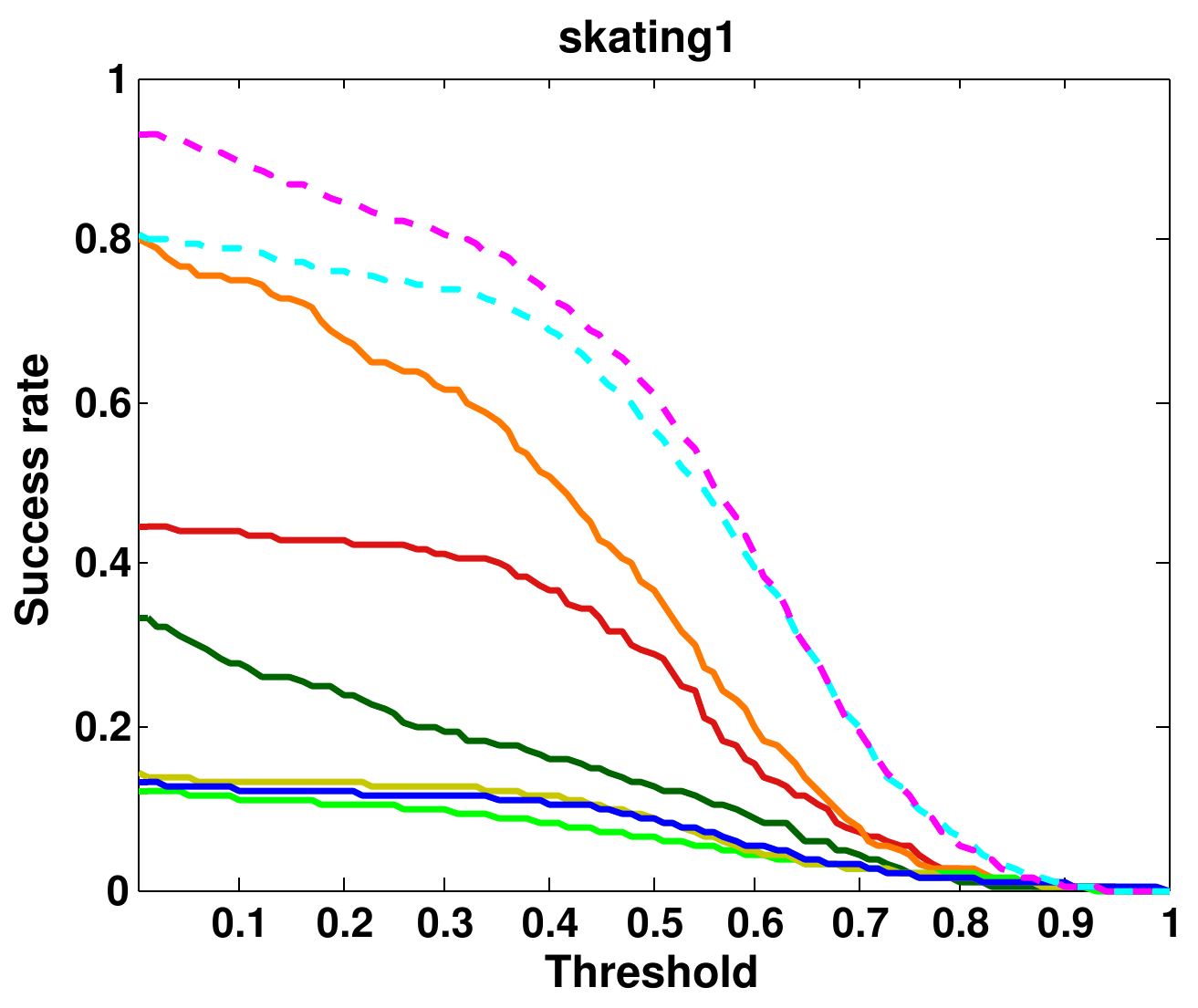}
\includegraphics[width=0.194\textwidth]{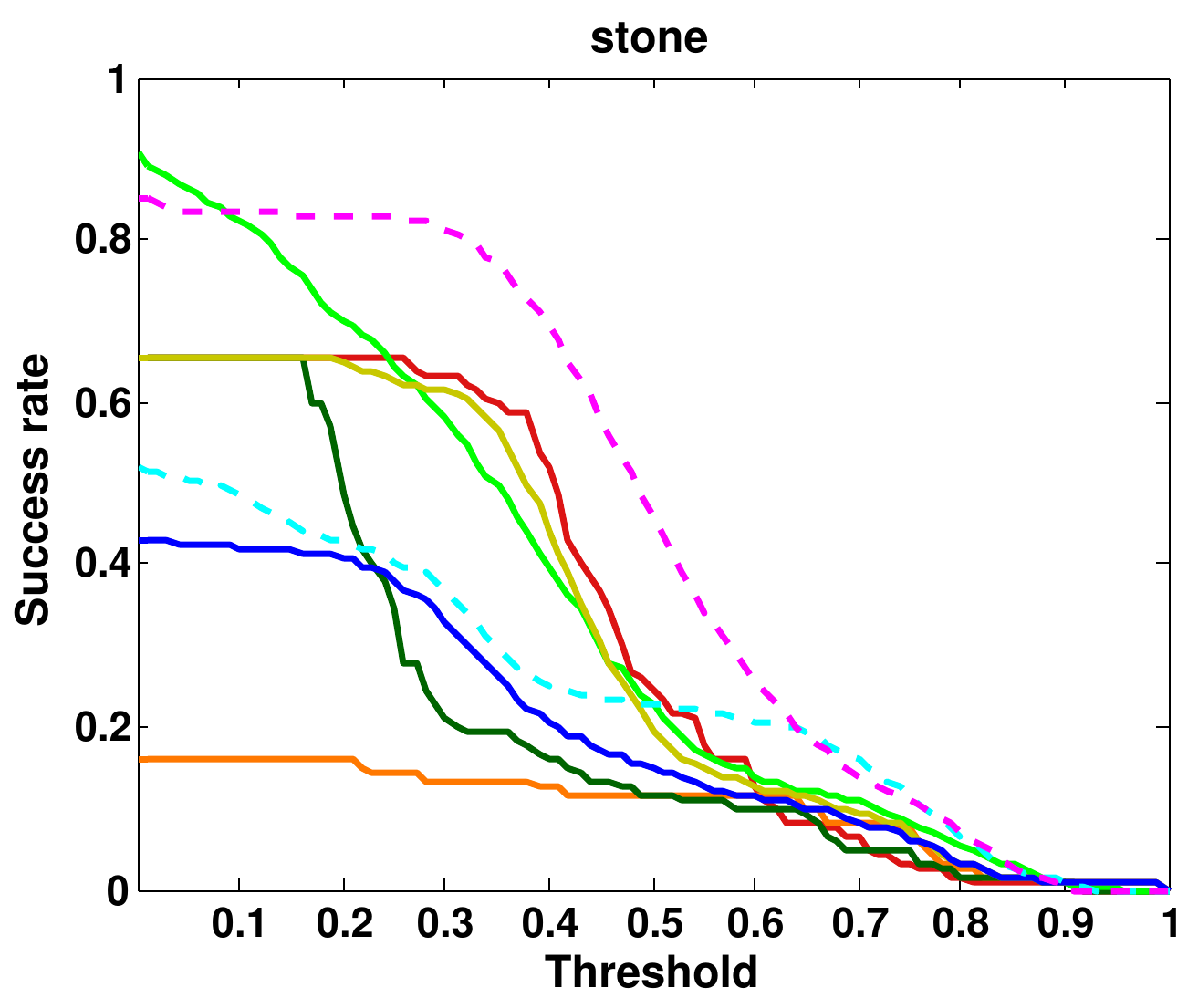}
\includegraphics[width=0.194\textwidth]{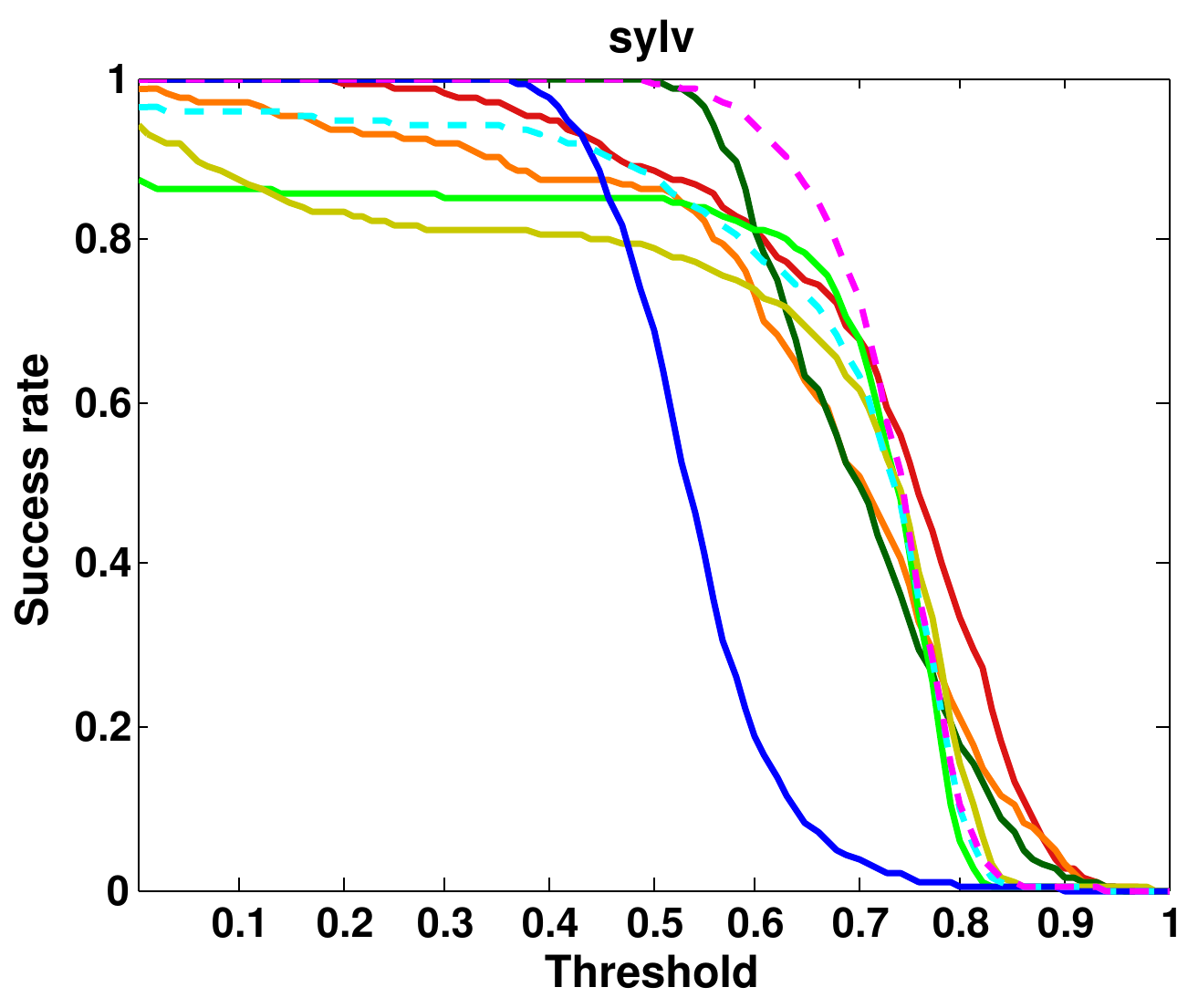}
\includegraphics[width=0.194\textwidth]{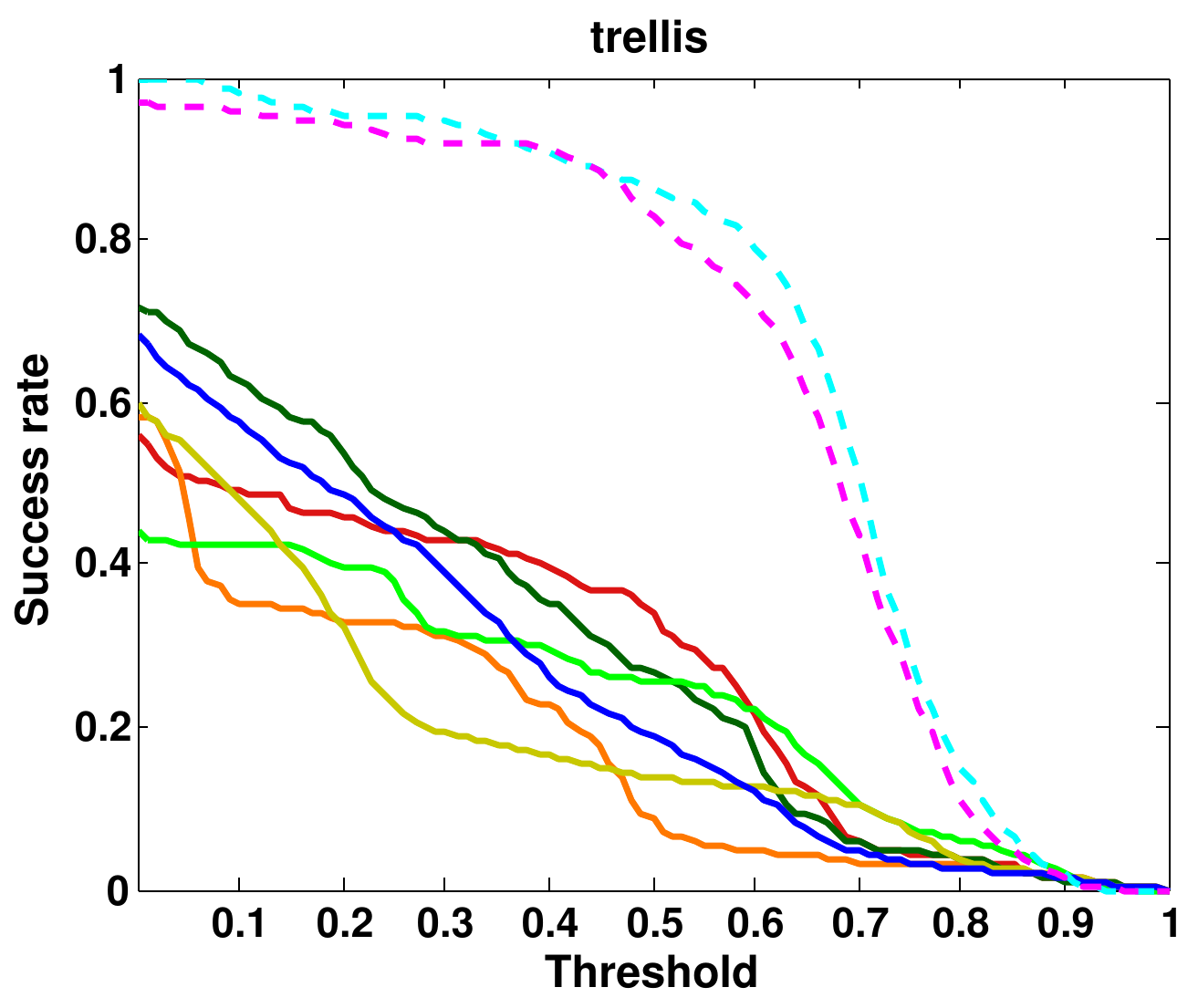}
\includegraphics[width=0.194\textwidth]{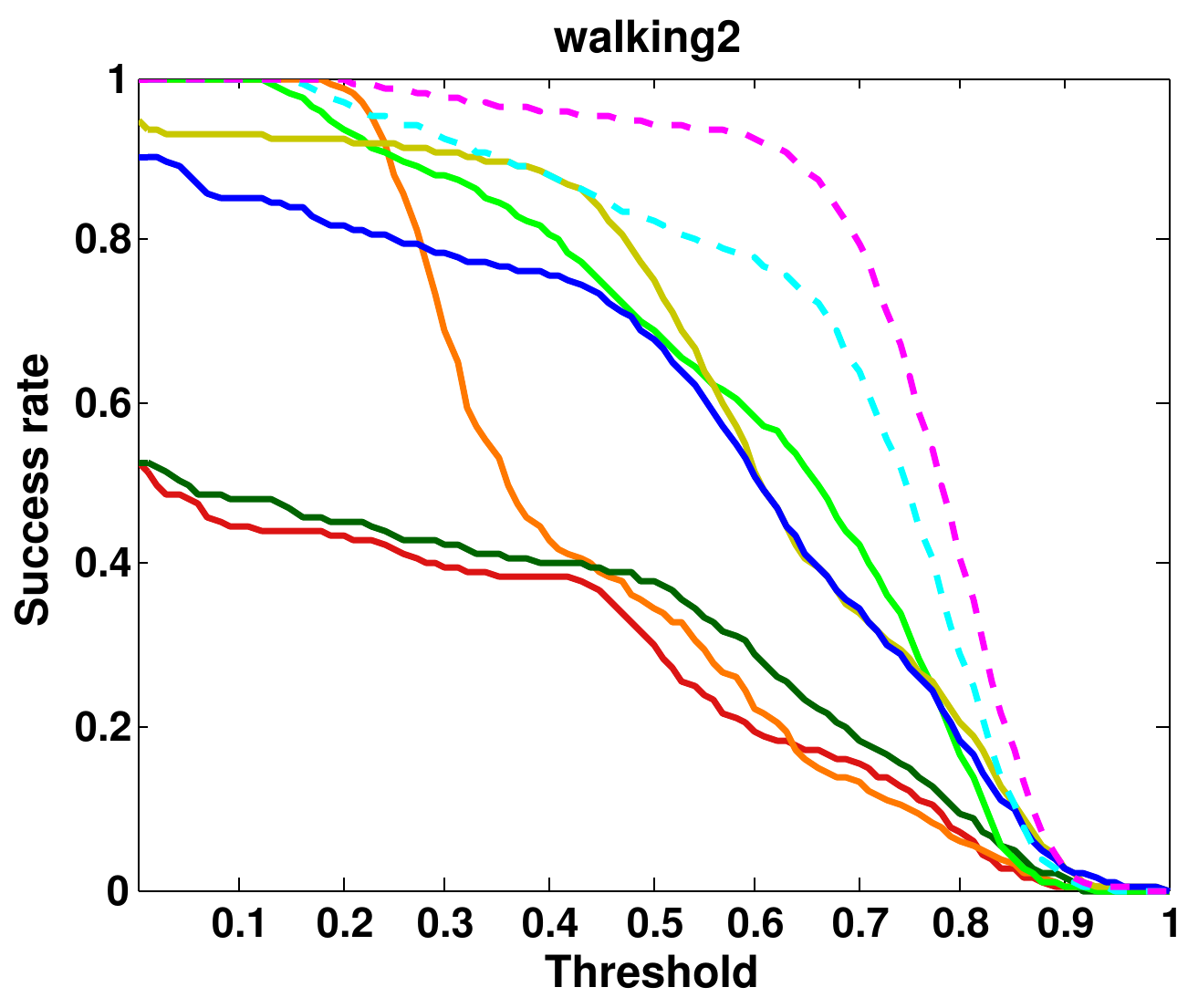}
\caption{\textbf{Success plot diagrams}. Success rates for thresholds in $[0\;1]$. Results were averaged over 10 runs.}
\label{fig:avgROC}
\end{figure*}

\subsection{Quantitative Comparison} \label{subsec:quantitative}
The Center Location Error (CLE), and PASCAL overlap score (VOC) are typical quantitative measures which are usually plotted as a function of frame number.
We have also calculated the mean values of these measures and presented them in tabular format for different trackers and datasets. The center location error measures the Euclidian distance between the centers of tracker, and ground truth bounding boxes. The frame and target size or orientation are not taken into account by CLE. Furthermore, after the target is lost, large values of CLE become meaningless and corrupt the mean CLE value. To resolve these deficiencies, the PASCAL overlap score is used. This score is defined as the ratio of intersection to union of tracker and ground truth bounding boxes: $\frac{|S_{gt} \cap S_{tr}|}{|S_{gt} \cup S_{tr}|}$, where $|R|$ denotes the number of pixels in region $R$. This measure which is normalized within $[0 \; 1]$, is independent of video and target sizes or orientation. The success rate measure which is equal to the ratio of successfully tracked frames in a video sequence, may be used to express the overall performance of a tracker. Moreover, a frame is successfully tracked if its overlap score is grater than a predefined threshold. 

\begin{table}[!t]
\centering
\caption{From top to bottom: CLE, overlap ratio, and  success rate averaged on 10 runs for different trackers on 10 datasets. Best two results are highlighted.}
\includegraphics[width=0.45\textwidth]{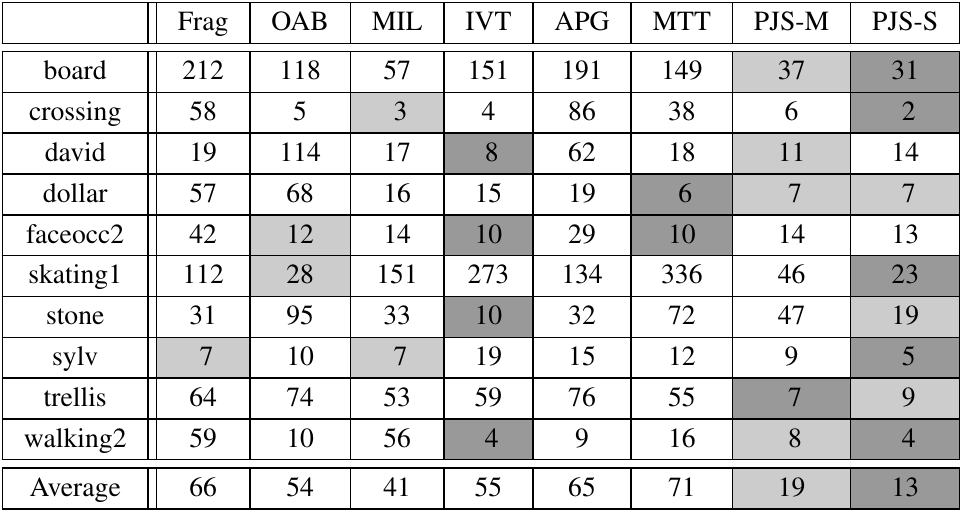}\vspace{4mm}
\includegraphics[width=0.45\textwidth]{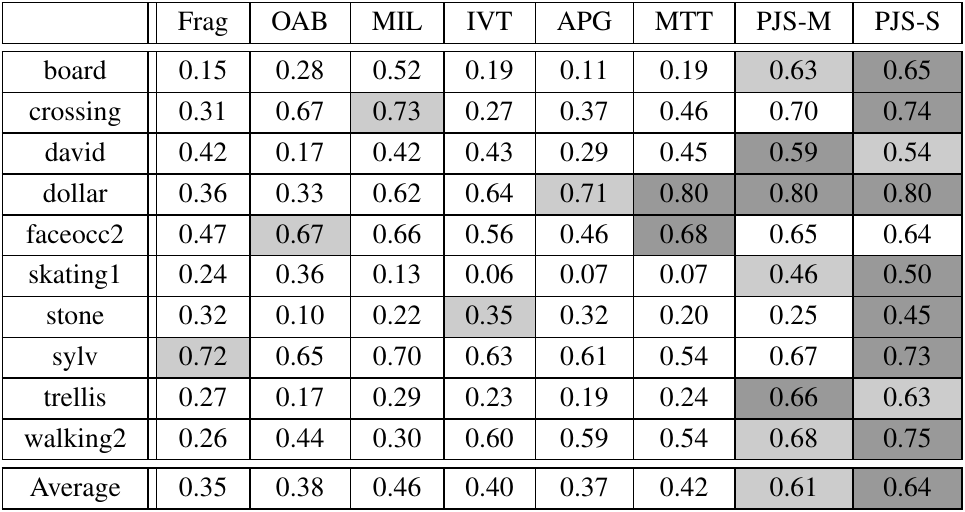}\vspace{4mm}
\includegraphics[width=0.45\textwidth]{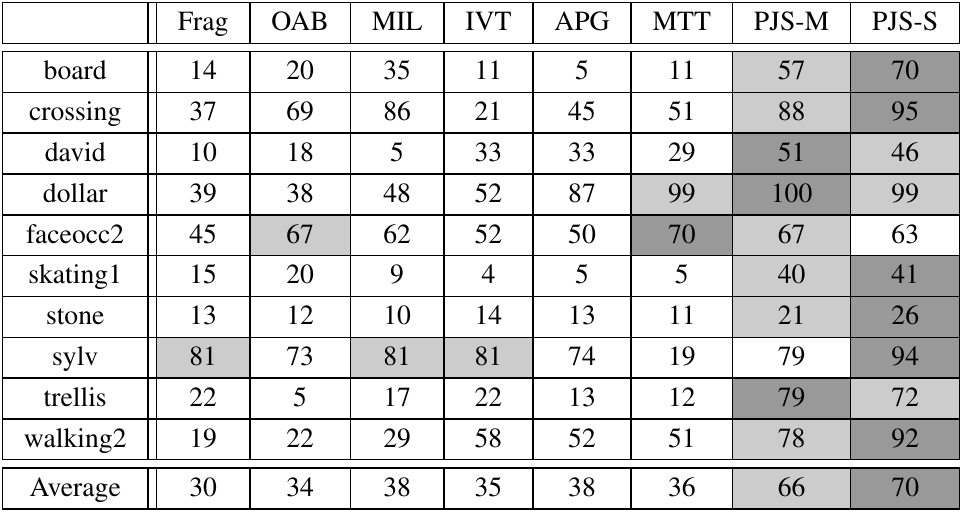}
\label{tbl:avgCLE}
\end{table}

 

Figures \ref{fig:avgCLE} and \ref{fig:avgVOC} show the CLE measure and overlap score vs frame number, for the average results of all trackers, respectively. As illustrated, PJS-S and PJS-M are among the trackers with lowest CLE and highest VOC. As we mentioned earlier, CLE alone is not a fair measure for comparison. For example, in the walking sequence PJS-S and IVT have the lowest CLE but when comparing the VOC, PJS-S and PJS-M gain the best results. From these figures, we can observe the average performance of the trackers for the entire video sequence, and evaluate them with regard to the challenges posed in different frames. In this regard, the results confirm the observations of Section \ref{subsec:qualitative}.
Table \ref{tbl:avgCLE} provides a numerical comparison of mean CLE, overlap ratio, and success rate for an overlap ratio threshold equal to 0.6, over 10 runs for each pair of dataset and tracker. The average results over all datasets in the last row of the tables shows that the proposed trackers are the best among the competing trackers. Also, in all sequences except \textit{faceocc2}, the proposed trackers achieve the first or second place.
The main reason is the patchwise joint-sparse nature of our method, which causes the bounding boxes to accurately align with the target.  
\begin{figure*}[t]
\centering
\includegraphics[width=0.25\textwidth]{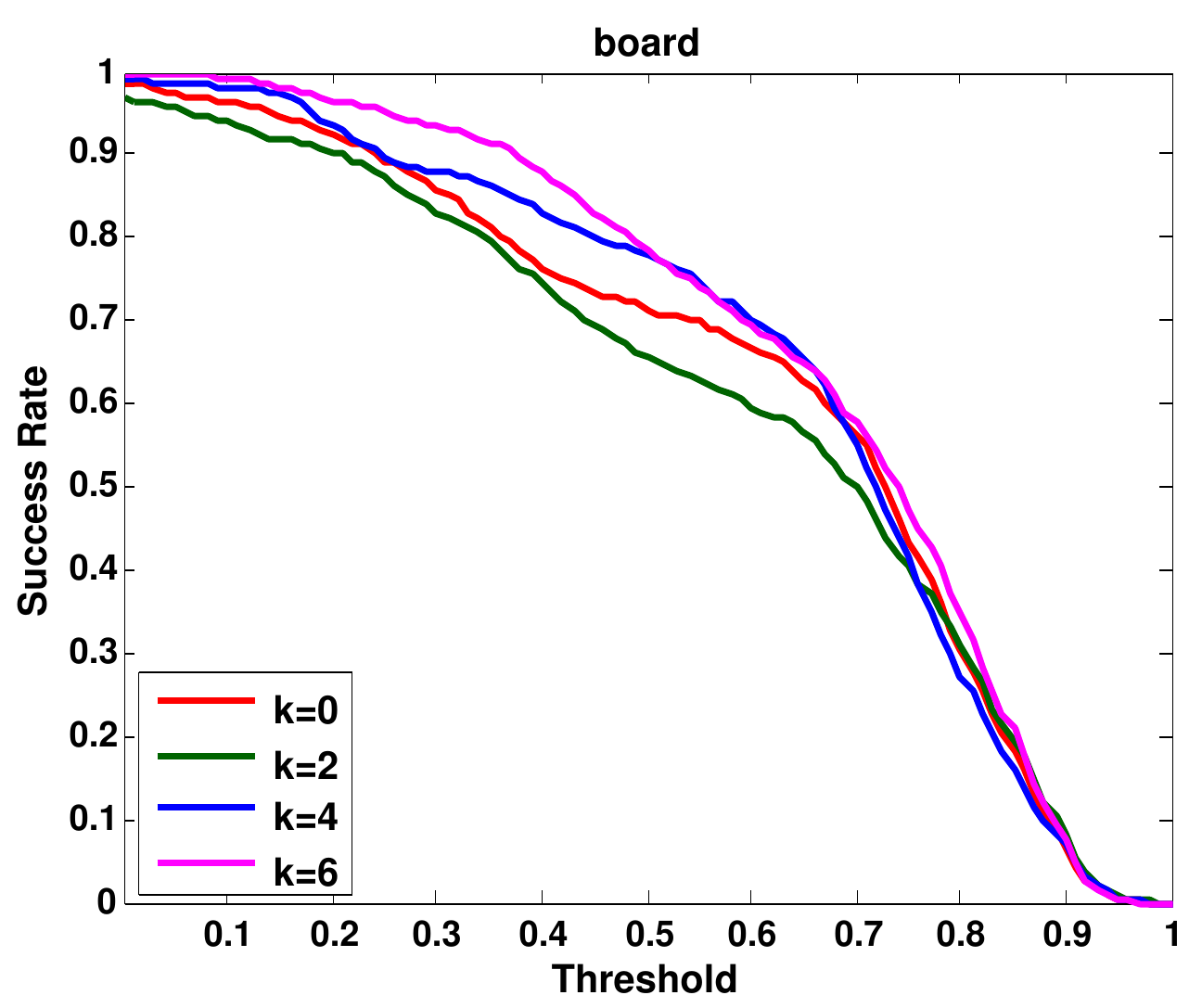}
 \includegraphics[width=0.25\textwidth]{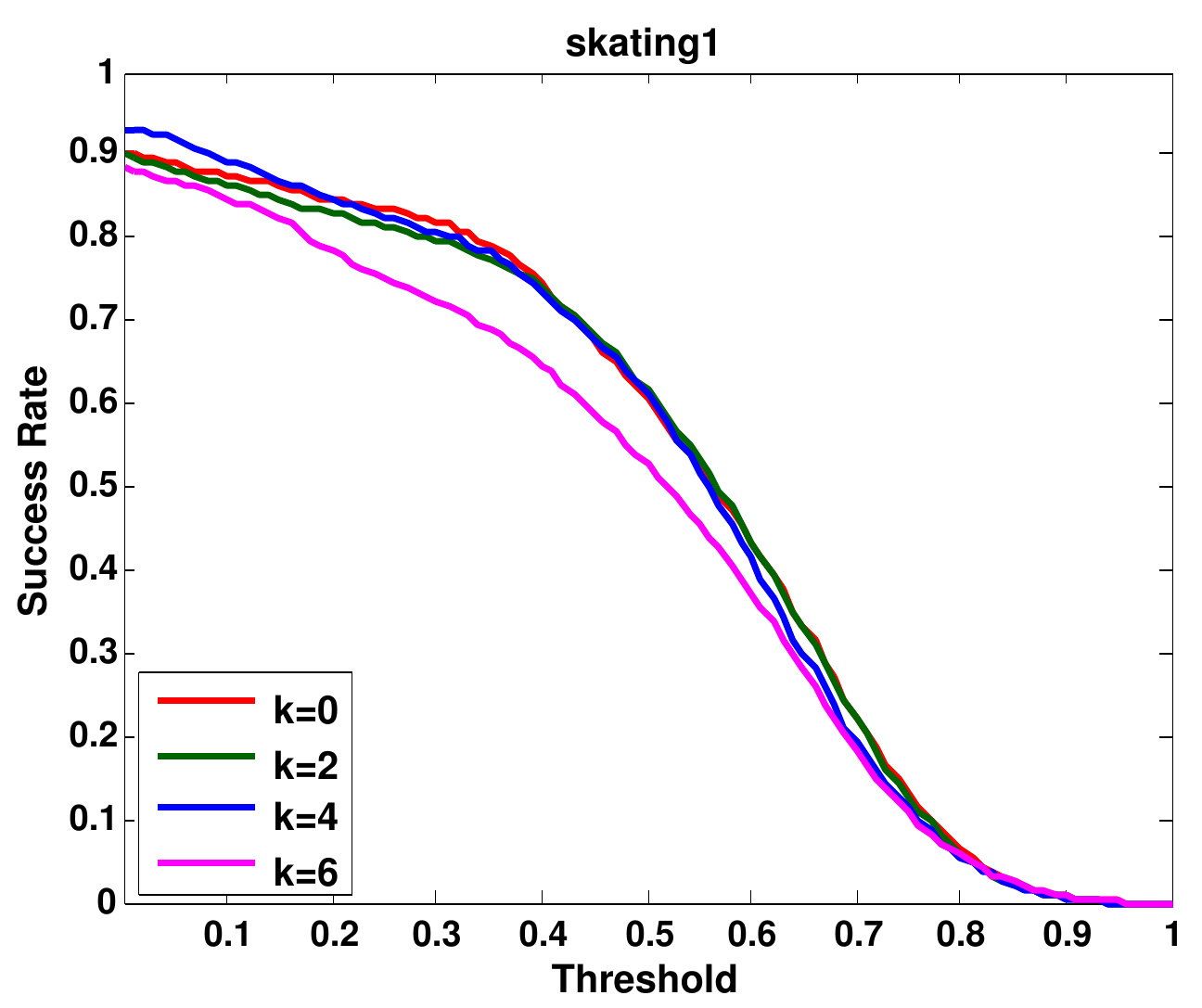}
\includegraphics[width=0.25\textwidth]{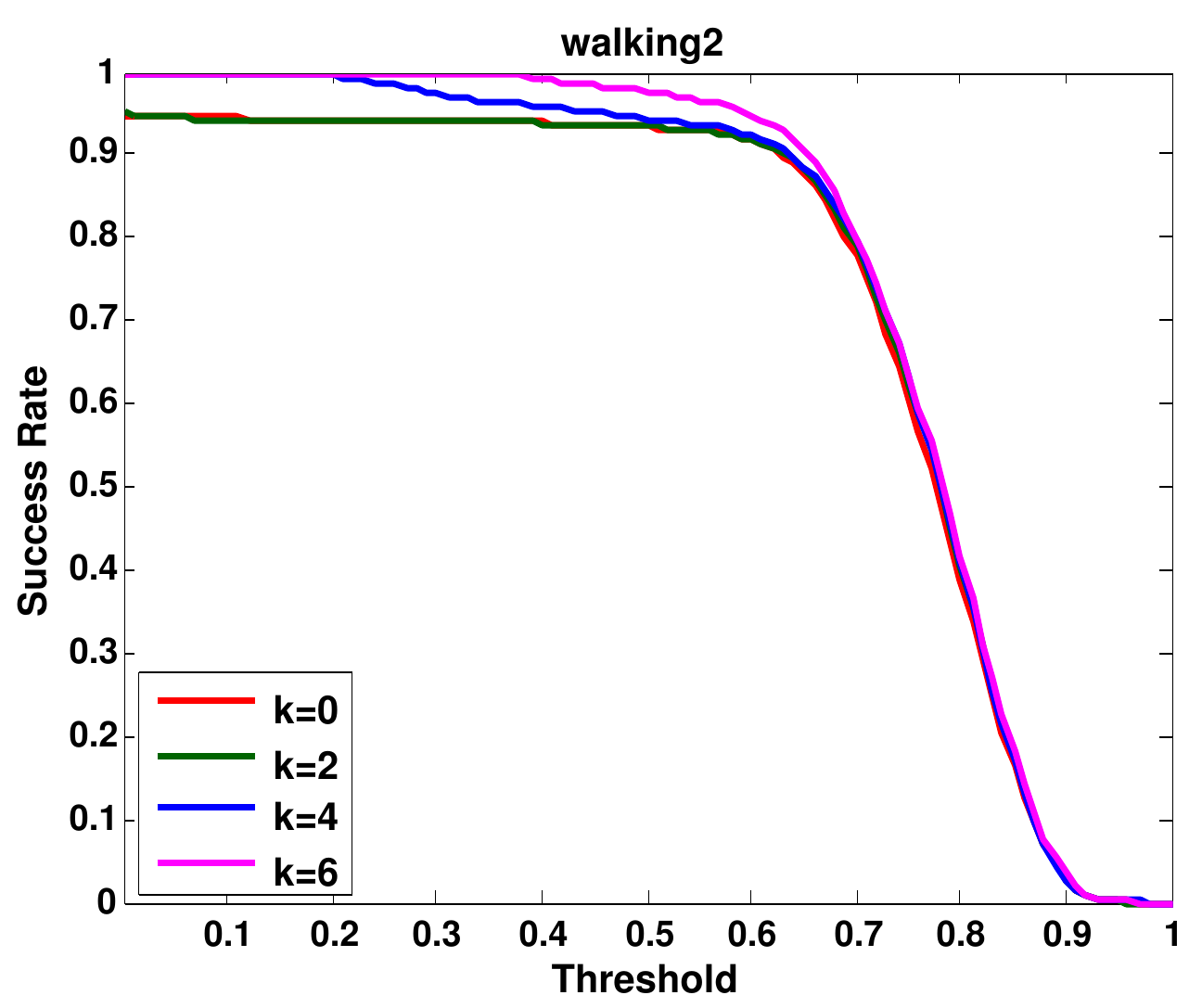}
\caption{\textbf{Group size effect.} Effect of group size on tracking performance of PJS-S. The results were averaged over 10 runs.}
\label{fig:gpsEffect}
\end{figure*}
The average success rate in all datasets for the two best trackers, PJS-S and PJS-M, is $69\%$ and $65\%$, respectively. Using one specific threshold for tracker evaluation may not be adequately representative of its performance; Therefore similar to \cite{Wu2013} we have used a success plot. The success plot shows the ratio of successful frames as the threshold varies from 0 to 1. The average success plots are shown in Fig. \ref{fig:avgROC} for all datasets and trackers. These plots show that the proposed trackers, as previously shown by using other measures, achieved the best success rate for all datasets, except \textit{faceocc2}. 

To further investigate the effect of joint sparse coding on the proposed tracker's (PJS-S) performance, the success plot for different values of group size, $k=0, 2, 4 ,6$ is depicted in Fig. \ref{fig:gpsEffect} for \textit{board},  \textit{skating1}, and \textit{walking2} datasets. As expected, increasing the group size will increase tracker performance, as long as the temporal similarity assumption is not violated . In \textit{board} and \textit{walking2} group size $k=6$ and $k=4$ obtained the best results. 
On the other hand, in \textit{skating1} due to sever occlusions and fast movements, group sizes higher than $k=4$ invalidate our assumption and consequently the performance has decreased for $k=6$. 

\section{Conclusion}\label{sec:conclusion}
In this paper, we exploited the temporal similarity assumption by using joint sparsity to enforce the targets in consecutive frames belong to a common subspace. The target image was viewed as a collection of non-overlapping patches.
This resulted in a likelihood measure which is calculated as a sum over patch-reconstruction errors. Furthermore, an occlusion detection scheme was introduced by taking advantage of patch-reconstruction errors, and a prior probability of occlusion derived from an adaptive Markov chain of occlusion states. 
Extensive experimental results showed the effectiveness of the proposed approach on a variety of video sequences. In particular, our observations show that interpreting the target as a collection of smaller patches reduced the effects of occlusion by isolating its effects to the occluded patches. This is further important for updating the dictionary, because the occluded patches can be excluded from this procedure.
In addition, we have observed that imposing the temporal similarity assumption generally improved the tracker's performance against appearance change, although this improvement was less profound for video sequences with rapid changes in appearance.
Many directions may be followed for future work. The temporal similarity assumption may be imposed in other manners, the tracker's speed can be improved to make it suitable for real-time applications, and finally a weighted sum of patch-reconstruction errors may be able to improve the tracker's performance by assigning less weight to marginal patches that often contain the background and are more prone to occlusion.

\ifCLASSOPTIONcaptionsoff
  \newpage
\fi


\bibliographystyle{IEEEtran}
\bibliography{IEEEabrv,ref}

\begin{IEEEbiography}[{\includegraphics[width=1in,height=1.25in,clip,keepaspectratio]{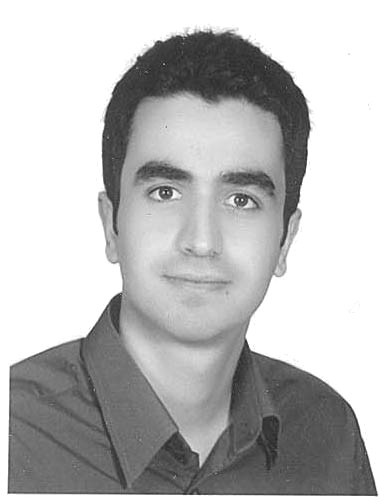}}]{Ali Zarezade}
received his B.Sc. and M.Sc in Aerospace Engineering from Amirkabir and Sharif University of Technology, Tehran, Iran, in 2007 and 2009, respectively. He also received an  M.Sc. in Artificial Intelligence from Sharif University of Technology, Tehran, Iran in 2013, and is currently working toward the Ph.D. degree in the Department of Computer Engineering at Sharif University of Technology.

His current research interests include the application of machine learning and sparse representation in computer vision and visual tracking.
\end{IEEEbiography}

\vspace*{-2\baselineskip}
\begin{IEEEbiography}[{\includegraphics[width=1in,height=1.25in,clip,keepaspectratio]{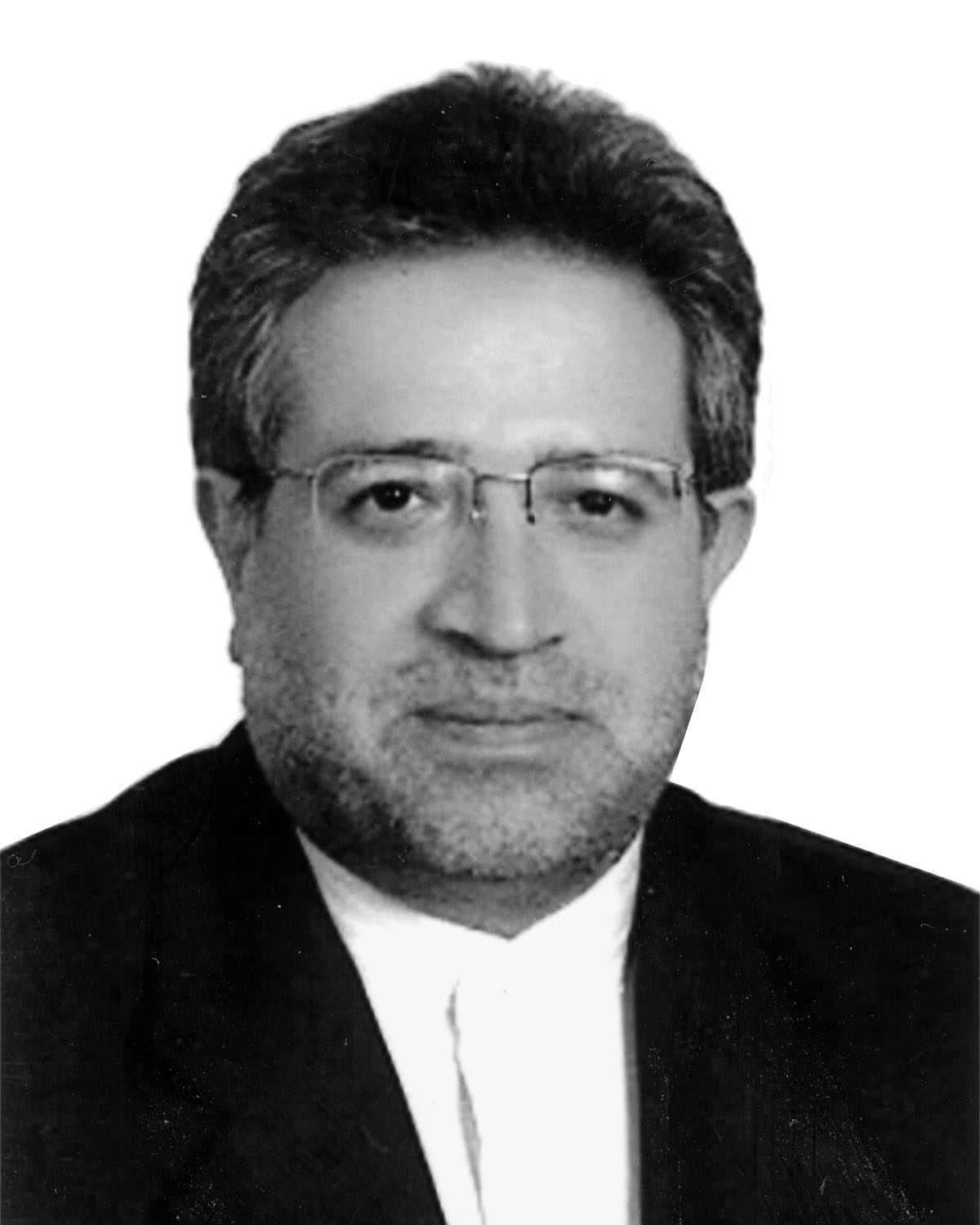}}]{Hamid R. Rabiee}
(SM'07) received his B.S. (1987) and M.S. (1989) degrees (with great distinction) in Electrical Engineering from CSULB, USA, his EEE in Electrical and Computer Engineering from USC, USA, and his Ph.D. (1996) in Electrical and Computer Engineering from Purdue University, West Lafayette, USA.

From 1993 to 1996 he was a Member of Technical Staff at AT\&T Bell Laboratories. From 1996 to 1999 he worked as a Senior Software Engineer at Intel Corporation. He was also with PSU, OGI, and OSU Universities as an adjunct professor of Electrical and Computer Engineering from 1996 to 2000. Since September 2000, he has joined Sharif University of Technology (SUT), Tehran, Iran. He is the founder of Sharif University Advanced Information and Communication Technology Research Center (AICT), Advanced Technologies Incubator (SATI), Digital Media Laboratory (DML) and Mobile Value Added Services Laboratory (VASL). He is currently a Professor of Computer Engineering at Sharif University of Technology and the Director of AICT, DML and VASL. He has been the initiator and director of national and international level projects in the context of UNDP International Open Source Network (IOSN) and Iran National ICT Development Plan.

Prof. Rabiee has received numerous awards and honors for his industrial, scientific and academic contributions. He has acted as chairman in a number of national and international conferences, and holds three patents.
\end{IEEEbiography}
\vspace*{-2\baselineskip}

\begin{IEEEbiography}[{\includegraphics[width=1in,height=1.25in,clip,keepaspectratio]{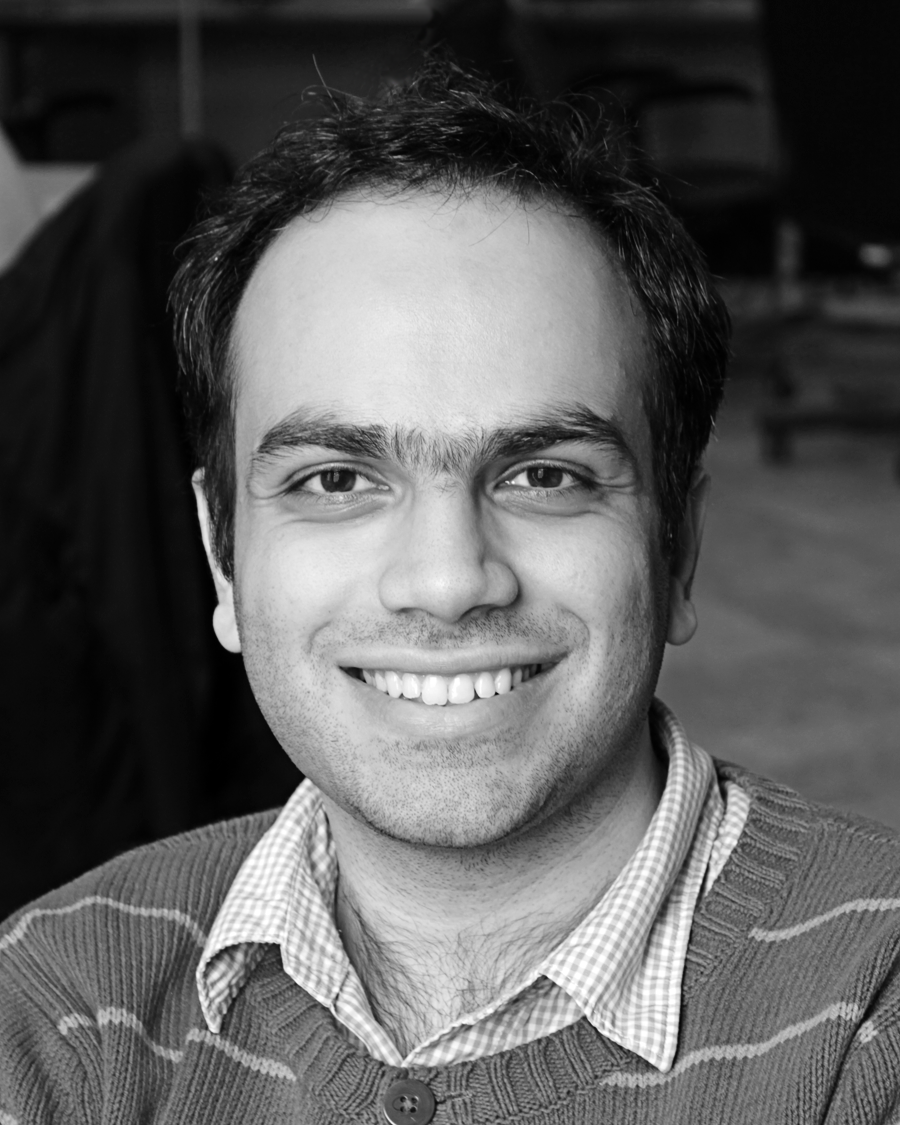}}]{Ali Soltani-Farani}
received his M.Sc. in Computer Architecture from Sharif University of Technology, Tehran, Iran, in 2010 and a B.Sc. in Hardware Engineering from the same university in 2008. He is currently working toward the Ph.D. degree in the Department of Computer Engineering at Sharif University of Technology.

His current research interests include structured sparse representations, dictionary learning, and their application to signal and image processing.
\end{IEEEbiography}
\vspace*{-2\baselineskip}

\begin{IEEEbiography}[{\includegraphics[width=1in,height=1.25in,clip,keepaspectratio]{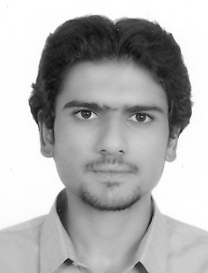}}]{Ahmad Khajenezhad}
received his M.Sc. in Artificial Intelligence from Sharif University of Technology, Tehran, Iran, in 2012 and a B.Sc. in Software Engineering from the same university in 2010. He is currently working toward the Ph.D. degree in the Department of Computer Engineering at Sharif University of Technology.

His current research interests include the application of machine learning and sparse representation in computer vision and visual tracking.
\end{IEEEbiography}

\end{document}